\pdfoutput=1
\documentclass[10pt,twocolumn,letterpaper]{article}

\usepackage[pagenumbers]{cvpr}  %

\usepackage[final]{microtype}
\usepackage{cuted}      %
\usepackage{currfile}
\usepackage{caption}   %
\usepackage{multirow}
\usepackage{makecell}
\usepackage{tabularx}
\usepackage{xspace}
\usepackage{bm}
\usepackage[inline]{enumitem}

\graphicspath{{figs/}{../figures/}}

\usepackage{colortbl}
\AtBeginDocument{\setlength{\textfloatsep}{6pt plus 2pt minus 2pt}\setlength{\floatsep}{6pt plus 2pt minus 2pt}%
  \setlength{\dbltextfloatsep}{6pt plus 2pt minus 2pt}\setlength{\dblfloatsep}{6pt plus 2pt minus 2pt}%
  \setlength{\intextsep}{6pt plus 2pt minus 2pt}}
\makeatletter\g@addto@macro\normalsize{\setlength\abovedisplayskip{4pt plus 2pt minus 2pt}\setlength\belowdisplayskip{4pt plus 2pt minus 2pt}%
  \setlength\abovedisplayshortskip{1pt plus 1pt}\setlength\belowdisplayshortskip{2pt plus 1pt minus 1pt}}\makeatother
\renewcommand{\arraystretch}{0.85}
\AtBeginDocument{\setlength{\aboverulesep}{0.15ex}\setlength{\belowrulesep}{0.3ex}}  %
\AtBeginEnvironment{table}{\renewcommand{\arraystretch}{0.8}}\AtBeginEnvironment{table*}{\renewcommand{\arraystretch}{0.8}}
\usepackage{pifont}
\usepackage{tikz}
\usetikzlibrary{arrows.meta,calc,shadows,shadows.blur}
\usepackage[scaled=0.92]{helvet}   %
\definecolor{oursbg}{HTML}{E6F6F8}     %
\definecolor{groupfg}{HTML}{4A45C9}    %
\definecolor{fireO}{HTML}{EF5323}
\definecolor{fireM}{HTML}{FF8A1F}
\definecolor{fireI}{HTML}{FFD23F}
\DeclareRobustCommand{\fire}{\tikz[baseline=0.05em,x=0.0105em,y=0.0105em]{%
  \fill[fireO] (50,0) .. controls (18,0) and (2,24) .. (8,50) .. controls (12,70) and (26,80) .. (30,100)
       .. controls (40,88) and (42,76) .. (40,64) .. controls (52,74) and (58,90) .. (56,108)
       .. controls (80,92) and (96,64) .. (94,40) .. controls (92,14) and (74,0) .. (50,0) -- cycle;
  \fill[fireM] (50,6) .. controls (28,6) and (18,24) .. (22,42) .. controls (25,56) and (34,62) .. (36,74)
       .. controls (44,66) and (46,56) .. (45,48) .. controls (55,56) and (60,68) .. (60,80)
       .. controls (76,66) and (84,48) .. (80,30) .. controls (76,14) and (64,6) .. (50,6) -- cycle;
  \fill[fireI] (51,10) .. controls (38,10) and (31,22) .. (34,34) .. controls (37,44) and (44,50) .. (48,58)
       .. controls (54,50) and (66,40) .. (66,28) .. controls (66,16) and (60,10) .. (51,10) -- cycle;}}
\definecolor{snowD}{HTML}{3D7BD8}
\definecolor{snowL}{HTML}{BCD7F4}
\DeclareRobustCommand{\snow}{\tikz[baseline=0.05em,x=0.0105em,y=0.0105em]{%
  \foreach \a in {90,150,210,270,330,390}{\draw[snowL, line width=0.17em, line cap=round] (50,52) -- ++(\a:44);}%
  \foreach \a in {90,150,210,270,330,390}{%
    \draw[snowD, line width=0.075em, line cap=round] (50,52) -- ++(\a:43);%
    \draw[snowD, line width=0.075em, line cap=round] ([shift={(\a:24)}]50,52) -- ++(\a+42:13);%
    \draw[snowD, line width=0.075em, line cap=round] ([shift={(\a:24)}]50,52) -- ++(\a-42:13);}%
  \fill[snowD] (50,52) circle (7);\fill[white] (50,52) circle (3);}}

\DeclareRobustCommand{\sota}[1]{\textbf{#1}\kern0.08em\fire}   %

\newcommand{\best}[1]{\textbf{#1}}                  %
\newcommand{\second}[1]{\underline{#1}}             %
\newcommand{\cmark}{\ding{51}}
\newcommand{\xmark}{\ding{55}}
\newcommand{\grouprow}[2]{\multicolumn{#1}{@{}l}{\color{groupfg}\textit{#2}}\\}
\definecolor{dnC}{HTML}{C13C95}\definecolor{upC}{HTML}{0E9AA7}
\newcommand{\dn}[1]{{\color{dnC}$#1$}}
\newcommand{\dnb}[1]{{\color{dnC}\boldmath$#1$}}
\newcommand{\up}[1]{{\color{upC}$#1$}}
\definecolor{venueC}{HTML}{6B7280}
\newcommand{\venue}[1]{\,{\color{venueC}\scriptsize(#1)}}
\newcommand{\pub}[1]{{\leavevmode\color{gray}#1}}             %

\newif\ifdraft
\draftfalse  %

\newcommand{\pending}[1]{{\color{red}\textbf{pending}}}

\newcommand{\ours}{When2Go\xspace}
\newcommand{\cod}{CoDriving\xspace}
\newcommand{\dtp}{Defer to Plan\xspace}

\DeclareMathOperator*{\argmax}{arg\,max}

\DeclareMathOperator{\cummax}{cummax}
\newcommand{\R}{\mathbb{R}}

\newcommand{\vect}[1]{\bm{#1}}

\newcommand{\gr}[1]{{\leavevmode\color{gray}#1}}

\usepackage{times}
\usepackage{graphicx}
\usepackage{amsmath}
\usepackage{amssymb}
\usepackage{booktabs}

\definecolor{cvprblue}{rgb}{0.21,0.49,0.74}
\usepackage[pagebackref,breaklinks,colorlinks,allcolors=cvprblue]{hyperref}

\def\confName{CVPR}
\def\confYear{2027}

\title{Route-Time Conflict Maps for Cooperative\texorpdfstring{\\}{ }End-to-End Autonomous Driving}

\author{Haohua Que \quad Mingkai Liu \quad Tianle Zhu \quad Zhipeng Bao \quad Yiping Li \quad Handong Yao\textsuperscript{*}\\
University of Georgia\\[2pt]
{\small \textsuperscript{*}Corresponding author, \href{mailto:handong.yao@uga.edu}{\texttt{handong.yao@uga.edu}}}}

\begin{document}
\maketitle
\begin{strip}
  \centering
  \input{sections/teaser_body}
  \captionsetup{hypcap=false}
  \captionof{figure}{\textbf{When2Go learns when to go.} InterDrive at $t{=}5$\,s: eight connected vehicles swap lanes. Left: each vehicle senses with its LiDAR (dots), broadcasts its detections and its $2$\,s plan, and decides when to go; numbers and arcs give the pass order the vehicles agree on (equal numbers: no conflict), bars mark vehicles that wait. Cards: the vehicles' route-time conflict maps, the others' plans as occupancy over route position $s$ and time $t$, with the executed speed profile. Top right: the same instant in CARLA; all eight vehicles arrive without a collision. Bottom right: \ours against the best other method on each axis.}
  \captionsetup{hypcap=true}
  \label{fig:teaser}
  \vspace{-2mm}%
\end{strip}

\begin{abstract}
Existing cooperative end-to-end driving works at Class~A (status-sharing) of the SAE J3216 classes of cooperative driving automation (CDA): vehicles exchange what they perceive, not what they intend, so each plans alone when to go, which limits cooperation and leads to deadlock when both vehicles yield or to collisions when both go.
We present \ours, a cooperative end-to-end driving system that learns when to go by sharing intent under CDA Class~B or higher.
Each vehicle broadcasts its LiDAR detections with forecast motion as instance tokens, together with its plan and intended trajectory.
The receiver draws the fused detections, forecasts and partner plans into a route-time conflict map, the occupancy along its route over the planning horizon, on which a planner scores candidate speed profiles.
Where two intended trajectories conflict, an order gate, a precedence network distilled from a right-of-way rule, decides who goes first; both vehicles evaluate it on the same messages and so reach the same order.
\ours imitates the rule-based expert of V2Xverse, which sees the simulator's ground truth; stops that \ours cannot observe within its $2$\,s planning horizon are relabelled, so that it does not learn to wait without an observable cause.
For our runs only, the benchmarks' PID controller steers by \ours's waypoints instead of the route, and its forced push of long-waiting vehicles is off.
With one model, \ours sets the state of the art on three closed-loop CARLA benchmarks: $97.8$ and $90.7$ driving score on InterDrive without and with traffic (best published $90.1$ and $78.9$), $70.0$ on the public V2Xverse protocol (best of sixteen rerun methods $59.1$) and $88.2$ on MDrive (best rerun $85.1$), with $95.0\,\%$ of its vehicles collision-free ($87.7\,\%$).
\end{abstract}

\vspace{-1.5mm}\section{Introduction}\label{sec:intro}

Connected vehicles can see for each other: a partner's LiDAR covers the crosswalk behind a truck, a roadside unit sees the whole junction.
SAE J3216~\cite{sae2025j3216} divides cooperative driving automation (CDA) into four classes: vehicles share their status and perception (Class~A, status-sharing) or what they plan to do (Class~B, intent-sharing), exchange messages to agree on their actions (Class~C, agreement-seeking), or one party directs the others (Class~D, prescriptive).
Existing cooperative end-to-end driving~\cite{liu2025codriving,li2026defer,yu2025univ2x} works at Class~A: it exchanges what the vehicles perceive, not what they intend.
Yet where vehicles meet, at intersections, merges and lane changes, the difficult part is not to see the other vehicle but to decide, for every vehicle, when to go, and each decides alone: if both yield, they wait until the benchmark ends the run as blocked; if both go, they collide.
Language-based negotiation~\cite{liu2025colmdriver,colmin2026}, a form of Class~C, settles such conflicts with language and vision-language models but degrades in dense traffic and under pose noise~\cite{mdrive2026}.

We start from the observation that this decision is one-dimensional along the vehicle's own route: is the stretch ahead free at the time the vehicle would be there?
\ours draws everything that can block this stretch, the fused detections with their forecast motion and the plans that partner vehicles broadcast, as occupancy over route position and future time: a \emph{route-time conflict map}.
A planner scores candidate speed profiles on this map and learns how far to drive in the next two seconds, from standing still to full speed (\cref{fig:overview}); the benchmarks' PID controller tracks its waypoints, and no hand-written rule overrides them.

Two further ingredients make the decisions of several vehicles fit together.
First, the vehicles share intent (Class~B): each also broadcasts an intended trajectory, how it would drive along its route if the road were free.
Where two intended trajectories meet, an \emph{order gate}, a precedence network distilled from a right-of-way rule, decides who goes first.
Both vehicles evaluate it on the same two messages, so with an ideal channel they reach the same order without the message rounds of Class~C; each draws the order into its conflict map, and the planner chooses the speed within it.
Second, \ours imitates the rule-based expert of V2Xverse~\cite{liu2025codriving}, which sees the simulator's ground truth and often stops for reasons that \ours cannot observe within its $2$\,s planning horizon, \eg because another vehicle, assumed to keep driving straight, might cross its path.
Copying these stops teaches \ours to wait without an observable cause, and two such vehicles deadlock; we relabel them with targets based only on what it can observe within this horizon, a temporal form of the learner-expert asymmetry~\cite{nguyen2026lead}.

We evaluate one model in closed loop on three CARLA~\cite{dosovitskiy2017carla} benchmarks with one to eight controlled vehicles (\cref{fig:teaser}).
On InterDrive~\cite{liu2025colmdriver}, \ours reaches $97.8$ and $90.7$ driving score without and with background traffic (best published $90.1$ and $78.9$).
On MDrive~\cite{mdrive2026}, it reaches $88.2$ (best rerun $85.1$), with $95.0\,\%$ of its vehicles collision-free against at most $87.7\,\%$.
On V2Xverse~\cite{liu2025codriving}, where we rerun sixteen published methods and two variants on the public protocol (all from released checkpoints but one reimplementation), it reaches $70.0$ against $59.1$ and collides with another vehicle in $6$ of $100$ runs, against $38$ to $69$ for the cooperative reruns. %
Our contributions are:
\begin{itemize}[leftmargin=*,itemsep=1pt,topsep=2pt]
  \item \textbf{A route-time conflict map for cooperative end-to-end planning}: fused detections, forecasts and partner plans as occupancy along the ego's future route, from which a planner learns whether the route ahead will be clear on arrival, supervised by an expert that sees more, whose stops beyond the $2$\,s horizon are relabelled; to our knowledge, the first CDA Class~B-or-higher end-to-end planner.
  \item \textbf{An order gate for agreement from shared messages}: an antisymmetric precedence network on the broadcast intended trajectories yields a cycle-free order, drawn into the conflict map as occupancy.
  \item \textbf{State-of-the-art closed-loop results} on three benchmarks with one end-to-end model, with reruns of sixteen methods and two variants on the public V2Xverse protocol and paired $95\,\%$ confidence intervals for the differences.
\end{itemize}

\vspace{-1.5mm}\section{Related Work}\label{sec:related}

\noindent\textbf{Cooperative perception and end-to-end driving.}
Cooperative perception fuses raw points~\cite{chen2019cooper}, features~\cite{chen2019fcooper,wang2020v2vnet,xu2022opv2v,xu2022v2xvit} or objects, selects what to send~\cite{hu2022where2comm} and copes with latency~\cite{lei2022syncnet,wei2023cobevflow}, pose error~\cite{lu2023coalign} and lossy links~\cite{ren2024incop,li2023lossy}.
Cooperative end-to-end planners build on it, in closed loop on V2Xverse~\cite{liu2025codriving,li2026defer,select2drive2025}, on logged data~\cite{yu2025univ2x,unimmv2x2026,omniv2x2026,cosv2x2026,riskmm2026,aurora2026,you2026v2xwam} and on real vehicles~\cite{qu2026cooperdrive}.
They exchange perception, not intent, and leave who goes first to each planner.

\noindent\textbf{Coordinating connected vehicles.}
CoLMDriver~\cite{liu2025colmdriver}, introduced with InterDrive, lets a language model prompted with right-of-way rules negotiate a speed command for each vehicle of a conflict group; CoLMIN~\cite{colmin2026} scores several joint assignments per round, related agents reason over shared intents with language or vision-language-action models~\cite{fang2025codrivingllm,gao2025langcoop,chiu2026v2vgot,li2026n2cdrive,chiu2026cmudrive}, and negotiation degrades in dense traffic and under pose noise~\cite{mdrive2026}.
Without language, intersection managers reserve space-time cells centrally~\cite{dresner2008multiagent} or by broadcast~\cite{vanmiddlesworth2008replacing,azimi2014stip}, vehicles broadcast a desired next to a planned trajectory~\cite{lehmann2018generic,wartnaby2019decentralised}, prioritized planning fixes an order among robots~\cite{erdmann1987multiple,vandenberg2005prioritized,cap2015prioritized,ma2019searching}, learned models choose crossing orders or priority graphs~\cite{zhang2023alphaorder,klimke2022cooperative,klimke2023automatic,mertens2025generalized,zhang2026tsc}, relation heads order agents in motion prediction~\cite{sun2022m2i,rowe2023fjmp}, and agents acting on common knowledge coordinate without negotiation~\cite{dewitt2019mackrl}.
Our precedence network applies the antisymmetric form of pairwise ranking~\cite{burges2005ranknet,koppel2019directranker} and ranked-pairs decoding~\cite{tideman1987independence} to decentralised agreement, and its result enters the planner as occupancy.

\noindent\textbf{Planning representations and supervision.}
End-to-end planners use module stacks~\cite{hu2023uniad}, vectorised~\cite{jiang2023vad} or sparse~\cite{sun2025sparsedrive} scene elements.
Speed profiles along a path are searched in path-time graphs~\cite{kant1986path,werling2010optimal,fan2018baidu} or scored on space-time costs, predicted occupancy or forecasts~\cite{zeng2019nmp,sadat2020p3,casas2021mp3,biswas2024quad,liu2023opgp,dauner2023parting}, also with factorised path and speed vocabularies~\cite{sun2026sparsedrivev2} or as displacement along the path~\cite{wu2026aligndrive}; our map adds the broadcast partner plans along the route.
Imitating a privileged rule-based expert~\cite{chitta2023transfuser,liu2025codriving} brings covariate shift~\cite{ross2011dagger,cui2022coopernaut,zhong2026rogdagger}, causal confusion~\cite{dehaan2019causal,codevilla2019limitations,wen2020copycat,bansal2019chauffeurnet} and a dependence on the expert's style~\cite{zimmerlin2024hidden}; rule-based experts score candidates~\cite{li2024hydramdp}, and imitation adapts to what the learner cannot observe~\cite{weihs2021advisor,warrington2021a2d,nguyen2026lead}.
We address the same asymmetry in time: expert stops that \ours cannot explain within its $2$\,s planning horizon are relabelled.

\vspace{-1.5mm}\section{Method}\label{sec:method}

\begin{figure*}[t]
  \centering
  \input{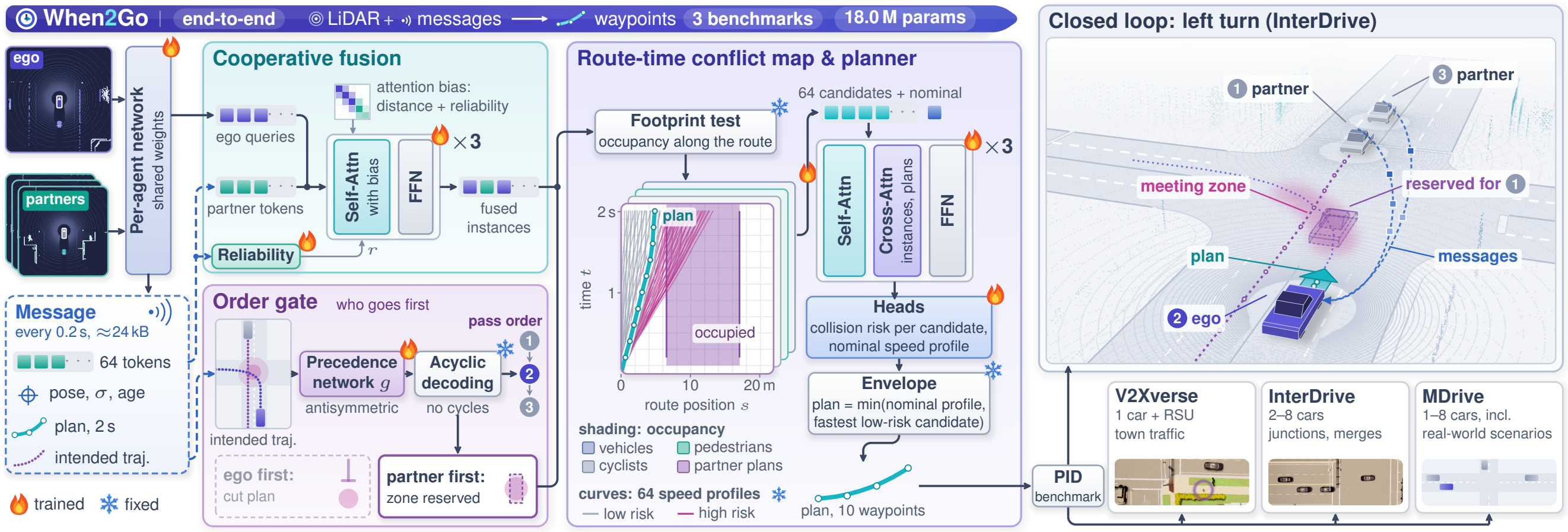}
  \caption{\textbf{Overview of \ours.} Each vehicle runs the same network on its LiDAR and broadcasts its detections, forecasts and plan. The receiver draws them into a route-time conflict map, on which the planner scores $64$ speed profiles; the order gate decides who goes first. Right: a logged step, in which the oncoming car goes first (numbers: pass order).}
  \label{fig:overview}
\end{figure*}

\vspace{-1mm}\subsection{Problem setting and overview}\label{sec:setting}

\ours drives $N$ connected vehicles in closed loop; each runs its own copy of one end-to-end network, without a coordinator: one forward pass maps its LiDAR sweeps, status and received messages to waypoints (\cref{fig:overview}), and no rule overrides them.
Every $\Delta_t{=}0.2$\,s a vehicle reads three sweeps of the benchmark's LiDAR (\cref{sec:supp_setup}) within $\pm48$\,m and a status vector: speed, navigation command, a target point, its positions over the last second and the route ahead (ten points $3$\,m apart).
Vehicles and roadside units (RSUs) broadcast instance tokens; vehicles add their plan and an intended trajectory. A receiver fuses at most $P{=}3$ partner sources and, on InterDrive and MDrive, reads up to three more plans within $70$\,m.
A vehicle outputs a plan $\vect{p}_{1:T}$: the positions of its \emph{reference point} (its LiDAR) at the times $t_k{=}k\Delta_t$, $k{=}1,\dots,T$, with $T{=}10$.
The benchmarks' PID controller~\cite{liu2025codriving} tracks the plan: speed from the first waypoints and, unlike its default, steering towards the mean of those at $0.8$ and $1.0$\,s instead of a point of the global route.

\vspace{-1mm}\subsection{Perception, forecasting and cooperative fusion}\label{sec:perception}

Perception uses standard parts, and vehicles and RSUs share one network: a pillar encoder~\cite{lang2019pointpillars} and BEV backbone, $K{=}200$ queries from the peaks of a centre heatmap~\cite{yin2021centerpoint}, and three decoder layers with deformable attention~\cite{zhu2021deformable}.
Per instance $n$, it predicts class probabilities $q_{n,c}$, a box with centre $\vect{x}_n$, a Laplace scale $\vect{\beta}_n$ of the centre and $J{=}6$ forecast modes over $2$\,s with probabilities $\pi_{n,j}$.
A message holds the $64$ highest-scoring instances of the sender (a $64$-dimensional feature, centre, heading, mean forecast, scale, class probabilities), the sender pose with standard deviations $\sigma_{xy}$, $\sigma_\psi$, the message age $\alpha$ in frames and the source type.
With plan and intended trajectory, a vehicle's message has about $6{,}000$ values ($24$\,kB in float32, $12$\,kB at $16$ bits) every $0.2$\,s, without packet format.
The receiver moves partner instances into its frame with the broadcast pose and advances each centre along its mean forecast by the age.
An instance has the position variance $V_n$: the mean of $\beta_n^2$ over both axes plus, for partner instances, $\sigma_{xy}^2+(\sigma_\psi D_n)^2+(0.5\,\alpha)^2$, with $D_n$ the distance from the sending sensor.
Each source $m$, the ego included, gets a reliability logit $r_m$ from an MLP of message statistics.
Three fusion layers process ego queries and partner tokens as one sequence, with the self-attention bias from query $n$ to key $n'$ of source $m(n')$
\begin{equation}\label{eq:bias}
  B_{nn'} = -\frac{\lVert\vect{x}_n-\vect{x}_{n'}\rVert^2}{2\,(V_n+V_{n'}+1)} + r_{m(n')} .
\end{equation}

\vspace{-1mm}\subsection{Route-time conflict map}\label{sec:map}

\begin{figure}[t]
  \centering
  \resizebox{\linewidth}{!}{\input{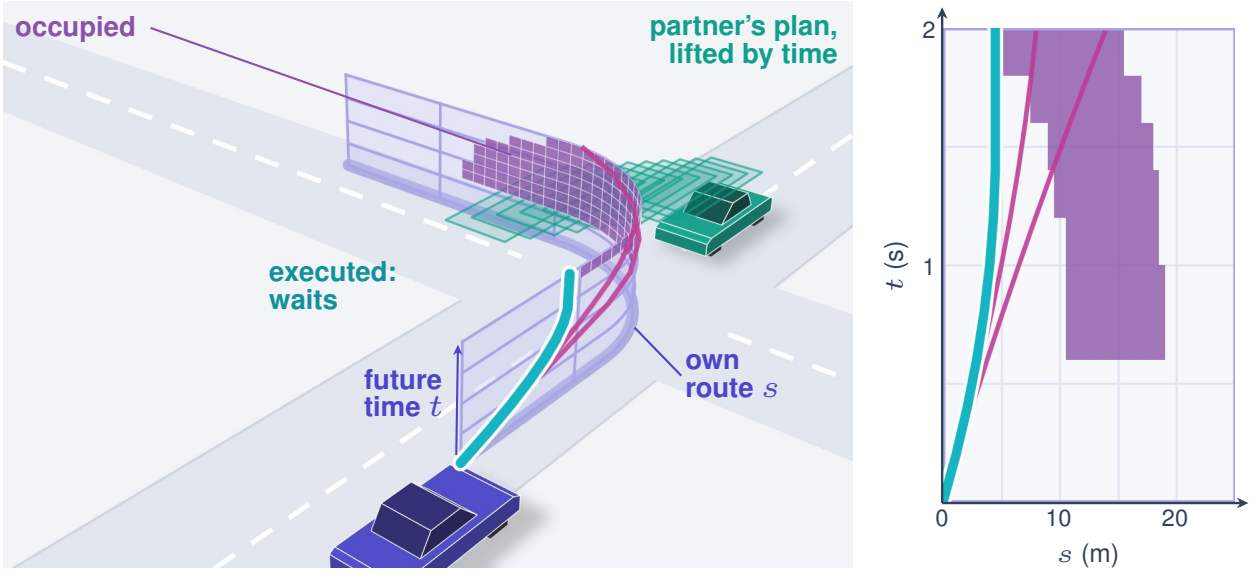}}
  \caption{\textbf{Route-time conflict map.} Left: a curtain along the ego's route, its height future time; a box lifted to its future time overlaps the ego where it crosses the curtain. The speed is capped by the fastest profile that stays clear (magenta: those that do not); here the ego waits. Right: the curtain unrolled is the map.}
  \label{fig:conflict}
\end{figure}

Let $s$ be the arclength along the route $\gamma$, the polyline from the reference point through its ten points.
The map $\vect{M}\in[0,1]^{4\times T\times S}$ (cf.\ BEV cost volumes~\cite{zeng2019nmp}, safe-distance bounds along a path~\cite{shao2022interfuser}) has one row per future time $t_k$ and $S{=}60$ columns of $\Delta_s{=}0.5$\,m with centres $\bar s_i=(i-\tfrac12)\Delta_s$; its channels are vehicles, pedestrians, cyclists and partner plans.
$M_c(k,i)$ is the probability that an ego with reference point at $\bar s_i$ at time $t_k$ overlaps an object of channel $c$ (\cref{fig:conflict}).

\noindent\textbf{Footprint test.}
For a box $b$ ($l_b\times w_b$) whose centre at $t_k$ has arclength $s_b$, lateral offset $d_b$ and heading $\phi_b$ relative to $\gamma$,
\begin{align}
  \xi^{\parallel}_b &= \tfrac{l_b}{2}\lvert\cos\phi_b\rvert+\tfrac{w_b}{2}\lvert\sin\phi_b\rvert, \notag\\
  \xi^{\perp}_b &= \tfrac{l_b}{2}\lvert\sin\phi_b\rvert+\tfrac{w_b}{2}\lvert\cos\phi_b\rvert, \notag\\
  I_b &= \bigl[\,s_b-\xi^{\parallel}_b-e_f,\;\; s_b+\xi^{\parallel}_b+e_r\,\bigr], \label{eq:stretch}\\
  \lambda_b &= \operatorname{clip}\Bigl(\frac{h+\xi^{\perp}_b+\mu-\lvert d_b\rvert}{\epsilon}+\frac12,\;0,\;1\Bigr), \label{eq:lat}\\
  o_b(k,i) &= \lambda_b\,\mathbf{1}[\,\bar s_i\in I_b\,]. \label{eq:occ}
\end{align}
$I_b$ is the stretch of reference-point positions where ego and box overlap along the route (ego bumpers $e_f{=}1.15$\,m ahead, $e_r{=}3.75$\,m behind).
The weight $\lambda_b$ (ego half width $h{=}1.06$\,m, margin $\mu{=}0.3$\,m, soft edge $\epsilon{=}0.5$\,m) is $\tfrac12$ at a clearance of $\mu$; $\lambda_b>\tfrac12$ means \emph{in the corridor}.

\noindent\textbf{Channels.}
Each forecast mode $j$ of a fused instance $n$ is a box trajectory with occupancy $o_{n,j}$.
Modes exclude each other, so $o_n=\min\bigl(1,\sum_{j}\pi_{n,j}\,o_{n,j}\bigr)$, and instances are treated as independent, so for each object class $c$
\begin{equation}\label{eq:union}
  M_c = 1-\textstyle\prod_{n}\bigl(1-q_{n,c}\,o_n\bigr).
\end{equation}
The fourth channel holds the plans of partner vehicles, moved into the ego frame and drawn as boxes of the ego's own size ($4.9\times2.12$\,m) with weight one: no class probability, no reliability, no shift for the message age.
A connected vehicle then appears twice, so the \emph{supersede rule} removes fused detections within $2.5$\,m of a partner with a plan.

\vspace{-1mm}\subsection{Planning on the map}\label{sec:planner}

The planner holds $A{=}64$ fixed speed profiles $\vect{s}^a\in\R^T$ ($s^a_k$: distance covered along the route by $t_k$), k-means centres of the expert's profiles in the training towns, from the stop ($\vect{s}^1{=}\vect{0}$) to $13.267$\,m in $2$\,s.
Three convolutions turn $\vect{M}$ into features $\vect{F}$, and candidate $a$ samples both bilinearly along its own profile (row $k$, position $s^a_k$).
Its query holds these samples, the maximum and mean over $k$ of $M_c(k,s^a_k)$ per channel, the profile, its waypoints and the status vector.
A \emph{nominal} query embeds the status vector and pooled $\vect{F}$; three attention layers relate the queries to each other, the fused instances, the received plans and the ego BEV map.
A candidate thus meets each conflict when it would get there, not where the object is now: this separates waiting from going.

Each candidate query predicts an imitation logit $z_a$, a residual $\vect{\delta}^a$ on its profile, lateral offsets and a collision-risk logit $\rho_a$; the nominal query regresses a profile $\vect{\nu}\in\R^T$ and lateral offsets $\vect{\ell}$ from the route.
The executed plan is the regressed profile, capped by the fastest candidate that the risk head accepts (the \emph{envelope}).
With $[\cdot]_+=\max(\cdot,0)$, the running maximum $\cummax_k$ and the logistic function $\operatorname{sig}$,
\begin{equation}\label{eq:envelope}
\begin{aligned}
  \hat s_k &= \min\bigl(\cummax_k[\vect{\nu}]_+,\;v_{\max}t_k\bigr),\\
  a^\star &= \argmax_{a:\;\operatorname{sig}(\rho_a)<\theta}\bigl(s^a_T+\delta^a_T\bigr),\\
  s^\star_k &= \min\bigl(\hat s_k,\;\cummax_k[\vect{s}^{a^\star}\!+\vect{\delta}^{a^\star}]_+\bigr),
\end{aligned}
\end{equation}
with $v_{\max}{=}7.0$\,m/s and $\theta=\theta_0\min_p\operatorname{sig}(r_p)$ over the delivered partners $p$ ($\theta{=}\theta_0$ without partners), $\theta_0{=}0.5$; without a candidate below $\theta$, $a^\star$ is the one of lowest risk.
The plan is $\vect{p}_k=\gamma(s^\star_k)+\ell_k\,\gamma^{\perp}(s^\star_k)$, with $\gamma^{\perp}$ the left normal of the route.

\vspace{-1mm}\subsection{Agreement from shared messages}\label{sec:agreement}

Plans in the map do not say who gives way; instead of negotiating in rounds~\cite{liu2025colmdriver}, vehicles that meet compute one learned pairwise decision from the same messages.

\noindent\textbf{Precedence network.}
Each vehicle broadcasts an \emph{intended trajectory}: its free-flow drive along its route ($2.0$\,m/s$^2$ up to $6.0$\,m/s, or its speed if higher) as ten points $0.6$\,s apart; it ignores other vehicles, so it does not react to the other's plan.
For vehicles $u$, $u'$ with trajectories $\vect{y}^u$, $\vect{y}^{u'}$, a convolutional network $g$ over frame-independent features of all point pairs gives the logit
\begin{equation}\label{eq:prec}
  \eta_{uu'}=g(\vect{y}^u,\vect{y}^{u'})-g(\vect{y}^{u'},\vect{y}^u)=-\eta_{u'u} ,
\end{equation}
and $u$ goes first if $\eta_{uu'}>0$: for any weights, two vehicles that evaluate the same two messages cannot both claim precedence.
$g$ is trained on $400{,}000$ counterfactual pairs labelled by the right-of-way rule of \cref{sec:supervision} (with a window of $0.6$ instead of $0.4$\,s).

\noindent\textbf{Order gate.}
The precedence network is a part of \ours, and its decisions enter the conflict map.
A vehicle evaluates $\eta$ for every pair of the vehicles it hears, itself included, whose intended trajectories come within $3.0$\,m of each other.
Since pairwise precedence is not transitive, it decodes these decisions into an order without cycles: pairs are taken by decreasing $\lvert\eta\rvert$, and a pair that would close a cycle is reversed (ranked pairs~\cite{tideman1987independence}).
For each partner whose intended trajectory meets its own, the decoded order sets how that partner appears in the map: if the ego goes first, as its broadcast plan cut before the meeting zone; if the partner goes first, as a vehicle standing across the ego's path in the zone.
Both alternatives follow from the two intended trajectories by fixed geometry; the choice between them is made by the network in its forward pass, and the planner then decides how fast to drive, as for any other occupancy in the map.
Two vehicles agree when they evaluate the same messages, as with the ideal channel of our runs, which has no delay (a vehicle that plans later in a step reads the new plans of earlier ones); latency or loss can break this (\cref{sec:supp_gate}).

\vspace{-1mm}\subsection{Relabelling unexplained stops}\label{sec:supervision}

The recorded plans come from the privileged rule-based expert of the benchmark~\cite{liu2025codriving}, which logs why it brakes.
\ours plans $2$\,s ahead from LiDAR and cannot reproduce many of the expert's stops: red lights and stop signs are not in its input, the junction and cyclist rules also fire when no path enters its corridor, the pedestrian rule fires for walkers whose recorded paths never enter it in time ($28\,\%$ of the frames in which it fires in the training towns).
When the data-collecting expert has stood for $2$\,s, it deletes every vehicle and walker within $20$\,m that has stood for $1$\,s; \ours has no such rescue.

\noindent\textbf{Relabelling.}
Labels are rebuilt from the logged reasons and the recorded futures of the ground-truth boxes (privileged, training only).
The expert plan stays if its pedestrian rule fires, or if no rule fires and it moves.
Otherwise the label is a \emph{distance-keeping} profile: the fastest profile up to $6.5$\,m/s that keeps a gap of $4.0$\,m plus $1.0$\,s of travel from every box.
It is used only if it gets $0.5$\,m further in $2$\,s than the expert plan and no pedestrian is in the way.
Where the $6$\,s path of a vehicle or cyclist (three times the horizon) crosses the corridor, the label commits or yields early.
The label commits if the unhindered ego leaves the crossed stretch $1.5$\,s before the other enters it, or arrives after it has left; otherwise a label ending inside the stretch stops $3.0$\,m before it.
Red-light and stop-sign stops alone become distance keeping.

\noindent\textbf{Virtual partners.}
The recordings contain almost no conflicts between connected vehicles, so with probability $0.5$ a sample receives one to three virtual partners that exist only as broadcast plans.
A right-of-way rule on the two $2$\,s plans makes both vehicles decide complementarily: the vehicle that first comes within $3.0$\,m of the other's path goes; within $0.4$\,s, going straight precedes turning, then the vehicle from the right.
If the ego yields, its label stops $0.5$\,m before the partner's stretch; if the partner yields, its future is re-timed, and in half the samples its broadcast plan too.

\noindent\textbf{Risk labels.}
Candidate $a$ has risk target $1$ if at some $t_k$ the ego at $s^a_k$ overlaps a recorded or virtual box in the corridor (the \emph{box test}, \cref{eq:occ}); for a candidate moving above $0.5$\,m/s, pedestrians and cyclists are padded by $0.7$\,m and stretches start $0.4$\,m earlier.

\begin{table*}[t]
  \centering
  \footnotesize
  \setlength{\tabcolsep}{1.6pt}
  \renewcommand{\arraystretch}{0.8}  %
  \begin{tabularx}{\textwidth}{@{}l c *{8}{>{\centering\arraybackslash}X}@{}}
    \toprule
\multirow{2}{*}{Method} & \multirow{2}{*}{V2X} & \multicolumn{2}{c}{V2Xverse} & \multicolumn{2}{c}{InterDrive} & \multicolumn{2}{c}{InterDrive, traffic} & \multicolumn{2}{c}{MDrive} \\
\cmidrule(lr){3-4}\cmidrule(lr){5-6}\cmidrule(lr){7-8}\cmidrule(l){9-10}
& & DS$\uparrow$ & SR$\uparrow$ & DS$\uparrow$ & SR$\uparrow$ & DS$\uparrow$ & SR$\uparrow$ & No coll.$\uparrow$ & DS$\uparrow$ \\
\midrule
\grouprow{10}{Single-vehicle end-to-end}
WOR~\cite{chen2021wor}\venue{ICCV'21} & \xmark & 55.85\,{\fontsize{5.5}{6}\selectfont\pub{12.05}} & 0.27 & 58.16 & 0.28 & 51.26 & 0.24 & 0.78 & 70.43 \\
LAV~\cite{chen2022lav}\venue{CVPR'22} & \xmark & 46.55\,{\fontsize{5.5}{6}\selectfont\pub{38.27}} & 0.18 & 52.07 & 0.30 & 52.61 & 0.30 & 0.66 & 71.93 \\
TransFuser~\cite{chitta2023transfuser}\venue{TPAMI'23} & \xmark & 57.59\,{\fontsize{5.5}{6}\selectfont\pub{30.23}} & 0.28 & 57.76 & 0.50 & 53.05 & 0.37 & 0.71 & 66.31 \\
InterFuser~\cite{shao2022interfuser}\venue{CoRL'22} & \xmark & 45.28\,{\fontsize{5.5}{6}\selectfont\pub{40.31}} & 0.13 & 18.91 & 0.09 & 16.58 & 0.07 & 0.39 & 54.20 \\
TCP~\cite{wu2022tcp}\venue{NeurIPS'22} & \xmark & 53.53\,{\fontsize{5.5}{6}\selectfont\pub{47.48}} & 0.18 & 74.58\,{\fontsize{5.5}{6}\selectfont\pub{73.68}} & 0.59 & 70.91\,{\fontsize{5.5}{6}\selectfont\pub{74.18}} & 0.46 & \second{0.88} & 77.01\,{\fontsize{5.5}{6}\selectfont\pub{76.79}} \\
VAD~\cite{jiang2023vad}\venue{ICCV'23} & \xmark & 11.92 & 0.00 & 28.78\,{\fontsize{5.5}{6}\selectfont\pub{25.37}} & 0.11 & 24.28\,{\fontsize{5.5}{6}\selectfont\pub{25.18}} & 0.09 & 0.49 & 48.57\,{\fontsize{5.5}{6}\selectfont\pub{57.84}} \\
UniAD~\cite{hu2023uniad}\venue{CVPR'23} & \xmark & 16.81 & 0.02 & 33.95\,{\fontsize{5.5}{6}\selectfont\pub{35.17}} & 0.11 & 29.02\,{\fontsize{5.5}{6}\selectfont\pub{37.13}} & 0.11 & 0.48 & 61.27\,{\fontsize{5.5}{6}\selectfont\pub{61.04}} \\
LMDrive~\cite{shao2024lmdrive}\venue{CVPR'24} & \xmark & 29.73 & 0.05 & 53.89\,{\fontsize{5.5}{6}\selectfont\pub{48.83}} & 0.20 & 51.58\,{\fontsize{5.5}{6}\selectfont\pub{49.95}} & 0.15 & 0.79 & 63.01\,{\fontsize{5.5}{6}\selectfont\pub{61.92}} \\
\midrule
\grouprow{10}{Cooperative}
No fusion$^{\ast}$~\cite{liu2025codriving}\venue{TPAMI'25} & \xmark & 56.37\,{\fontsize{5.5}{6}\selectfont\pub{33.81}} & 0.29 & 72.77 & 0.65 & 58.23 & 0.48 & 0.78 & 83.35 \\
Early fusion~\cite{liu2025codriving}\venue{TPAMI'25} & \cmark & 57.20\,{\fontsize{5.5}{6}\selectfont\pub{59.12}} & \second{0.32} & 68.32 & 0.59 & 50.33 & 0.37 & 0.77 & 82.45 \\
Late fusion~\cite{liu2025codriving}\venue{TPAMI'25} & \cmark & \second{59.09}\,{\fontsize{5.5}{6}\selectfont\pub{52.40}} & 0.31 & 70.13 & 0.54 & 63.50 & 0.46 & 0.82 & 84.79 \\
F-Cooper~\cite{chen2019fcooper}\venue{SEC'19} & \cmark & 51.45\,{\fontsize{5.5}{6}\selectfont\pub{44.00}} & 0.27 & 65.59 & 0.52 & 53.72 & 0.28 & 0.79 & 82.36 \\
V2X-ViT~\cite{xu2022v2xvit}\venue{ECCV'22} & \cmark & 53.34\,{\fontsize{5.5}{6}\selectfont\pub{39.35}} & 0.19 & 47.19 & 0.02 & 40.40 & 0.00 & 0.70 & 68.81 \\
\cod~\cite{liu2025codriving}\venue{TPAMI'25} & \cmark & 56.59\,{\fontsize{5.5}{6}\selectfont\pub{77.15}} & 0.30 & 69.59\,{\fontsize{5.5}{6}\selectfont\pub{74.13}} & 0.59 & 60.59\,{\fontsize{5.5}{6}\selectfont\pub{64.50}} & 0.48 & 0.77 & 82.80\,{\fontsize{5.5}{6}\selectfont\pub{82.67}} \\
Defer to Plan$^{\ddagger}$~\cite{li2026defer}\venue{ICME'26} & \cmark & 56.50\,{\fontsize{5.5}{6}\selectfont\pub{79.72}} & 0.31 & 73.10 & 0.59 & 69.87 & 0.48 & 0.84 & \second{85.13} \\
Rule-based$^{\ast\ast}$~\cite{liu2025colmdriver,mdrive2026}\venue{ICCV'25} & \cmark & 40.32 & 0.19 & 76.55\,{\fontsize{5.5}{6}\selectfont\pub{78.38}} & 0.70 & 70.78\,{\fontsize{5.5}{6}\selectfont\pub{69.71}} & 0.57 & 0.81 & 73.23\,{\fontsize{5.5}{6}\selectfont\pub{76.12}} \\
CoLMDriver~\cite{liu2025colmdriver}\venue{ICCV'25} & \cmark & 40.79 & 0.18 & 72.12\,{\fontsize{5.5}{6}\selectfont\pub{88.53}} & 0.63 & 69.18\,{\fontsize{5.5}{6}\selectfont\pub{77.09}} & 0.54 & 0.74 & 69.11\,{\fontsize{5.5}{6}\selectfont\pub{77.16}} \\
CoLMIN~\cite{colmin2026}\venue{TMM'26} & \cmark & 40.15 & 0.18 & \second{90.12} & \second{0.81} & \second{78.93} & \second{0.67} & 0.83$^{\S}$ & 74.09$^{\S}$ \\
\midrule
\gr{Expert (privileged)}~\cite{liu2025codriving,liu2025colmdriver,mdrive2026} & \gr{--} & \gr{62.95} & \gr{0.39} & \gr{83.89} & \gr{0.67} & \gr{78.42} & \gr{0.48} & \gr{0.80} & \gr{67.52} \\
\midrule
\rowcolor{oursbg}
\textbf{\ours (ours)} & \cmark & \sota{70.00} & \sota{0.42} & \sota{97.79} & \sota{0.89} & \sota{90.70} & \sota{0.75} & \sota{0.95} & \sota{88.16} \\
\bottomrule
  \end{tabularx}
  \caption{\textbf{Closed-loop results.} One \ours model (LiDAR only) for all benchmarks. Every row is our rerun of the released code, except CoLMIN on InterDrive (published), $^{\ddagger}$our reimplementation and $^{\S}$our adaptation (MDrive ships no CoLMIN); small \pub{gray}: DS as reported (V2Xverse: on $67$ unreleased routes, \cref{sec:setup}). \gr{Expert}: privileged, our rerun. MDrive: share of vehicles without a collision and DS, over its three buckets. \textbf{Bold}: best (\fire: ours); \underline{underline}: second. $^{\ast}$\cod without messages. $^{\ast\ast}$CoLMDriver negotiating by rules. Seeds and further metrics: \cref{sec:supp_id_md,tab:supp_seeds}.}
  \label{tab:main}
\end{table*}

\vspace{-1mm}\subsection{Training and inference details}\label{sec:training}

The model is the end of a fine-tuning chain whose stages after the first train perception, fusion and planner together.
The stages are the per-agent network ($20$ epochs, batch size $8$), the full model without the map ($6$ epochs, batch size $4$, with the degradation severity raised from $0.3$ to $1.0$ over the first five), $3$ epochs with the route taken from the global plan, and $2$ epochs with the map modules.
These are added through zero-initialised layers and trained on the labels of \cref{sec:supervision} and on degraded samples (sensor, delay, loss, pose).
The loss is
\begin{equation}\label{eq:loss}
\begin{aligned}
  \mathcal{L} ={}& \mathcal{L}_{\mathrm{det}}+\mathcal{L}_{\mathrm{aux}}+0.2\,\mathcal{L}_{\mathrm{rel}}+\mathcal{L}_{\mathrm{cls}}+\mathcal{L}_{\mathrm{reg}}\\
  &+0.5\,\mathcal{L}_{\mathrm{lat}}+0.5\,\mathcal{L}_{\mathrm{risk}}+\mathcal{L}_{\mathrm{nom}} ,
\end{aligned}
\end{equation}
with the detection term and the five planning terms summed over the three fusion or planner layers (all terms: \cref{sec:supp_loss}).

At inference, only these parts are written by hand: the envelope, the gate's geometry and decoding, partner selection, the supersede rule, the route's straight continuation past its end, the rescaling by the configuration's speed factor (\cref{sec:supp_setup}) and the benchmarks' PID controller.
We switch off the controller's forced push (throttle $0.8$ for $2.4$\,s after $20$\,s below $0.2$\,m/s, whatever the plan says).
None of them sets a speed of its own: the envelope takes the lower of two network outputs, at most $v_{\max}$, and the decoding reverses a pairwise decision only to break a cycle ($51$ planning steps in the $46$ tests without traffic).

\vspace{-1.5mm}\section{Experiments}\label{sec:experiments}

\begin{figure*}[t]
  \centering
  \includegraphics[width=\textwidth]{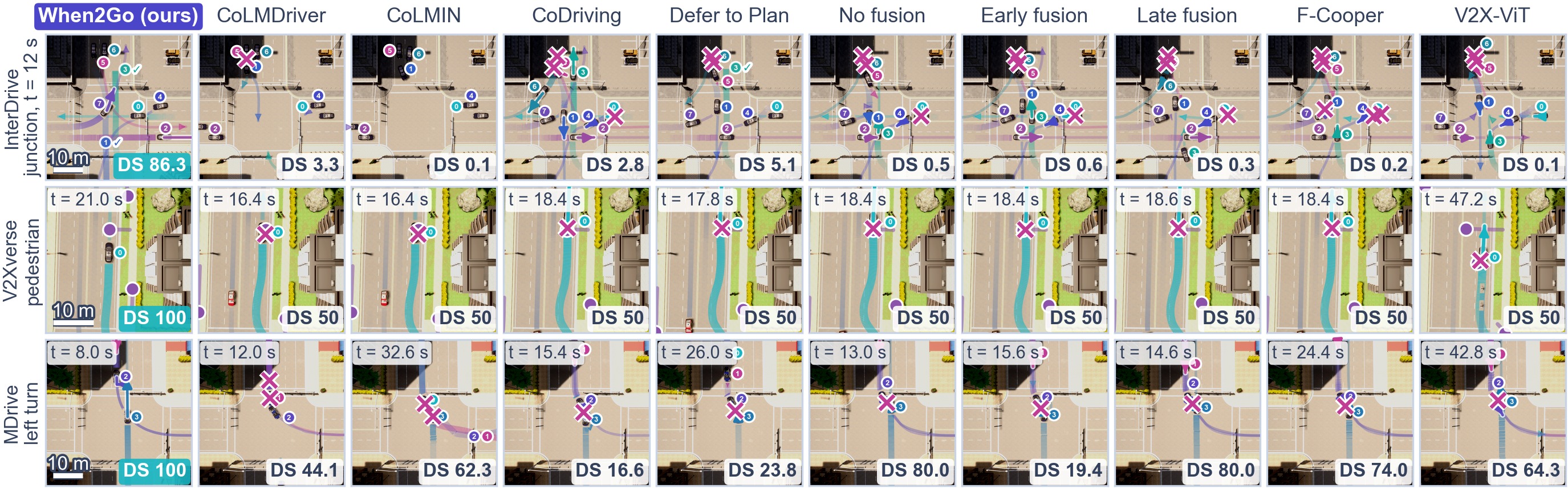}
  \caption{\textbf{Ten methods on the three benchmarks.} One camera per row; on MDrive, each baseline at its first collision. Only \ours never collides (\textcolor[HTML]{C13C95}{$\times$}). Corner: DS; \checkmark: arrived. CoLMIN's runs: \cref{sec:supp_qual}.}
  \label{fig:qual}
\end{figure*}

\begin{table}[t]
  \centering
  \footnotesize
  \setlength{\tabcolsep}{4pt}
  \begin{tabularx}{\linewidth}{@{}>{\raggedright\arraybackslash}X ccc@{}}
    \toprule
    \multirow{2}{*}{Variant} & V2Xverse & \multicolumn{2}{c}{InterDrive DS} \\
    \cmidrule(lr){2-2}\cmidrule(l){3-4}
    & DS & no traffic & traffic \\
    \midrule
    \rowcolor{oursbg}
    \ours, repetition 3 & 72.24 & 97.79 & 90.24 \\
    \grouprow{4}{Route-time conflict map}
    \quad w/o map (tokens only) & \dnb{-6.4} & \dn{-4.6} & \dn{-0.5} \\
    \quad w/o partner plans in the map & \dn{-2.0} & \dn{-6.0} & \dn{-1.6} \\
    \quad w/o forecasts in the map & \dn{-1.3} & \dn{-3.3} & \dn{-2.0} \\
    \quad one class channel & \up{+3.0} & \dn{-0.5} & \dn{-1.0} \\
    \quad BEV occupancy instead of the map & \dn{-5.9} & \dn{-1.8} & \dn{-1.3} \\
    \grouprow{4}{Cooperative fusion and planner}
    \quad w/o reliability estimate & \dnb{-6.4} & \dn{-2.2} & \up{+1.2} \\
    \quad w/o risk envelope (regressed profile) & \dn{-1.0} & \dn{-4.2} & \dn{-1.4} \\
    \quad w/o candidate imitation loss & \dn{-1.9} & \up{+0.4} & \up{+0.3} \\
    \grouprow{4}{Supervision}
    \quad expert plans as recorded & \dnb{-6.4} & \dn{-4.5} & \dn{-2.5} \\
    \quad unexplained stops released only & \up{+0.5} & \dn{-2.2} & \dn{-3.0} \\
    \quad distance keeping only & \dn{-3.9} & \dn{-2.3} & \up{+0.3} \\
    \quad w/o commit or yield early & \dn{-4.2} & \dn{-3.0} & \dn{-4.5} \\
    \quad w/o box test in the risk labels & \dn{-3.6} & \dnb{-10.9} & \dnb{-4.8} \\
    \quad w/o virtual partners & \dn{-0.7} & \dn{-6.1} & \dn{-1.4} \\
    \bottomrule
  \end{tabularx}
  \caption{\textbf{Ablations.} Each row changes one component and is retrained (envelope row: inference only), on the seeds of the first line (repetition $3$). Values: change in driving score, paired over the same runs (bold: largest drop). One repetition resolves about $7$ points on V2Xverse; intervals: \cref{tab:supp_ablation_ci}.}
  \label{tab:ablation}
\end{table}

\begin{table}[t]
  \centering
  \footnotesize
  \setlength{\tabcolsep}{3.5pt}
  \begin{tabularx}{\linewidth}{@{}>{\raggedright\arraybackslash}X ccc ccc@{}}
    \toprule
    Order & Learned & Decides & Acyclic & DS$\uparrow$ & RC$\uparrow$ & SR$\uparrow$ \\
    \midrule
    None & \xmark & \xmark & \xmark & 93.64 & 94.63 & 0.83 \\
    Map channel & \cmark & \xmark & \xmark & 92.94 & 95.23 & 0.80 \\
    Pairwise & \cmark & \cmark & \xmark & 96.53 & 97.74 & \textbf{0.89} \\
    \rowcolor{oursbg}
    Gate (\ours) & \cmark & \cmark & \cmark & \textbf{97.79} & \textbf{98.99} & \textbf{0.89} \\
    \midrule
    \gr{Priority rule} & \multicolumn{3}{c}{\gr{hand-written}} & \gr{97.43} & \gr{98.45} & \gr{0.89} \\
    \bottomrule
  \end{tabularx}
  \caption{\textbf{Learning the order.} InterDrive without traffic, every vehicle controlled, same planner weights except Map channel, which is retrained ($46$ tests). Learned: precedence from the network; Decides: it picks each pair's plans (else only a map channel); Acyclic: decoded into an order without cycles. \gr{Gray}: a hand-written rule.}
  \label{tab:order}
\end{table}

\vspace{-1.5mm}\subsection{Setup}\label{sec:setup}

\noindent\textbf{Benchmarks.}
We evaluate in closed loop on three CARLA~\cite{dosovitskiy2017carla} benchmarks.
\emph{V2Xverse}~\cite{liu2025codriving}: the $85$ runs of the public release in Town05, one controlled vehicle with one roadside unit among background vehicles, pedestrians and cyclists.
\emph{InterDrive}~\cite{liu2025colmdriver}: $46$ tests in which two to eight controlled vehicles meet at intersections, in merges and in lane changes, without and with background traffic.
\emph{MDrive}~\cite{mdrive2026}: $225$ scenarios with one to eight controlled vehicles in three buckets, one of them replayed from real-world logs.
We report the driving score (DS: route completion times the product of the infraction penalties), averaged as each benchmark's script does, and the success rate (SR: the share of runs, tests or, on MDrive, vehicles with a perfect score); details in \cref{sec:supp_setup}.

\noindent\textbf{Protocol.}
Repetition $r$ seeds the simulator with $2000{+}r$.
Design choices used repetitions $0$ to $2$ of the same routes; we report unseen repetitions: $3$ to $8$ on V2Xverse ($510$ runs) and $3$ to $5$ on InterDrive with traffic and on MDrive. InterDrive without traffic is deterministic, so repetition $3$ repeats the development episodes; MDrive's InterDrive bucket uses its routes.
On V2Xverse the published scores were obtained on $67$ routes that are not in the public release~\cite{liu2025codriving,li2026defer}, and users report lower scores for the released checkpoints~\cite{v2xverse_issue44,v2xverse_issue48}.
We therefore rerun every released method on the public protocol with its released checkpoint, agent code, controller and configuration (interface fixes only, \cref{sec:supp_setup}); ours runs unrendered (same LiDAR).
On InterDrive and MDrive, the released cooperative baselines fuse only their own roadside unit (CoLMDriver and CoLMIN also exchange intents); \ours fuses up to three sources and reads up to three more plans within $70$\,m (\cref{sec:supp_gate}); \cref{sec:results} also runs it with the baselines' inputs.

\noindent\textbf{Implementation.}
One model ($18.0$\,M parameters, LiDAR only) is trained on the released V2Xverse recordings of Town01 to 04 and Town06 ($31{,}755$ samples, \cref{sec:training}).
Town05, the town of all V2Xverse runs, is held out, and InterDrive and MDrive also test in training towns (\cref{tab:supp_scale} splits InterDrive by town). The precedence network of its order gate ($0.27$\,M parameters) is trained separately on rule-labelled counterfactual pairs (\cref{sec:agreement}).
One forward pass takes about $0.1$\,s per vehicle on one GPU.

\vspace{-1.5mm}\subsection{Main results}\label{sec:results}

\Cref{tab:main} compares one \ours model with published and rerun methods (\cref{fig:qual}).

\noindent\textbf{InterDrive.}
\ours reaches $97.79$ DS without and $90.70$ with background traffic, $7.7$ and $11.8$ points above the best published method, CoLMIN~\cite{colmin2026}. Our rerun of CoLMDriver scores below its reported values, while the same harness reproduces those of \cod ($74.46$ against $74.13$ under the benchmark's earlier result script) and of the rule variant ($78.38$); the difference lies in its language-model negotiation, whose sampling the release does not fix (\cref{sec:supp_v2xverse}).
\begin{table}[t]
  \centering
  \footnotesize
  \setlength{\tabcolsep}{2pt}
  \begin{tabularx}{\linewidth}{@{}l *{6}{>{\centering\arraybackslash}X}@{}}
    \toprule
    & \multicolumn{3}{c}{no traffic} & \multicolumn{3}{c}{traffic} \\
    \cmidrule(lr){2-4}\cmidrule(l){5-7}
    Method & IC & LM & LC & IC & LM & LC \\
    \midrule
    CoLMIN~\cite{colmin2026}              & \second{90.71} & \second{98.75} & \second{67.20} & 66.47 & \second{94.11} & \second{67.45} \\
    No fusion$^{\ast}$                    & 61.80 & 96.80 & 37.39 & 40.56 & 83.14 & 35.73 \\
    Early fusion                          & 60.07 & 93.60 & 23.71 & 45.61 & 69.81 & 12.22 \\
    Late fusion                           & 51.81 & 96.17 & 46.24 & 48.14 & 81.84 & 52.19 \\
    F-Cooper                              & 70.30 & 75.79 & 29.48 & 46.55 & 66.95 & 36.75 \\
    V2X-ViT                               & 41.61 & 69.05 & 5.10 & 37.47 & 58.44 & 1.87 \\
    \cod                                  & 54.95 & 96.49 & 35.31 & 47.87 & 82.40 & 34.69 \\
    Defer to Plan$^{\ddagger}$            & 54.14 & 94.49 & 62.28 & 54.99 & 86.65 & 61.40 \\
    Rule-based$^{\ast\ast}$             & 74.18 & 96.80 & 31.28 & \second{68.07} & 87.72 & 34.55 \\
    CoLMDriver~\cite{liu2025colmdriver}   & 65.32 & 91.36 & 39.32 & 61.27 & 90.52 & 33.63 \\
    \midrule
    \rowcolor{oursbg}
    \textbf{\ours}                        & \textbf{95.12} & \textbf{100.0} & \textbf{98.25} & \textbf{80.30} & \textbf{97.59} & \textbf{96.84} \\
    \bottomrule
  \end{tabularx}
  \caption{\textbf{InterDrive by scenario type.} Driving score at intersections (IC, $18$ tests), merges (LM, $20$) and lane changes (LC, $8$). CoLMIN as published, the other baselines our reruns; CoLMDriver's tables split the tests differently.}
  \label{tab:categories}
\vspace{-1mm}
\end{table}

\begin{figure}[t]
  \centering
  \includegraphics[width=\columnwidth]{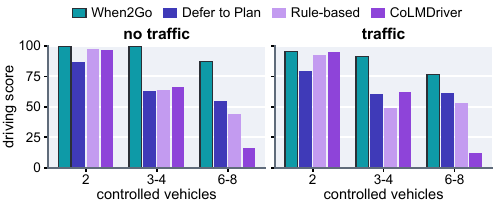}
  \vspace{-5mm}
  \caption{\textbf{InterDrive by number of controlled vehicles} ($22$, $16$ and $8$ tests).}
  \label{fig:scale}
  \vspace{-3mm}
\end{figure}

The largest margin is in lane changes ($98.25$ and $96.84$ against $67.20$ and $67.45$, \cref{tab:categories}), where a vehicle must enter a gap that another controlled vehicle is leaving; there the order comes from the broadcast intended trajectories instead of a negotiation in language~\cite{liu2025colmdriver,colmin2026}.
Most of the margin comes from fewer collisions.
Infractions (RC${-}$DS) cost our reruns of \cod and its fusion variants $24$ to $31$ points without traffic and $26$ to $41$ with it, almost all through collisions between controlled vehicles; they cost CoLMIN $4.2$ and $13.9$, and \ours $1.2$ and $0.8$ ($0$ and $1$ such collisions per $100$ tests).
The margin grows with the number of controlled vehicles (\cref{fig:scale}): negotiation comes close with two and falls behind with more, where \ours scores $100.0$ and $87.3$ without traffic.
Most of it does not come from more inputs: given only the baselines' perception, its own roadside unit, \ours scores $94.1$ and $85.7$, and with the other vehicles' broadcast plans added, $96.1$ and $90.6$ (\cref{sec:supp_id_md}).

\noindent\textbf{MDrive.}
\ours keeps $95.0\,\%$ of its vehicles free of collisions, against $87.7\,\%$ for the safest baseline (TCP; $+7.3$, $[+4.0, +10.8]$).
It also has the highest driving score, $88.16$, against $85.13$ for Defer to Plan ($+3.0$, $[+0.03, +6.22]$) and $82.67$ published for \cod~\cite{mdrive2026}. The CoLMDriver family (CoLMDriver, its rule variant, our CoLMIN adaptation) scores $69.1$ to $74.1$, $13.9$ to $18.9$ below \ours on the same scenarios (\cref{sec:supp_setup}).
The margin comes from the buckets where controlled vehicles interact: \ours leads the best baseline of the Interaction bucket, F-Cooper, by $3.3$ ($[+0.3, +6.7]$), and Defer to Plan on the InterDrive development routes by $5.4$ ($[-0.8, +12.4]$).
On the V2X-PnP log replays, where messages do not help the baselines, \ours trails six cooperative reruns by $1.5$ to $4.9$ and InterFuser by $0.2$, mostly because it waits behind obstacles that need a lane change (\cref{tab:supp_mdrive}).

\noindent\textbf{V2Xverse.}
On the public runs, \ours reaches $70.00$ DS over six repetitions ($510$ runs), $10.9$ points above the strongest rerun baseline (late fusion, $59.09$); even its lowest repetition ($64.5$) is above every rerun baseline.
Rerun on the seeds of \ours, each of twelve rerun baselines scores $12.8$ to $24.4$ points lower, every paired $95\,\%$ interval above $+8$ (\cod: $47.22$, $[+18.5, +27.2]$; \cref{sec:supp_v2xverse}).
\ours finishes $42\,\%$ of the runs with a perfect score, against at most $32\,\%$ for the rerun baselines, and collides with a vehicle in $6$ of $100$ runs, against $9$ to $69$.
In exchange it waits more: $29$ of $100$ runs end blocked, against $10$ to $18$ for the rerun fusion baselines (\cref{tab:supp_v2xverse}).

\noindent\textbf{Against the privileged expert.}
Each benchmark's rule-based expert sees ground truth, yet without the actor deletion it uses to collect data it scores below \ours on all three benchmarks (\cref{tab:main}): its vehicles decide alone, from straight-line predictions of the others ($73.43$ at intersections and $67.15$ in lane changes without traffic, against $95.12$ and $98.25$), and its binary stop rule both blocks and collides (\cref{sec:supp_id_md}).

\vspace{-1.5mm}\subsection{Ablations}\label{sec:ablations}

\noindent\textbf{Control at test time.}
\begin{table}[t]
  \centering
  \footnotesize
  \setlength{\tabcolsep}{4pt}
  \begin{tabularx}{\linewidth}{@{}l *{4}{>{\centering\arraybackslash}X}@{}}
    \toprule
    \multirow{2}{*}{Control} & V2Xverse & \multicolumn{2}{c}{InterDrive} & MDrive \\
    \cmidrule(lr){2-2}\cmidrule(lr){3-4}\cmidrule(l){5-5}
    & DS & no traffic & traffic & Avg \\
    \midrule
    \grouprow{5}{Steering target}
    Benchmark route point & 65.40 & 97.81 & 89.94 & 88.43 \\
    \rowcolor{oursbg}
    Own plan (\ours) & 70.00 & 97.79 & 90.70 & 88.16 \\
    \grouprow{5}{Benchmark's forced push switched on}
    $\Delta$DS, paired & \dn{-4.5} & \up{+0.8} & \dn{-3.7} & \up{+1.0} \\
    Collisions after a push & 21 of 26 & 0 of 0 & 13 of 17 & 68 of 89 \\
    \bottomrule
  \end{tabularx}
  \caption{\textbf{Control at test time.} Top: the same weights steered towards the controller's route point or the network's waypoints (repetitions of \cref{tab:main}; MDrive: repetition $3$). Bottom: with the forced push on (repetition $3$): change in DS and vehicle collisions within $5$\,s after a push.}
  \label{tab:control}
\vspace{-1mm}
\end{table}

Steering by the network's own waypoints instead of the route point gains $4.6$ points on V2Xverse ($[+1.8, +7.6]$, paired) and is within noise elsewhere, although the success rate on MDrive is $3.2$ points lower ($77.1$ against $80.3$).
The controller's forced push (\cref{sec:training}) makes a waiting vehicle go, which is exactly the decision \ours learns; our runs therefore switch it off, while every baseline keeps it, as released.
Switched on, it trades blocked runs for collisions (\cref{tab:control}): $102$ of the $132$ vehicle collisions of \ours then follow a push within $5$\,s, while the driving score changes within noise.
With both default settings (route-point steering, push on), \ours scores $59.77$ instead of $68.31$ on V2Xverse (repetitions $3$ to $5$), still above every baseline on the same seeds and resolved for all but TransFuser ($+3.3$, $[-1.9, +8.5]$).
On MDrive and InterDrive with traffic it scores $89.49$ and $90.21$, within noise, but collides more, mostly after a push (\cref{sec:supp_stock}).

\noindent\textbf{Route-time conflict map and supervision.}
\Cref{tab:ablation} changes one component at a time.
Since another training seed alone costs up to $2.9$ points without traffic and $6.8$ on V2Xverse (\cref{tab:supp_seeds}), effects of this size are not resolved.
Without the map, the score drops by $6.4$ on V2Xverse (paired $95\,\%$ interval $[-12.9, -0.1]$) and by $4.6$ without traffic ($[-10.6, +0.3]$; $1.3$ over three training seeds, \cref{sec:supp_ablation_ci}).
Without traffic, removing the partner plans from the map costs $6.0$ ($[-11.8, -1.5]$) and removing the forecasts $3.3$ ($[-6.2, -0.8]$).
Drawing the same inputs as bird's-eye-view occupancy over future time, the planner interface of cooperative planners evaluated open loop~\cite{yu2025univ2x,riskmm2026,you2026v2xwam}, costs $5.9$ points on V2Xverse, with more blocked runs, and $1.8$ and $1.3$ on InterDrive (within noise, \cref{tab:supp_ablation_ci}).
On InterDrive, the risk head matters most: risk labels without the box test of \cref{sec:supervision} cost $10.9$ points without traffic ($[-17.6, -5.0]$) and $4.8$ with it.
Without traffic, executing the regressed profile without the risk envelope costs $4.2$, with more collisions, and removing the reliability estimate costs $2.2$.
Each simpler relabelling loses $2.2$ to $3.0$ points without traffic (within training noise), training on the expert's plans as recorded loses $4.5$ ($[-8.4, -1.2]$), and training without virtual partners $6.1$ ($[-11.2, -1.9]$), since the recordings hold almost no conflicts between connected vehicles.

\noindent\textbf{Learning the order.}
\Cref{tab:order} compares ways to decide which of two controlled vehicles goes first.
Without an order, the vehicles resolve each meeting only through their plans in the conflict map, and $17\,\%$ of the tests end with a penalty.
The order gate raises the driving score by $4.1$ points (paired $95\,\%$ interval $[+1.3, +7.7]$) and the success rate from $0.83$ to $0.89$.
Acyclic decoding adds $1.3$ points over using the pairwise decisions directly ($[-0.1, +3.9]$), and the learned order matches a hand-written right-of-way rule on the same broadcasts ($+0.4$, $[-0.01, +1.05]$).

\vspace{-1.5mm}\subsection{Imperfect communication}\label{sec:robust}

\begin{table}[t]
  \centering
  \footnotesize
  \setlength{\tabcolsep}{4pt}
  \begin{tabularx}{\linewidth}{@{}l *{4}{>{\centering\arraybackslash}X}@{}}
    \toprule
    Channel & DS$\uparrow$ & RC$\uparrow$ & SR$\uparrow$ & $\Delta$DS \\
    \midrule
    \rowcolor{oursbg}
    Ideal & 97.79 & 98.99 & 0.89 & -- \\
    \grouprow{5}{Latency}
    \quad $200$\,ms & 94.63 & 97.77 & 0.83 & $-3.2$ \\
    \quad $400$\,ms & 91.35 & 96.11 & 0.78 & $-6.4$ \\
    \quad $800$\,ms & 90.22 & 93.44 & 0.76 & $-7.6$ \\
    \grouprow{5}{Message loss}
    \quad $50\,\%$, independent & 94.53 & 97.44 & 0.87 & $-3.3$ \\
    \quad $50\,\%$, in bursts of $1$\,s & 92.99 & 98.27 & 0.83 & $-4.8$ \\
    \quad $100\,\%$ (no messages) & 87.06 & 96.93 & 0.72 & $-10.7$ \\
    \grouprow{5}{Pose error (standard deviation)}
    \quad $0.5$\,m, $0.5^\circ$ & 98.26 & 99.45 & 0.93 & $+0.5$ \\
    \quad $1.0$\,m, $1.0^\circ$ & 94.99 & 97.61 & 0.85 & $-2.8$ \\
    \grouprow{5}{Combined}
    \quad $200$\,ms, $30\,\%$ loss, $0.3$\,m, $0.5^\circ$ & 92.51 & 96.66 & 0.83 & $-5.3$ \\
    \midrule
    \gr{CoLMIN~\cite{colmin2026}, ideal channel} & \gr{90.12} & \gr{94.36} & \gr{0.81} & \gr{--} \\
    \bottomrule
  \end{tabularx}
  \caption{\textbf{Imperfect channels.} InterDrive without traffic, one repetition; the degradation acts on every message. $\Delta$DS: paired difference to the ideal channel. \gr{Gray}: the best published method, with an ideal channel.}
  \label{tab:robust}
\vspace{-1mm}
\end{table}

We degrade the channel on InterDrive without traffic (\cref{tab:robust}): latency, independent and bursty message loss, and errors in the senders' poses, whose standard deviation the receiver is given.
Of the partial degradations, latency costs most, $7.6$ points at $800$\,ms; losing half of the messages costs $3.3$ points, or $4.8$ when the losses come in bursts.
Pose errors of $0.5$\,m and $0.5^\circ$ have no effect that one repetition resolves; $1$\,m and $1^\circ$ cost $2.8$ points.
With an ideal channel, two vehicles that order each other never disagree ($0$ of $42{,}789$ pair-steps); they disagree in $5.7\,\%$ of the pair-steps at $800$\,ms of latency and in $5.6\,\%$ with $1$\,m pose errors (\cref{tab:supp_coord}).
If both vehicles decide on the same two delayed messages (common-knowledge decoding, \cref{sec:supp_gate}; not used in the reported runs), the score stays within noise of the ideal channel up to $800$\,ms ($95.96$; $92.35$ without it in the same implementation), and disagreements almost vanish.
Without any message, each vehicle drives on its own LiDAR and the score falls by $10.7$ points to $87.06$; lane changes fall from $98.25$ to $60.92$.
The planner alone is thus already strong: it scores above every cooperative method of \cref{tab:main} except the published values of the two language-model negotiators, and without messages it also stays above every rerun baseline on V2Xverse and MDrive ($64.59$ and $86.71$, \cref{sec:supp_gate}).
Cooperation adds the rest: messages without an order raise the score to $93.64$, and the order to $97.79$ (\cref{tab:order}).

\vspace{-2mm}\section{Conclusion}\label{sec:conclusion}

\ours decides when each vehicle goes by reading a route-time conflict map, and its vehicles reach the same order from the same messages; one end-to-end network leads three closed-loop benchmarks.
\noindent\textbf{Limitations.}
The planner follows the given route without lane changes of its own and has no traffic-light or stop-sign input; the network senses with LiDAR only; its labels and those of the order gate come from privileged simulator state and rules; the order gate chooses between two fixed plans; pairs stall in $24$ of $138$ runs with traffic; and all results are in simulation.

\section*{Acknowledgments}
We thank the Georgia Advanced Computing Resource Center (GACRC) for computing resources and the TRAVELS project for its support.

{
    \small
    \bibliographystyle{ieeenat_fullname}
    \bibliography{refs}

\begin{thebibliography}{92}
\providecommand{\natexlab}[1]{#1}
\providecommand{\url}[1]{\texttt{#1}}
\expandafter\ifx\csname urlstyle\endcsname\relax
  \providecommand{\doi}[1]{doi: #1}\else
  \providecommand{\doi}{doi: \begingroup \urlstyle{rm}\Url}\fi

\bibitem[Azimi et~al.(2014)Azimi, Bhatia, Rajkumar, and
  Mudalige]{azimi2014stip}
Reza Azimi, Gaurav Bhatia, Ragunathan~(Raj) Rajkumar, and Priyantha Mudalige.
\newblock {STIP}: Spatio-temporal intersection protocols for autonomous
  vehicles.
\newblock In \emph{ACM/IEEE International Conference on Cyber-Physical Systems
  (ICCPS)}, 2014.

\bibitem[Bansal et~al.(2019)Bansal, Krizhevsky, and
  Ogale]{bansal2019chauffeurnet}
Mayank Bansal, Alex Krizhevsky, and Abhijit Ogale.
\newblock {ChauffeurNet}: Learning to drive by imitating the best and
  synthesizing the worst.
\newblock In \emph{Proceedings of Robotics: Science and Systems (RSS)}, 2019.

\bibitem[Biswas et~al.(2024)Biswas, Casas, Sykora, Agro, Sadat, and
  Urtasun]{biswas2024quad}
Sourav Biswas, Sergio Casas, Quinlan Sykora, Ben Agro, Abbas Sadat, and Raquel
  Urtasun.
\newblock {QuAD}: Query-based interpretable neural motion planning for
  autonomous driving.
\newblock In \emph{IEEE International Conference on Robotics and Automation
  (ICRA)}, 2024.

\bibitem[Burges et~al.(2005)Burges, Shaked, Renshaw, Lazier, Deeds, Hamilton,
  and Hullender]{burges2005ranknet}
Chris Burges, Tal Shaked, Erin Renshaw, Ari Lazier, Matt Deeds, Nicole
  Hamilton, and Greg Hullender.
\newblock Learning to rank using gradient descent.
\newblock In \emph{International Conference on Machine Learning (ICML)}, 2005.

\bibitem[{\v{C}}{\'a}p et~al.(2015){\v{C}}{\'a}p, Nov{\'a}k, Kleiner, and
  Seleck{\'y}]{cap2015prioritized}
Michal {\v{C}}{\'a}p, Peter Nov{\'a}k, Alexander Kleiner, and Martin
  Seleck{\'y}.
\newblock Prioritized planning algorithms for trajectory coordination of
  multiple mobile robots.
\newblock \emph{IEEE Transactions on Automation Science and Engineering},
  12\penalty0 (3):\penalty0 835--849, 2015.

\bibitem[Carion et~al.(2020)Carion, Massa, Synnaeve, Usunier, Kirillov, and
  Zagoruyko]{carion2020detr}
Nicolas Carion, Francisco Massa, Gabriel Synnaeve, Nicolas Usunier, Alexander
  Kirillov, and Sergey Zagoruyko.
\newblock End-to-end object detection with transformers.
\newblock In \emph{Proceedings of the European Conference on Computer Vision
  (ECCV)}, 2020.

\bibitem[Casas et~al.(2021)Casas, Sadat, and Urtasun]{casas2021mp3}
Sergio Casas, Abbas Sadat, and Raquel Urtasun.
\newblock {MP3}: A unified model to map, perceive, predict and plan.
\newblock In \emph{Proceedings of the IEEE/CVF Conference on Computer Vision
  and Pattern Recognition (CVPR)}, 2021.

\bibitem[Chen and Kr{\"a}henb{\"u}hl(2022)]{chen2022lav}
Dian Chen and Philipp Kr{\"a}henb{\"u}hl.
\newblock Learning from all vehicles.
\newblock In \emph{Proceedings of the IEEE/CVF Conference on Computer Vision
  and Pattern Recognition (CVPR)}, pages 17201--17210, 2022.

\bibitem[Chen et~al.(2021)Chen, Koltun, and Kr{\"a}henb{\"u}hl]{chen2021wor}
Dian Chen, Vladlen Koltun, and Philipp Kr{\"a}henb{\"u}hl.
\newblock Learning to drive from a world on rails.
\newblock In \emph{Proceedings of the IEEE/CVF International Conference on
  Computer Vision (ICCV)}, pages 15570--15579, 2021.

\bibitem[Chen et~al.(2019{\natexlab{a}})Chen, Ma, Tang, Guo, Yang, and
  Fu]{chen2019fcooper}
Qi Chen, Xu Ma, Sihai Tang, Jingda Guo, Qing Yang, and Song Fu.
\newblock {F-Cooper}: Feature based cooperative perception for autonomous
  vehicle edge computing system using {3D} point clouds.
\newblock In \emph{Proceedings of the 4th ACM/IEEE Symposium on Edge Computing
  (SEC)}, 2019{\natexlab{a}}.

\bibitem[Chen et~al.(2019{\natexlab{b}})Chen, Tang, Yang, and
  Fu]{chen2019cooper}
Qi Chen, Sihai Tang, Qing Yang, and Song Fu.
\newblock {Cooper}: Cooperative perception for connected autonomous vehicles
  based on {3D} point clouds.
\newblock In \emph{Proceedings of the IEEE 39th International Conference on
  Distributed Computing Systems (ICDCS)}, 2019{\natexlab{b}}.

\bibitem[Chitta et~al.(2023)Chitta, Prakash, Jaeger, Yu, Renz, and
  Geiger]{chitta2023transfuser}
Kashyap Chitta, Aditya Prakash, Bernhard Jaeger, Zehao Yu, Katrin Renz, and
  Andreas Geiger.
\newblock {TransFuser}: Imitation with transformer-based sensor fusion for
  autonomous driving.
\newblock \emph{IEEE Transactions on Pattern Analysis and Machine
  Intelligence}, 45\penalty0 (11), 2023.

\bibitem[Chiu and Smith(2026)]{chiu2026cmudrive}
Hsu-kuang Chiu and Stephen~F. Smith.
\newblock {CMU-Drive} and {V2V-VLA}: Cooperative multi-agent unified driving
  with reasoning benchmark and vehicle-to-vehicle vision-language-action
  models.
\newblock \emph{arXiv preprint arXiv:2608.07621}, 2026.

\bibitem[Chiu et~al.(2026)Chiu, Hachiuma, Wang, Wang, Chen, and
  Smith]{chiu2026v2vgot}
Hsu-kuang Chiu, Ryo Hachiuma, Chien-Yi Wang, Yu-Chiang~Frank Wang, Min-Hung
  Chen, and Stephen~F. Smith.
\newblock {V2V-GoT}: Vehicle-to-vehicle cooperative autonomous driving with
  multimodal large language models and graph-of-thoughts.
\newblock In \emph{IEEE International Conference on Robotics and Automation
  (ICRA)}, 2026.

\bibitem[Cho et~al.(2026)Cho, Kang, Lee, Yang, Park, and Yoon]{cosv2x2026}
Hoonhee Cho, Jae-Young Kang, Giwon Lee, Hyemin Yang, Heejun Park, and Kuk-Jin
  Yoon.
\newblock {VIPS}: Vehicle-infrastructure cooperative planning benchmark via
  pseudo-simulation.
\newblock In \emph{Proceedings of the European Conference on Computer Vision
  (ECCV)}, 2026.

\bibitem[Codevilla et~al.(2019)Codevilla, Santana, L{\'o}pez, and
  Gaidon]{codevilla2019limitations}
Felipe Codevilla, Eder Santana, Antonio~M. L{\'o}pez, and Adrien Gaidon.
\newblock Exploring the limitations of behavior cloning for autonomous driving.
\newblock In \emph{Proceedings of the IEEE/CVF International Conference on
  Computer Vision (ICCV)}, 2019.

\bibitem[Coscoy et~al.(2026)Coscoy, Zhou, Zhao, Wei, Magtoto, Liu, Song,
  Zimmer, Huang, Tang, Zhou, and Ma]{mdrive2026}
Marco Coscoy, Zewei Zhou, Seth~Z. Zhao, Henry Wei, Angela Magtoto, Johnson Liu,
  Rui Song, Walter Zimmer, Zhiyu Huang, Chen Tang, Bolei Zhou, and Jiaqi Ma.
\newblock {MDrive}: Benchmarking closed-loop cooperative driving for end-to-end
  multi-agent systems.
\newblock \emph{arXiv preprint arXiv:2605.10904}, 2026.

\bibitem[Cui et~al.(2022)Cui, Qiu, Chen, Stone, and Zhu]{cui2022coopernaut}
Jiaxun Cui, Hang Qiu, Dian Chen, Peter Stone, and Yuke Zhu.
\newblock {Coopernaut}: End-to-end driving with cooperative perception for
  networked vehicles.
\newblock In \emph{Proceedings of the IEEE/CVF Conference on Computer Vision
  and Pattern Recognition (CVPR)}, 2022.

\bibitem[Dauner et~al.(2023)Dauner, Hallgarten, Geiger, and
  Chitta]{dauner2023parting}
Daniel Dauner, Marcel Hallgarten, Andreas Geiger, and Kashyap Chitta.
\newblock Parting with misconceptions about learning-based vehicle motion
  planning.
\newblock In \emph{Proceedings of the Conference on Robot Learning (CoRL)},
  2023.

\bibitem[de~Haan et~al.(2019)de~Haan, Jayaraman, and Levine]{dehaan2019causal}
Pim de Haan, Dinesh Jayaraman, and Sergey Levine.
\newblock Causal confusion in imitation learning.
\newblock In \emph{Advances in Neural Information Processing Systems
  (NeurIPS)}, 2019.

\bibitem[de~Witt et~al.(2019)de~Witt, Foerster, Farquhar, Torr, Boehmer, and
  Whiteson]{dewitt2019mackrl}
Christian A.~Schroeder de Witt, Jakob~N. Foerster, Gregory Farquhar, Philip
  H.~S. Torr, Wendelin Boehmer, and Shimon Whiteson.
\newblock Multi-agent common knowledge reinforcement learning.
\newblock In \emph{Advances in Neural Information Processing Systems
  (NeurIPS)}, 2019.

\bibitem[Dosovitskiy et~al.(2017)Dosovitskiy, Ros, Codevilla, L{\'o}pez, and
  Koltun]{dosovitskiy2017carla}
Alexey Dosovitskiy, German Ros, Felipe Codevilla, Antonio L{\'o}pez, and
  Vladlen Koltun.
\newblock {CARLA}: An open urban driving simulator.
\newblock In \emph{Proceedings of the Conference on Robot Learning (CoRL)},
  2017.

\bibitem[Dresner and Stone(2008)]{dresner2008multiagent}
Kurt Dresner and Peter Stone.
\newblock A multiagent approach to autonomous intersection management.
\newblock \emph{Journal of Artificial Intelligence Research}, 31:\penalty0
  591--656, 2008.

\bibitem[Erdmann and Lozano-P{\'e}rez(1987)]{erdmann1987multiple}
Michael Erdmann and Tom{\'a}s Lozano-P{\'e}rez.
\newblock On multiple moving objects.
\newblock \emph{Algorithmica}, 2:\penalty0 477--521, 1987.

\bibitem[Fan et~al.(2018)Fan, Zhu, Liu, Zhang, Zhuang, Li, Zhu, Hu, Li, and
  Kong]{fan2018baidu}
Haoyang Fan, Fan Zhu, Changchun Liu, Liangliang Zhang, Li Zhuang, Dong Li,
  Weicheng Zhu, Jiangtao Hu, Hongye Li, and Qi Kong.
\newblock Baidu apollo {EM} motion planner.
\newblock \emph{arXiv preprint arXiv:1807.08048}, 2018.

\bibitem[Fang et~al.(2025)Fang, Liu, Ding, Cui, Lv, Hang, and
  Sun]{fang2025codrivingllm}
Shiyu Fang, Jiaqi Liu, Mingyu Ding, Yiming Cui, Chen Lv, Peng Hang, and Jian
  Sun.
\newblock Towards interactive and learnable cooperative driving automation: A
  large language model-driven decision-making framework.
\newblock \emph{IEEE Transactions on Vehicular Technology}, 2025.

\bibitem[Gao et~al.(2025)Gao, Wu, Wang, Liu, Zhou, and Tu]{gao2025langcoop}
Xiangbo Gao, Yuheng Wu, Rujia Wang, Chenxi Liu, Yang Zhou, and Zhengzhong Tu.
\newblock {LangCoop}: Collaborative driving with language.
\newblock In \emph{Proceedings of the IEEE/CVF Conference on Computer Vision
  and Pattern Recognition Workshops (CVPRW)}, 2025.

\bibitem[{GitHub user ikerrrrrrrrrrr}(2025)]{v2xverse_issue44}
{GitHub user ikerrrrrrrrrrr}.
\newblock {Evaluating Problems}.
\newblock \url{https://github.com/CollaborativePerception/V2Xverse/issues/44},
  2025.
\newblock GitHub issue \#44, opened August 27, 2025.

\bibitem[{GitHub user Kiki2049}(2025)]{colmdriver_issue6}
{GitHub user Kiki2049}.
\newblock {Question about Driving performance in InterDrive Benchmark}.
\newblock \url{https://github.com/cxliu0314/CoLMDriver/issues/6}, 2025.
\newblock GitHub issue \#6 of CoLMDriver, opened October 9, 2025.

\bibitem[{GitHub user Rookie-From-East}(2025)]{v2xverse_issue48}
{GitHub user Rookie-From-East}.
\newblock {Questions on the calculation of assessment indicators}.
\newblock \url{https://github.com/CollaborativePerception/V2Xverse/issues/48},
  2025.
\newblock GitHub issue \#48, opened December 24, 2025.

\bibitem[Hu et~al.(2022)Hu, Fang, Lei, Zhong, and Chen]{hu2022where2comm}
Yue Hu, Shaoheng Fang, Zixing Lei, Yiqi Zhong, and Siheng Chen.
\newblock {Where2comm}: Communication-efficient collaborative perception via
  spatial confidence maps.
\newblock In \emph{Advances in Neural Information Processing Systems}, 2022.

\bibitem[Hu et~al.(2023)Hu, Yang, Chen, Li, Sima, Zhu, Chai, Du, Lin, Wang, Lu,
  Jia, Liu, Dai, Qiao, and Li]{hu2023uniad}
Yihan Hu, Jiazhi Yang, Li Chen, Keyu Li, Chonghao Sima, Xizhou Zhu, Siqi Chai,
  Senyao Du, Tianwei Lin, Wenhai Wang, Lewei Lu, Xiaosong Jia, Qiang Liu,
  Jifeng Dai, Yu Qiao, and Hongyang Li.
\newblock Planning-oriented autonomous driving.
\newblock In \emph{Proceedings of the IEEE/CVF Conference on Computer Vision
  and Pattern Recognition (CVPR)}, 2023.

\bibitem[Huang et~al.(2025)Huang, Zhu, Li, Zhao, and Zhang]{select2drive2025}
Jiahao Huang, Jianhang Zhu, Rongpeng Li, Zhifeng Zhao, and Honggang Zhang.
\newblock {Select2Drive}: Pragmatic communications for real-time collaborative
  autonomous driving.
\newblock \emph{IEEE Transactions on Intelligent Transportation Systems},
  26\penalty0 (12), 2025.

\bibitem[Huang et~al.(2026)Huang, Fan, Wang, Wu, Zhao, and Liu]{colmin2026}
Zhe Huang, Zhaoxin Fan, Shuo Wang, Wenjun Wu, Xuan Zhao, and Min Liu.
\newblock {CoLMIN}: {LLM}-based multi-decision path negotiation for cooperative
  autonomous driving.
\newblock \emph{IEEE Transactions on Multimedia}, 2026.

\bibitem[Jiang et~al.(2023)Jiang, Chen, Xu, Liao, Chen, Zhou, Zhang, Liu,
  Huang, and Wang]{jiang2023vad}
Bo Jiang, Shaoyu Chen, Qing Xu, Bencheng Liao, Jiajie Chen, Helong Zhou, Qian
  Zhang, Wenyu Liu, Chang Huang, and Xinggang Wang.
\newblock {VAD}: Vectorized scene representation for efficient autonomous
  driving.
\newblock In \emph{Proceedings of the IEEE/CVF International Conference on
  Computer Vision (ICCV)}, 2023.

\bibitem[Kant and Zucker(1986)]{kant1986path}
Kamal Kant and Steven~W. Zucker.
\newblock Toward efficient trajectory planning: The path-velocity
  decomposition.
\newblock \emph{The International Journal of Robotics Research}, 5\penalty0
  (3):\penalty0 72--89, 1986.

\bibitem[Klimke et~al.(2022)Klimke, V{\"o}lz, and
  Buchholz]{klimke2022cooperative}
Marvin Klimke, Benjamin V{\"o}lz, and Michael Buchholz.
\newblock Cooperative behavior planning for automated driving using graph
  neural networks.
\newblock In \emph{IEEE Intelligent Vehicles Symposium (IV)}, 2022.

\bibitem[Klimke et~al.(2023)Klimke, V{\"o}lz, and
  Buchholz]{klimke2023automatic}
Marvin Klimke, Benjamin V{\"o}lz, and Michael Buchholz.
\newblock Automatic intersection management in mixed traffic using
  reinforcement learning and graph neural networks.
\newblock In \emph{IEEE Intelligent Vehicles Symposium (IV)}, 2023.

\bibitem[K{\"o}ppel et~al.(2019)K{\"o}ppel, Segner, Wagener, Pensel, Karwath,
  and Kramer]{koppel2019directranker}
Marius K{\"o}ppel, Alexander Segner, Martin Wagener, Lukas Pensel, Andreas
  Karwath, and Stefan Kramer.
\newblock Pairwise learning to rank by neural networks revisited:
  Reconstruction, theoretical analysis and practical performance.
\newblock In \emph{European Conference on Machine Learning and Principles and
  Practice of Knowledge Discovery in Databases (ECML PKDD)}, 2019.

\bibitem[Lang et~al.(2019)Lang, Vora, Caesar, Zhou, Yang, and
  Beijbom]{lang2019pointpillars}
Alex~H. Lang, Sourabh Vora, Holger Caesar, Lubing Zhou, Jiong Yang, and Oscar
  Beijbom.
\newblock {PointPillars}: Fast encoders for object detection from point clouds.
\newblock In \emph{Proceedings of the IEEE/CVF Conference on Computer Vision
  and Pattern Recognition (CVPR)}, 2019.

\bibitem[Lehmann et~al.(2018)Lehmann, G{\"u}nther, and
  Wolf]{lehmann2018generic}
Bernd Lehmann, Hendrik-J{\"o}rn G{\"u}nther, and Lars Wolf.
\newblock A generic approach towards maneuver coordination for automated
  vehicles.
\newblock In \emph{IEEE International Conference on Intelligent Transportation
  Systems (ITSC)}, 2018.

\bibitem[Lei et~al.(2026)Lei, Zhou, Li, Ma, and Hu]{riskmm2026}
Mingyue Lei, Zewei Zhou, Hongchen Li, Jiaqi Ma, and Jia Hu.
\newblock Risk map as middleware: Toward interpretable cooperative end-to-end
  autonomous driving for risk-aware planning.
\newblock \emph{IEEE Robotics and Automation Letters}, 11\penalty0 (1), 2026.

\bibitem[Lei et~al.(2022)Lei, Ren, Hu, Zhang, and Chen]{lei2022syncnet}
Zixing Lei, Shunli Ren, Yue Hu, Wenjun Zhang, and Siheng Chen.
\newblock Latency-aware collaborative perception.
\newblock In \emph{Proceedings of the European Conference on Computer Vision
  (ECCV)}, 2022.

\bibitem[Li et~al.(2023)Li, Xu, Liu, Ma, Chi, Ma, and Yu]{li2023lossy}
Jinlong Li, Runsheng Xu, Xinyu Liu, Jin Ma, Zicheng Chi, Jiaqi Ma, and Hongkai
  Yu.
\newblock Learning for vehicle-to-vehicle cooperative perception under lossy
  communication.
\newblock \emph{IEEE Transactions on Intelligent Vehicles}, 8\penalty0 (4),
  2023.

\bibitem[Li et~al.(2026{\natexlab{a}})Li, Zhang, Zhao, Wang, and
  Sun]{li2026defer}
Nuoran Li, Zhang Zhang, Yueran Zhao, Tianze Wang, and Chao Sun.
\newblock Defer to plan: Adaptive multi-agent fusion for end-to-end {V2X}
  driving.
\newblock \emph{arXiv preprint arXiv:2607.19774}, 2026{\natexlab{a}}.
\newblock Accepted at IEEE International Conference on Multimedia and Expo
  (ICME) 2026.

\bibitem[Li et~al.(2026{\natexlab{b}})Li, Hu, Liu, Jin, and
  Cai]{li2026n2cdrive}
Xingpeng Li, Enwen Hu, Jingrong Liu, Siyuan Jin, and Huirong Cai.
\newblock {N2CDrive}: Negotiate to cooperate for multi-agent autonomous driving
  via large vision-language model.
\newblock In \emph{IEEE International Conference on Acoustics, Speech and
  Signal Processing (ICASSP)}, 2026{\natexlab{b}}.

\bibitem[Li et~al.(2024)Li, Li, Wang, Lan, Yu, Ji, Li, Zhu, Kautz, Wu, Jiang,
  and Alvarez]{li2024hydramdp}
Zhenxin Li, Kailin Li, Shihao Wang, Shiyi Lan, Zhiding Yu, Yishen Ji, Zhiqi Li,
  Ziyue Zhu, Jan Kautz, Zuxuan Wu, Yu-Gang Jiang, and Jose~M. Alvarez.
\newblock {Hydra-MDP}: End-to-end multimodal planning with multi-target
  {Hydra}-distillation.
\newblock \emph{arXiv preprint arXiv:2406.06978}, 2024.

\bibitem[Liu et~al.(2025{\natexlab{a}})Liu, Liu, Wang, Yang, and
  Chen]{liu2025colmdriver}
Changxing Liu, Genjia Liu, Zijun Wang, Jinchang Yang, and Siheng Chen.
\newblock {CoLMDriver}: {LLM}-based negotiation benefits cooperative autonomous
  driving.
\newblock In \emph{Proceedings of the IEEE/CVF International Conference on
  Computer Vision (ICCV)}, 2025{\natexlab{a}}.

\bibitem[Liu et~al.(2025{\natexlab{b}})Liu, Hu, Xu, Mao, Ge, Huang, Lu, Xu,
  Xia, Wang, and Chen]{liu2025codriving}
Genjia Liu, Yue Hu, Chenxin Xu, Weibo Mao, Junhao Ge, Zhengxiang Huang, Yifan
  Lu, Yinda Xu, Junkai Xia, Yafei Wang, and Siheng Chen.
\newblock Toward collaborative autonomous driving: Simulation platform and
  end-to-end system.
\newblock \emph{IEEE Transactions on Pattern Analysis and Machine
  Intelligence}, 47\penalty0 (8), 2025{\natexlab{b}}.

\bibitem[Liu et~al.(2023)Liu, Huang, and Lv]{liu2023opgp}
Haochen Liu, Zhiyu Huang, and Chen Lv.
\newblock Occupancy prediction-guided neural planner for autonomous driving.
\newblock In \emph{IEEE International Conference on Intelligent Transportation
  Systems (ITSC)}, 2023.

\bibitem[Loshchilov and Hutter(2019)]{loshchilov2019adamw}
Ilya Loshchilov and Frank Hutter.
\newblock Decoupled weight decay regularization.
\newblock In \emph{International Conference on Learning Representations
  (ICLR)}, 2019.

\bibitem[Lu et~al.(2023)Lu, Li, Liu, Dianati, Feng, Chen, and
  Wang]{lu2023coalign}
Yifan Lu, Quanhao Li, Baoan Liu, Mehrdad Dianati, Chen Feng, Siheng Chen, and
  Yanfeng Wang.
\newblock Robust collaborative {3D} object detection in presence of pose
  errors.
\newblock In \emph{Proceedings of the IEEE International Conference on Robotics
  and Automation (ICRA)}, 2023.

\bibitem[Ma et~al.(2019)Ma, Harabor, Stuckey, Li, and Koenig]{ma2019searching}
Hang Ma, Daniel Harabor, Peter~J. Stuckey, Jiaoyang Li, and Sven Koenig.
\newblock Searching with consistent prioritization for multi-agent path
  finding.
\newblock In \emph{Proceedings of the AAAI Conference on Artificial
  Intelligence}, 2019.

\bibitem[Mertens and Buchholz(2025)]{mertens2025generalized}
Max~Bastian Mertens and Michael Buchholz.
\newblock Generalized coordination of partially cooperative urban traffic.
\newblock \emph{arXiv preprint arXiv:2505.20879}, 2025.

\bibitem[Nguyen et~al.(2026)Nguyen, Fauth, Jaeger, Dauner, Igl, Geiger, and
  Chitta]{nguyen2026lead}
Long Nguyen, Micha Fauth, Bernhard Jaeger, Daniel Dauner, Maximilian Igl,
  Andreas Geiger, and Kashyap Chitta.
\newblock {LEAD}: Minimizing learner-expert asymmetry in end-to-end driving.
\newblock In \emph{Proceedings of the IEEE/CVF Conference on Computer Vision
  and Pattern Recognition (CVPR)}, 2026.
\newblock arXiv:2512.20563.

\bibitem[Peng et~al.(2026)Peng, Lu, Zhou, Cui, Chen, and Wang]{omniv2x2026}
Juntong Peng, Juanwu Lu, Yupeng Zhou, Can Cui, Yaobin Chen, and Ziran Wang.
\newblock {OmniV2X}: A generative foundation planner for efficient end-to-end
  cooperative driving.
\newblock \emph{arXiv preprint arXiv:2606.21165}, 2026.
\newblock Accepted to IEEE/RSJ International Conference on Intelligent Robots
  and Systems (IROS) 2026.

\bibitem[Qu et~al.(2026)Qu, Chen, Shimizu, and Altintas]{qu2026cooperdrive}
Deyuan Qu, Qi Chen, Takayuki Shimizu, and Onur Altintas.
\newblock {CooperDrive}: Enhancing driving decisions through cooperative
  perception.
\newblock In \emph{IEEE International Conference on Robotics and Automation
  (ICRA)}, 2026.
\newblock arXiv:2604.14454.

\bibitem[Ren et~al.(2024)Ren, Lei, Wang, Dianati, Wang, Chen, and
  Zhang]{ren2024incop}
Shunli Ren, Zixing Lei, Zi Wang, Mehrdad Dianati, Yafei Wang, Siheng Chen, and
  Wenjun Zhang.
\newblock Interruption-aware cooperative perception for {V2X}
  communication-aided autonomous driving.
\newblock \emph{IEEE Transactions on Intelligent Vehicles}, 9\penalty0 (4),
  2024.

\bibitem[Ross et~al.(2011)Ross, Gordon, and Bagnell]{ross2011dagger}
St{\'e}phane Ross, Geoffrey~J. Gordon, and J.~Andrew Bagnell.
\newblock A reduction of imitation learning and structured prediction to
  no-regret online learning.
\newblock In \emph{Proceedings of the International Conference on Artificial
  Intelligence and Statistics (AISTATS)}, 2011.

\bibitem[Rowe et~al.(2023)Rowe, Ethier, Dykhne, and Czarnecki]{rowe2023fjmp}
Luke Rowe, Martin Ethier, Eli-Henry Dykhne, and Krzysztof Czarnecki.
\newblock {FJMP}: Factorized joint multi-agent motion prediction over learned
  directed acyclic interaction graphs.
\newblock In \emph{Proceedings of the IEEE/CVF Conference on Computer Vision
  and Pattern Recognition (CVPR)}, 2023.

\bibitem[Sadat et~al.(2020)Sadat, Casas, Ren, Wu, Dhawan, and
  Urtasun]{sadat2020p3}
Abbas Sadat, Sergio Casas, Mengye Ren, Xinyu Wu, Pranaab Dhawan, and Raquel
  Urtasun.
\newblock Perceive, predict, and plan: Safe motion planning through
  interpretable semantic representations.
\newblock In \emph{European Conference on Computer Vision (ECCV)}, 2020.
\newblock arXiv:2008.05930.

\bibitem[{SAE International}(2025)]{sae2025j3216}
{SAE International}.
\newblock Taxonomy and definitions for terms related to cooperative driving
  automation for on-road motor vehicles.
\newblock SAE Standard J3216\_202504, 2025.

\bibitem[Shao et~al.(2022)Shao, Wang, Chen, Li, and Liu]{shao2022interfuser}
Hao Shao, Letian Wang, Ruobing Chen, Hongsheng Li, and Yu Liu.
\newblock Safety-enhanced autonomous driving using interpretable sensor fusion
  transformer.
\newblock In \emph{Proceedings of the Conference on Robot Learning (CoRL)},
  2022.

\bibitem[Shao et~al.(2024)Shao, Hu, Wang, Song, Waslander, Liu, and
  Li]{shao2024lmdrive}
Hao Shao, Yuxuan Hu, Letian Wang, Guanglu Song, Steven~L. Waslander, Yu Liu,
  and Hongsheng Li.
\newblock {LMDrive}: Closed-loop end-to-end driving with large language models.
\newblock In \emph{Proceedings of the IEEE/CVF Conference on Computer Vision
  and Pattern Recognition (CVPR)}, 2024.

\bibitem[Song et~al.(2026)Song, Xia, Wang, Yu, Zhou, and Niu]{unimmv2x2026}
Ziyi Song, Chen Xia, Chenbing Wang, Haibao Yu, Sheng Zhou, and Zhisheng Niu.
\newblock {UniMM-V2X}: {MoE}-enhanced multi-level fusion for end-to-end
  cooperative autonomous driving.
\newblock In \emph{Proceedings of the AAAI Conference on Artificial
  Intelligence}, 2026.

\bibitem[Sun et~al.(2022)Sun, Huang, Gu, Williams, and Zhao]{sun2022m2i}
Qiao Sun, Xin Huang, Junru Gu, Brian~C. Williams, and Hang Zhao.
\newblock {M2I}: From factored marginal trajectory prediction to interactive
  prediction.
\newblock In \emph{Proceedings of the IEEE/CVF Conference on Computer Vision
  and Pattern Recognition (CVPR)}, 2022.

\bibitem[Sun et~al.(2025)Sun, Lin, Shi, Zhang, Wu, and
  Zheng]{sun2025sparsedrive}
Wenchao Sun, Xuewu Lin, Yining Shi, Chuang Zhang, Haoran Wu, and Sifa Zheng.
\newblock {SparseDrive}: End-to-end autonomous driving via sparse scene
  representation.
\newblock In \emph{Proceedings of the IEEE International Conference on Robotics
  and Automation (ICRA)}, 2025.

\bibitem[Sun et~al.(2026)Sun, Lin, Chen, Pei, Li, Shi, and
  Zheng]{sun2026sparsedrivev2}
Wenchao Sun, Xuewu Lin, Keyu Chen, Zixiang Pei, Xiang Li, Yining Shi, and Sifa
  Zheng.
\newblock {SparseDriveV2}: Scoring is all you need for end-to-end autonomous
  driving.
\newblock \emph{arXiv preprint arXiv:2603.29163}, 2026.

\bibitem[Tideman(1987)]{tideman1987independence}
T.~Nicolaus Tideman.
\newblock Independence of clones as a criterion for voting rules.
\newblock \emph{Social Choice and Welfare}, 4\penalty0 (3):\penalty0 185--206,
  1987.

\bibitem[van~den Berg and Overmars(2005)]{vandenberg2005prioritized}
Jur~P. van~den Berg and Mark~H. Overmars.
\newblock Prioritized motion planning for multiple robots.
\newblock In \emph{IEEE/RSJ International Conference on Intelligent Robots and
  Systems (IROS)}, pages 430--435, 2005.

\bibitem[VanMiddlesworth et~al.(2008)VanMiddlesworth, Dresner, and
  Stone]{vanmiddlesworth2008replacing}
Mark VanMiddlesworth, Kurt Dresner, and Peter Stone.
\newblock Replacing the stop sign: Unmanaged intersection control for
  autonomous vehicles.
\newblock In \emph{International Joint Conference on Autonomous Agents and
  Multiagent Systems (AAMAS)}, 2008.

\bibitem[Wang et~al.(2020)Wang, Manivasagam, Liang, Yang, Zeng, and
  Urtasun]{wang2020v2vnet}
Tsun-Hsuan Wang, Sivabalan Manivasagam, Ming Liang, Bin Yang, Wenyuan Zeng, and
  Raquel Urtasun.
\newblock {V2VNet}: Vehicle-to-vehicle communication for joint perception and
  prediction.
\newblock In \emph{Proceedings of the European Conference on Computer Vision
  (ECCV)}, 2020.

\bibitem[Warrington et~al.(2021)Warrington, Lavington, {\'S}cibior, Schmidt,
  and Wood]{warrington2021a2d}
Andrew Warrington, J.~Wilder Lavington, Adam {\'S}cibior, Mark Schmidt, and
  Frank Wood.
\newblock Robust asymmetric learning in {POMDPs}.
\newblock In \emph{International Conference on Machine Learning (ICML)}, 2021.

\bibitem[Wartnaby and Bellan(2019)]{wartnaby2019decentralised}
Charles Wartnaby and Daniele Bellan.
\newblock Decentralised cooperative collision avoidance with reference-free
  model predictive control and desired versus planned trajectories.
\newblock \emph{arXiv preprint arXiv:1904.07053}, 2019.

\bibitem[Wei et~al.(2023)Wei, Wei, Hu, Lu, Zhong, Chen, and
  Zhang]{wei2023cobevflow}
Sizhe Wei, Yuxi Wei, Yue Hu, Yifan Lu, Yiqi Zhong, Siheng Chen, and Ya Zhang.
\newblock Asynchrony-robust collaborative perception via bird's eye view flow.
\newblock In \emph{Advances in Neural Information Processing Systems}, 2023.

\bibitem[Weihs et~al.(2021)Weihs, Jain, Liu, Salvador, Lazebnik, Kembhavi, and
  Schwing]{weihs2021advisor}
Luca Weihs, Unnat Jain, Iou-Jen Liu, Jordi Salvador, Svetlana Lazebnik,
  Aniruddha Kembhavi, and Alexander Schwing.
\newblock Bridging the imitation gap by adaptive insubordination.
\newblock In \emph{Advances in Neural Information Processing Systems
  (NeurIPS)}, 2021.

\bibitem[Wen et~al.(2020)Wen, Lin, Darrell, Jayaraman, and Gao]{wen2020copycat}
Chuan Wen, Jierui Lin, Trevor Darrell, Dinesh Jayaraman, and Yang Gao.
\newblock Fighting copycat agents in behavioral cloning from observation
  histories.
\newblock In \emph{Advances in Neural Information Processing Systems}, 2020.

\bibitem[Werling et~al.(2010)Werling, Ziegler, Kammel, and
  Thrun]{werling2010optimal}
Moritz Werling, Julius Ziegler, S{\"o}ren Kammel, and Sebastian Thrun.
\newblock Optimal trajectory generation for dynamic street scenarios in a
  {Fren\'{e}t} frame.
\newblock In \emph{IEEE International Conference on Robotics and Automation
  (ICRA)}, 2010.

\bibitem[Wu et~al.(2022)Wu, Jia, Chen, Yan, Li, and Qiao]{wu2022tcp}
Penghao Wu, Xiaosong Jia, Li Chen, Junchi Yan, Hongyang Li, and Yu Qiao.
\newblock Trajectory-guided control prediction for end-to-end autonomous
  driving: A simple yet strong baseline.
\newblock In \emph{Advances in Neural Information Processing Systems}, 2022.

\bibitem[Wu et~al.(2026)Wu, Zhang, He, Wu, Shan, Qiu, Gao, Ke, and
  Zhang]{wu2026aligndrive}
Yanhao Wu, Haoyang Zhang, Fei He, Rui Wu, Yanhu Shan, Congpei Qiu, Liang Gao,
  Wei Ke, and Tong Zhang.
\newblock {AlignDrive}: Aligned lateral-longitudinal planning for end-to-end
  autonomous driving.
\newblock In \emph{Advances in Neural Information Processing Systems
  (NeurIPS)}, 2026.
\newblock arXiv:2601.01762.

\bibitem[Xu et~al.(2022{\natexlab{a}})Xu, Xiang, Tu, Xia, Yang, and
  Ma]{xu2022v2xvit}
Runsheng Xu, Hao Xiang, Zhengzhong Tu, Xin Xia, Ming-Hsuan Yang, and Jiaqi Ma.
\newblock {V2X-ViT}: Vehicle-to-everything cooperative perception with vision
  transformer.
\newblock In \emph{Proceedings of the European Conference on Computer Vision
  (ECCV)}, 2022{\natexlab{a}}.

\bibitem[Xu et~al.(2022{\natexlab{b}})Xu, Xiang, Xia, Han, Li, and
  Ma]{xu2022opv2v}
Runsheng Xu, Hao Xiang, Xin Xia, Xu Han, Jinlong Li, and Jiaqi Ma.
\newblock {OPV2V}: An open benchmark dataset and fusion pipeline for perception
  with vehicle-to-vehicle communication.
\newblock In \emph{Proceedings of the IEEE International Conference on Robotics
  and Automation (ICRA)}, 2022{\natexlab{b}}.

\bibitem[Xu et~al.(2026)Xu, Wu, Wu, Xu, Jiang, Wang, Ke, Jiang, Zhang, and
  Wang]{aurora2026}
Yitao Xu, Tong Wu, Yiyan Wu, Guoji Xu, Yanbo Jiang, Jiahao Wang, Zehong Ke,
  Junkai Jiang, Fang Zhang, and Jianqiang Wang.
\newblock Roadside-cooperative autonomous driving: From data platform to
  vision-language end-to-end reasoning.
\newblock \emph{arXiv preprint arXiv:2608.21032}, 2026.

\bibitem[Yin et~al.(2021)Yin, Zhou, and Kr{\"a}henb{\"u}hl]{yin2021centerpoint}
Tianwei Yin, Xingyi Zhou, and Philipp Kr{\"a}henb{\"u}hl.
\newblock Center-based {3D} object detection and tracking.
\newblock In \emph{Proceedings of the IEEE/CVF Conference on Computer Vision
  and Pattern Recognition (CVPR)}, 2021.

\bibitem[You et~al.(2026)You, Tang, Wang, Zhao, Hua, Shi, Zhang, Wang, and
  Ran]{you2026v2xwam}
Junwei You, Weizhe Tang, Can Wang, Yan Zhao, Jun Hua, Haotian Shi, Wei Zhang,
  Lin Wang, and Bin Ran.
\newblock {V2X-WAM}: A cooperative world action model for end-to-end autonomous
  driving.
\newblock \emph{arXiv preprint arXiv:2609.37098}, 2026.

\bibitem[Yu et~al.(2025)Yu, Yang, Zhong, Yang, Fan, Luo, and Nie]{yu2025univ2x}
Haibao Yu, Wenxian Yang, Jiaru Zhong, Zhenwei Yang, Siqi Fan, Ping Luo, and
  Zaiqing Nie.
\newblock End-to-end autonomous driving through {V2X} cooperation.
\newblock In \emph{Proceedings of the AAAI Conference on Artificial
  Intelligence}, 2025.

\bibitem[Zeng et~al.(2019)Zeng, Luo, Suo, Sadat, Yang, Casas, and
  Urtasun]{zeng2019nmp}
Wenyuan Zeng, Wenjie Luo, Simon Suo, Abbas Sadat, Bin Yang, Sergio Casas, and
  Raquel Urtasun.
\newblock End-to-end interpretable neural motion planner.
\newblock In \emph{Proceedings of the IEEE/CVF Conference on Computer Vision
  and Pattern Recognition (CVPR)}, 2019.

\bibitem[Zhang et~al.(2023)Zhang, Li, and Li]{zhang2023alphaorder}
Jiawei Zhang, Shen Li, and Li Li.
\newblock Coordinating {CAV} swarms at intersections with a deep learning
  model.
\newblock \emph{IEEE Transactions on Intelligent Transportation Systems}, 2023.

\bibitem[Zhang et~al.(2026)Zhang, Xiong, Wang, Teng, Knoll, and
  Chen]{zhang2026tsc}
Xiaotong Zhang, Gang Xiong, Yuanjing Wang, Siyu Teng, Alois Knoll, and Long
  Chen.
\newblock {TSC}: Topology-conditioned stackelberg coordination for multi-agent
  reinforcement learning in interactive driving.
\newblock \emph{arXiv preprint arXiv:2602.23896}, 2026.

\bibitem[Zhong et~al.(2026)Zhong, Sicking, Hueger, and
  Gottschalk]{zhong2026rogdagger}
Liangyu Zhong, Joachim Sicking, Fabian Hueger, and Hanno Gottschalk.
\newblock {RoG-DAgger}: Rollout-guided post-training for end-to-end driving.
\newblock \emph{arXiv preprint arXiv:2608.24525}, 2026.

\bibitem[Zhu et~al.(2021)Zhu, Su, Lu, Li, Wang, and Dai]{zhu2021deformable}
Xizhou Zhu, Weijie Su, Lewei Lu, Bin Li, Xiaogang Wang, and Jifeng Dai.
\newblock {Deformable DETR}: Deformable transformers for end-to-end object
  detection.
\newblock In \emph{International Conference on Learning Representations
  (ICLR)}, 2021.

\bibitem[Zimmerlin et~al.(2024)Zimmerlin, Bei{\ss}wenger, Jaeger, Geiger, and
  Chitta]{zimmerlin2024hidden}
Julian Zimmerlin, Jens Bei{\ss}wenger, Bernhard Jaeger, Andreas Geiger, and
  Kashyap Chitta.
\newblock Hidden biases of end-to-end driving datasets.
\newblock \emph{arXiv preprint arXiv:2412.09602}, 2024.

\end{thebibliography}
}

\clearpage
\setcounter{page}{1}\renewcommand{\thepage}{S\arabic{page}}
\maketitlesupplementary
\DeclareRobustCommand{\sota}[1]{\textbf{#1}}  %
\raggedbottom  %
\renewcommand{\topfraction}{0.95}\renewcommand{\bottomfraction}{0.9}\renewcommand{\textfraction}{0.05}
\renewcommand{\floatpagefraction}{0.85}\renewcommand{\dbltopfraction}{0.95}\renewcommand{\dblfloatpagefraction}{0.85}
\setcounter{topnumber}{4}\setcounter{bottomnumber}{2}\setcounter{totalnumber}{6}\setcounter{dbltopnumber}{4}
\makeatletter\setlength{\@fptop}{0pt}\setlength{\@dblfptop}{0pt}%
\setlength{\@fpsep}{10pt plus 2pt}\setlength{\@dblfpsep}{10pt plus 2pt}\makeatother  %

\section{Loss terms}\label{sec:supp_loss}

The terms of \cref{eq:loss}: with $\mathcal{L}_{\mathrm{det}}$ and the five planning terms summed over the three fusion or planner layers.
$\mathcal{L}_{\mathrm{det}}$ and $\mathcal{L}_{\mathrm{aux}}$ are set losses with Hungarian matching~\cite{carion2020detr} on the fused and on the ego's own detections. $\mathcal{L}_{\mathrm{rel}}$ is a cross-entropy between $r_m$ and the share of confident instances of source $m$ near a ground-truth box.
$\mathcal{L}_{\mathrm{cls}}$ is a cross-entropy between $z_a$ and $\operatorname{softmax}_a\bigl(-\tfrac{1}{T\tau}\sum_k\lvert s^a_k-\tilde s_k\rvert\bigr)$, with $\tilde{\vect{s}}$ the label profile and $\tau{=}0.5$\,m. $\mathcal{L}_{\mathrm{reg}}$ and $\mathcal{L}_{\mathrm{lat}}$ are $L_1$ losses on profile and lateral offsets of the closest candidate. $\mathcal{L}_{\mathrm{risk}}$ is a binary cross-entropy (positive weight $3.0$), and $\mathcal{L}_{\mathrm{nom}}$ is $L_1$ on $\vect{\nu}$ plus $0.5$ times $L_1$ on $\vect{\ell}$.

\section{Order gate}\label{sec:supp_gate}
\begin{figure*}[t]
  \centering
  \includegraphics[width=\textwidth]{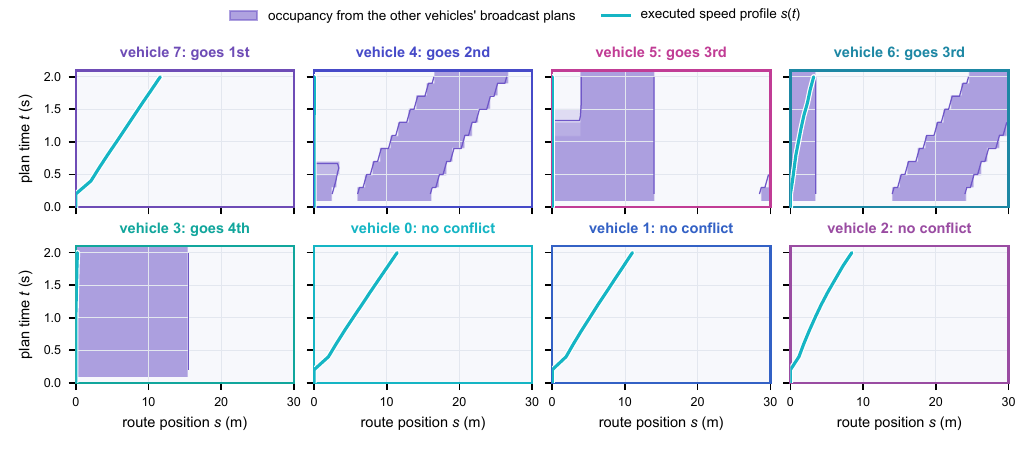}
  \caption{\textbf{The conflict maps of \cref{fig:teaser}, with axes.} InterDrive r40 at $t{=}5$\,s: the partner-plan channel of each vehicle's route-time conflict map, rebuilt from the log with the network's own function, and its executed speed profile, in the pass order the vehicles agreed on. A vehicle that goes later keeps its profile out of the cells that the earlier vehicles' plans occupy on its route (vehicles 4, 5 and 3 wait at their entry); vehicles without a conflict drive freely.}
  \label{fig:supp_maps}
\end{figure*}

\noindent\textbf{Inputs.}
Every vehicle broadcasts its intended trajectory: its free-flow drive along its route ($2.0$\,m/s$^2$ up to $6.0$\,m/s, or its current speed if that is higher) as ten points $0.6$\,s apart.
Two intended trajectories meet if their positions come within $3.0$\,m of each other.
A vehicle evaluates the precedence logit of \cref{eq:prec} for every pair of the vehicles it hears whose intended trajectories meet, itself included.
The features of a pair do not depend on the receiver's frame, so two vehicles that hear the same vehicles compute the same logits. $g$ reads a pair as an $11\times11$ image over its point pairs (each vehicle's start point and ten intended points). The image has seven channels: the distance of the two points, the cosine and sine of their heading difference, the two speeds and the two time indices. Two $3\times3$ convolutions ($48$ channels), a $1\times1$ convolution to $16$ and a two-layer MLP ($128$ hidden units) give one value, and \cref{eq:prec} takes the difference of the two orders.

\noindent\textbf{Decoding.}
Pairwise precedence is not transitive: three vehicles can each be told to wait for the next one.
The pairs are therefore taken by decreasing $\lvert\eta\rvert$ (ties by name), and a pair that would close a cycle with the pairs already taken is reversed. This yields an order without cycles, and vehicles that hear the same vehicles decode the same order. \Cref{fig:supp_maps} shows such an order for the eight vehicles of \cref{fig:teaser} and how each vehicle's conflict map follows it. The order is decoded anew at every planning step from the current intended trajectories, so it follows the vehicles as they move. A vehicle that holds precedence but stands for another reason keeps the other waiting, which causes the stalls named in \cref{sec:conclusion}.

\begin{figure*}[t]
  \centering
  \includegraphics[width=\textwidth]{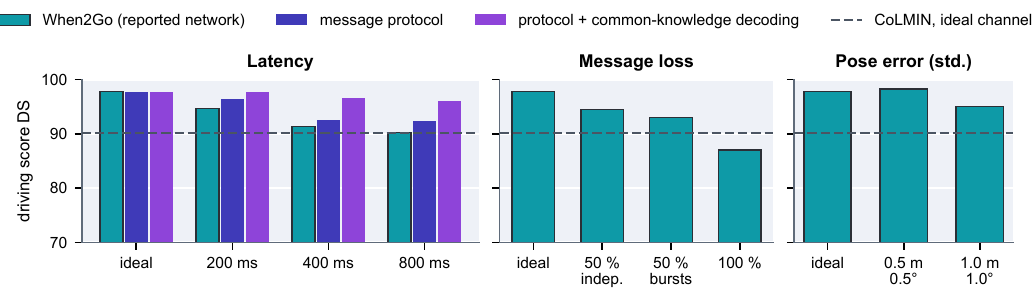}
  \caption{\textbf{Imperfect channels.} InterDrive without traffic, one repetition (\cref{tab:robust}). Left: latency, for the reported network, the separate message protocol and the protocol with common-knowledge decoding (\cref{sec:supp_gate}). Middle: message loss. Right: the senders' pose error (standard deviation). Dashed: CoLMIN, the best published method, with an ideal channel.}
  \label{fig:supp_channel}
\end{figure*}

\noindent\textbf{Communication model.}
The simulation steps synchronously at $20$\,Hz, and every vehicle plans and broadcasts every $0.2$\,s. With the ideal channel, a message is read at the step it is sent.
All intended trajectories of a step are formed before any vehicle plans, so every vehicle reads the same ones. The $2$\,s plans are read as soon as they exist. The vehicles plan one after the other within a step, so a vehicle reads the current plans of the vehicles before it and the previous step's plans of those after it. This is more than a real channel delivers; the $200$\,ms row of \cref{tab:robust}, which delays every message by one planning step, is the realisable reference.
On InterDrive and MDrive, a receiver fuses at most three sources, its own RSU first and then the nearest within $70$\,m. It also reads the broadcast plans and intended trajectories of up to three further vehicles heard within the same range.
Each vehicle evaluates a pair with its own current intended trajectory and the partner's latest message. With the ideal channel, both vehicles of a pair thus evaluate the same two trajectories. With latency or loss they may not, which is part of what \cref{tab:robust} measures.
The code can instead make both vehicles evaluate the same two broadcast messages under latency (common-knowledge decoding, on the separate message protocol of the next paragraphs); the reported runs do not use it. With it, the driving score stays within noise of the ideal channel ($97.79$) at every latency (\cref{fig:supp_channel}): $97.69$, $96.68$ and $95.96$ at $200$, $400$ and $800$\,ms. Paired against the ideal channel, the differences are $-0.1$ ($[-1.3, +1.0]$), $-1.1$ ($[-4.0, +1.0]$) and $-1.8$ ($[-4.2, +0.1]$). On the same protocol without it, the scores are $96.34$, $92.58$ and $92.35$, and the one network scores $94.63$, $91.35$ and $90.22$. Common-knowledge decoding removes every disagreement at $200$ and $400$\,ms, where $2.4$ and $4.3\,\%$ of the pair-steps disagree without it. At $800$\,ms, $2$ of the $10{,}156$ pair-steps still disagree, against $6.3\,\%$ without it (\cref{tab:supp_coord}). Under bursty loss without latency the two vehicles already evaluate the same messages, and the decoding changes nothing ($93.90$).
Two vehicles therefore act on complementary orders when five conditions hold. They (i) evaluate the same two intended trajectories, (ii) with the same weights of $g$ and (iii) in the same arithmetic. In addition, (iv) each hears the other and finds that their intended trajectories meet, and (v) for the decoding, both hear the same set of vehicles. The evaluated tests have up to eight controlled vehicles. The decoding breaks a tie $\eta_{uu'}=0$ by the vehicles' names, the same way on both sides.
Every controlled vehicle runs the same network; background vehicles and pedestrians are perceived but do not communicate.

\noindent\textbf{Alternatives.}
For each partner whose intended trajectory meets its own, the ego forms two plans by fixed geometry.
If the ego goes first, it reads the plan the partner broadcasts, cut where it would come within $3.5$\,m of the ego's intended path: the partner stands there.
If the partner goes first, it reads a vehicle of its own size standing across its path just inside the meeting zone. This vehicle is placed so that the planner stops the ego's reference point $1.0$\,m before the zone.
The decoded order chooses one of the two; the chosen plan enters the partner-plan channel of the conflict map and the planner's tokens.
The precedence network runs in float32 inside the forward pass.

\noindent\textbf{Training of the precedence network.}
$g$ is trained on $400{,}000$ counterfactual pairs built from the paths of the training towns and labelled by the right-of-way rule of \cref{sec:supervision}. The rule's tie window is $0.6$\,s, the spacing of the intended points, against $0.4$\,s on the $0.2$\,s plans of the planner's labels. On $40{,}000$ pairs built from the paths of the validation towns, the precedence network agrees with the rule in $98.4\,\%$ of the pairs ($99.5\,\%$ of the clear ones, $94.9\,\%$ of the near ties), and $\eta_{uu'}=-\eta_{u'u}$ holds exactly.

\noindent\textbf{One network.}
Before the gate became a part of the network, the same decisions were computed by a separate message protocol between the network and the controller.
On the CPU in float32, both make the same decision at all $162$ planning steps of InterDrive test r23, and their plans differ by at most $2.4\times10^{-7}$\,m.
Over the evaluation, the one network minus the protocol is $-0.01$ DS on InterDrive without traffic ($95\,\%$ interval $[-0.03, 0.00]$, $45$ tests). With background traffic, which is not deterministic, it is $-0.33$ ($[-1.56, +0.72]$, $46$ tests, three repetitions), and on MDrive $+0.61$ ($[-0.02, +1.25]$, the $218$ scenarios both completed).
The InterDrive and MDrive numbers of \cref{tab:main} are those of the one network.
The code also holds a soft gate, a latch of earlier decisions and a hold for lost messages; none of them is used in the reported runs.

\noindent\textbf{Coordination in closed loop.}
\Cref{tab:supp_coord} measures the order directly, from the logs of the reported InterDrive runs and of the runs of \cref{tab:robust}.
With the ideal channel, the two vehicles of an ordered pair never disagree, in $42{,}789$ steps at which both give the pair a role, without and with traffic.
Latency makes them evaluate different messages: $2.2$ to $5.7\,\%$ of these steps disagree, mostly with both going. Under the two levels of pose noise, $4.7$ and $5.6\,\%$ disagree.
With lost messages most pairs are ordered by one vehicle only, which has heard the other, and disagreement stays below $1.1\,\%$.
The decoding without cycles turns a pairwise decision around at $51$ planning steps in $9$ of the $46$ tests without traffic.
Stalls, in which both vehicles of an ordered pair stand for more than $10$\,s, occur in $3$ of the $46$ tests without traffic and in $24$ of the $138$ runs with traffic. None of them coincides with a disagreement, and most end when one vehicle leaves the simulation. These are the stalls named in \cref{sec:conclusion}.
An offline reconstruction of the decoding from the logged messages reproduces $97$ to $99\,\%$ of the logged roles ($89$ to $95\,\%$ with pose noise, whose noisy poses are not logged).

\noindent\textbf{Without messages.}
With background traffic (repetition $3$), the same network without any message scores $76.92$ against $91.11$ with them (paired $-14.2$, $95\,\%$ interval $[-21.5, -7.0]$, $46$ tests). Lane changes fall from $96.88$ to $41.48$, and $31$ vehicle collisions occur in $9$ of the $46$ tests, against none with messages. Without traffic the drop is $10.7$ points (\cref{tab:robust}). On V2Xverse (repetitions $3$ to $5$) it scores $64.59$ against $68.31$ (paired $-3.7$, $[-7.4, +0.1]$), with more runs ending blocked ($31$ against $27$ of $100$) and more vehicle collisions ($25$ against $16$ in the $255$ runs). On MDrive (repetition $3$) it scores $86.71$ against $88.16$ (paired $-1.5$, $[-3.5, +0.6]$; in the Interaction bucket $-3.1$, $[-5.1, -1.2]$). In this repetition, the share of collision-free vehicles falls from $94.9$ to $91.7\,\%$ ($29$ instead of $5$ vehicle collisions per $100$ scenarios). Without any message the planner thus still scores above every rerun baseline on V2Xverse and MDrive ($59.09$ and $85.13$ for the strongest); the messages add the rest, most where controlled vehicles meet.

\begin{table*}[t]
  \centering
  \footnotesize
  \setlength{\tabcolsep}{3pt}
  \begin{tabularx}{\textwidth}{@{}>{\raggedright\arraybackslash}X ccccccccccc@{}}
    \toprule
    Channel & runs & pair-steps & disagree (\%) & both go / both yield & one-sided (\%) & flips (\%) & cycles & stalls & cleared & roles (\%) \\
    \midrule
    Ideal, no traffic & 46 & 8{,}036 & \textbf{0.00} & \textbf{0 / 0} & 9.5 & 1.7 & 51 (9) & 4 (3) & 1 & 96.8 \\
    Ideal, with traffic & 138 & 34{,}753 & \textbf{0.00} & \textbf{0 / 0} & 9.6 & 1.7 & 312 (24) & 80 (24) & 35 & 98.4 \\
    \midrule
    \quad $200$\,ms & 46 & 9{,}806 & 2.24 & 201 / 19 & 13.0 & 1.9 & 129 (10) & 16 (5) & 8 & 97.4 \\
    \quad $400$\,ms & 46 & 9{,}817 & 4.26 & 377 / 41 & 12.8 & 1.9 & 86 (13) & 15 (8) & 7 & 98.6 \\
    \quad $800$\,ms & 46 & 11{,}488 & 5.74 & 589 / 70 & 15.2 & 2.0 & 165 (12) & 34 (8) & 10 & 98.9 \\
    \quad $50\,\%$ loss, independent & 46 & 2{,}821 & 0.28 & 4 / 4 & 67.6 & 0.8 & 47 (9) & 13 (2) & 3 & 97.8 \\
    \quad $50\,\%$ loss, in bursts & 46 & 2{,}545 & 1.02 & 17 / 9 & 68.1 & 1.6 & 99 (7) & 7 (2) & 2 & 98.2 \\
    \quad $0.5$\,m, $0.5^\circ$ & 46 & 8{,}807 & 4.73 & 220 / 197 & 16.7 & 9.0 & 55 (11) & 7 (4) & 5 & 92.2 \\
    \quad $1.0$\,m, $1.0^\circ$ & 46 & 10{,}330 & 5.63 & 314 / 268 & 22.1 & 9.3 & 91 (10) & 10 (4) & 5 & 88.8 \\
    \quad combined & 46 & 5{,}306 & 5.39 & 181 / 105 & 50.9 & 3.9 & 152 (10) & 12 (2) & 7 & 95.1 \\
    \midrule
    \quad $200$\,ms, protocol & 46 & 8{,}313 & 2.44 & 186 / 17 & 11.6 & 2.1 & 67 (9) & 4 (3) & 2 & 96.9 \\
    \quad $200$\,ms, common knowledge & 46 & 7{,}833 & \textbf{0.00} & \textbf{0 / 0} & 11.9 & 2.1 & 53 (10) & 3 (2) & 2 & 97.8 \\
    \quad $400$\,ms, protocol & 46 & 9{,}965 & 4.33 & 386 / 45 & 14.0 & 1.9 & 139 (10) & 20 (4) & 5 & 98.7 \\
    \quad $400$\,ms, common knowledge & 46 & 8{,}792 & \textbf{0.00} & \textbf{0 / 0} & 13.5 & 1.9 & 50 (10) & 7 (2) & 4 & 97.6 \\
    \quad $800$\,ms, protocol & 46 & 10{,}707 & 6.29 & 598 / 75 & 16.1 & 2.1 & 164 (12) & 25 (7) & 10 & 97.5 \\
    \quad $800$\,ms, common knowledge & 46 & 10{,}156 & 0.02 & 2 / 0 & 18.2 & 2.3 & 182 (10) & 20 (3) & 4 & 97.1 \\
    \quad $50\,\%$ loss in bursts, protocol$^{\dagger}$ & 46 & 2{,}417 & 0.54 & 12 / 1 & 68.1 & 1.5 & 35 (8) & 7 (2) & 2 & 97.7 \\
    \bottomrule
  \end{tabularx}
  \caption{\textbf{Coordination in closed loop.} InterDrive, the reported runs (ideal channel) and those of \cref{tab:robust}. A pair of controlled vehicles is ordered when one of them gives it a role. Pair-steps: planning steps ($0.2$\,s) at which both vehicles of an ordered pair give it a role. Disagree: their roles do not complement each other (both go or both yield). One-sided: share of the steps of ordered pairs at which only one vehicle gives a role. Flips: changes of a vehicle's role between consecutive steps, per step of an ordered pair. Cycles: planning steps at which the decoding turned a pairwise decision around (runs). Stalls: episodes in which both vehicles of an ordered pair stand for more than $10$\,s (runs), and how many end with one of them driving off (cleared). The other stalls end when one vehicle leaves the simulation, arrived or blocked out. Roles: share of the logged roles that an offline reconstruction of the decoding reproduces. Last rows: the separate message protocol of \cref{sec:supp_gate}, without and with common-knowledge decoding ($^{\dagger}$: without latency both are the same run).}
  \label{tab:supp_coord}
\end{table*}

\section{Experimental setup in detail}\label{sec:supp_setup}
\subsection{Benchmarks and metrics}\label{sec:benchmarks}

We evaluate in closed loop on three CARLA~\cite{dosovitskiy2017carla} benchmarks (\cref{tab:benchmarks}); InterDrive and MDrive build on the evaluator of V2Xverse and keep its infraction penalties and blocked test.

\begin{table}[t]
  \centering
  \footnotesize
  \setlength{\tabcolsep}{3pt}
  \begin{tabularx}{\linewidth}{@{}l*{3}{>{\raggedright\arraybackslash}X}@{}}
    \toprule
    & V2Xverse & InterDrive & MDrive \\
    \midrule
    CARLA & 0.9.10.1 & 0.9.10.1 & 0.9.12 \\
    Tests & 85 runs in 5 parts & 46 (IC 18, LM 20, LC 8), in 2 traffic settings & 225 scenarios in 3 buckets (46, 112, 67) \\
    Vehicles & 1 & 2 to 8 & 1 to 8 \\
    Traffic & 55 to 120 vehicles, 60 to 120 pedestrians, 60 to 120 bicycles & none; or 50 vehicles, 50 pedestrians, 50 bicycles & placed by each scenario \\
    DS of a test & RC $\times$ IS & mean RC $\times$ product of IS over its vehicles & mean of RC $\times$ IS over its vehicles \\
    Reported & mean over runs & mean over tests, per setting & mean of the 3 bucket means \\
    \bottomrule
  \end{tabularx}
  \caption{\textbf{Benchmarks.} Vehicles: controlled vehicles per test. Traffic: background actors. IC, LM, LC: intersection crossing, lane merging, lane changing. RC, IS, DS: see \cref{sec:benchmarks}.}
  \label{tab:benchmarks}
\end{table}

\noindent\textbf{V2Xverse.}
We run the evaluation script of the public release~\cite{liu2025codriving}: $85$ runs in Town05 in five parts. Each part has its own scenario parameters (background traffic as in \cref{tab:benchmarks}, traffic lights held at green) and agent configuration.
The LiDAR is a rotating unit with $64$ beams at $20$\,Hz, $85$\,m range and a vertical field of view from $-30^\circ$ to $10^\circ$. It is mounted $1.3$\,m ahead of the vehicle centre and $1.85$\,m above the ground.
A run is one route driven by one controlled vehicle with one RSU. It ends early at its time limit, when the vehicle leaves its route, or after $30$\,s below $0.5$\,m/s (the \emph{blocked test}).
Averaged over the runs, we report route completion (RC, in per cent), the infraction score (IS) and the driving score $\mathrm{DS}=\mathrm{RC}\cdot\mathrm{IS}$. IS is the product of $0.50$ per collision with a pedestrian, $0.60$ with a vehicle, $0.65$ with the layout, $0.70$ per red light, $0.80$ per stop sign, and the share of the route driven inside its lanes.

\noindent\textbf{InterDrive}~\cite{liu2025colmdriver} tests two to eight controlled vehicles that meet at intersections, in merges and in lane changes, without and with background traffic.
Its script multiplies the mean RC of a test's vehicles by the product of their IS, so a collision between two of them is penalised twice.

\noindent\textbf{MDrive}~\cite{mdrive2026} has three buckets: InterDrive (the routes above, without background traffic), Interaction and V2X-PnP (replays of real-world logs).
Its average is the unweighted mean of the three bucket means, not the mean over all scenarios.

\subsection{Protocol}\label{sec:protocol}

\noindent\textbf{Repetitions and seeds.}
Repetition $r$ sets the seeds of the simulator's actor provider and traffic manager to $2000+r$; $r{=}0$ is the seed of the benchmarks' scripts.
Design choices were made on repetitions $0$ to $2$. The numbers of \ours come from repetitions on which no choice was made. These are repetitions $3$ to $8$ on V2Xverse ($510$ runs), $3$ to $5$ on InterDrive with traffic and on MDrive, and repetition $3$ on InterDrive without traffic, which is deterministic. On V2Xverse, repetitions $3$ to $5$ were run first (mean $68.31$). Repetitions $6$ to $8$ were then added under a commitment, made before they ran, to report the mean over all six, whatever it would be ($70.00$). The rerun baselines on V2Xverse run in their release configuration (rendered, with the release's roadside units) on repetition $0$, and on three repetitions, $0$ to $2$, for no fusion, \cod and \dtp. Their runs are listed in \cref{tab:supp_v2xverse}. On repetition $2$, without rendering and with LiDAR-only roadside units, as \ours runs, \cod scores $56.96$ against $59.58$ rendered (paired $-2.6$, $[-9.8, +4.6]$ over $85$ routes).

\noindent\textbf{Noise and intervals.}
On the public V2Xverse runs, the $95\,\%$ bootstrap interval of the paired difference between two single repetitions is about $\pm5$ to $6$ DS for two runs of the same policy. Between the ablations of \cref{tab:supp_ablation_ci}, it is $\pm6$ to $8$ DS, so one repetition resolves only differences of about $7$ DS.
We average the repetitions of each route first and report paired differences with $95\,\%$ percentile-bootstrap intervals over routes ($10{,}000$ resamples, stratified by category on InterDrive and by bucket on MDrive).

\noindent\textbf{Machines.}
The runs come from a GPU cluster (NVIDIA L4, L40S and A100 GPUs with AMD EPYC CPUs) and a workstation (one NVIDIA RTX PRO 6000 GPU with an Intel CPU).
\ours runs on the cluster's CPUs in single precision on all benchmarks; the baselines run on GPUs, as released, and the runs of one baseline on one benchmark may come from both machines.
This favours no method: every benchmark advances the simulation synchronously in steps of $0.05$\,s and waits for the actions of all vehicles before the next step. The time a method takes thus never changes what happens in the simulation.
A machine changes only floating-point rounding, which the closed loop amplifies like a different seed and which the repetitions and the paired comparisons absorb.
Rerun on the workstation, the evaluation of \ours on InterDrive without traffic scores $97.36$ against the cluster's $97.79$. Of the $46$ tests, $42$ end with the same score, and the paired difference is $+0.43$ ($95\,\%$ interval $[-0.02, +1.30]$).
On one node type of the cluster (NVIDIA V100S), nearly every forward pass of early fusion failed with a cuDNN error, after which its vehicles brake. Its $140$ MDrive scenarios from these nodes were discarded and rerun on other GPUs.

\noindent\textbf{Published rows and reruns.}
Rows copied from publications are marked.
On V2Xverse they are for reference only. The published scores~\cite{liu2025codriving,li2026defer} were obtained on $67$ routes whose list is not in the public release, and users report lower scores for the released checkpoint with the public script~\cite{v2xverse_issue44,v2xverse_issue48}.
Our reruns of released baselines use the released checkpoints with the benchmark's own agent code, controller and configuration, unmodified except where stated below.
On MDrive this is MDrive's copy of the code. Unlike V2Xverse's, its perception gives zero features instead of raising an error when a LiDAR sweep has no point in the detection range. This matters on the elevated roads of the V2X-PnP map: the released preprocessing adds the sensor's height above the map origin to every point, which there lifts all points out of the range.
Our first pass used V2Xverse's copy and braked at every such step. The $125$ runs that hit an inference error were rerun with MDrive's copy and its handling of a failed step (a brake for one simulation step, as in MDrive's agent).
Where a protocol is public, the same \cod setup reproduces a third party's number. On MDrive, with the released weights (byte-identical to those MDrive uses), our rerun averages $82.80$ against the published $82.67$~\cite{mdrive2026} (repetition $0$, all $225$ scenarios). By bucket it scores $79.20$ against $79.06$ on Interaction and $91.00$ against $84.97$ on V2X-PnP. On the InterDrive bucket it scores $83.02$ against $83.99$ over the scenarios the released harness can start. The four it cannot start (see the MDrive harness below) score $2$ to $75$ with our fix, which gives $78.19$ over all $46$.
On the public V2Xverse routes the privileged rule-based expert also falls, from $82.58$ to $74.24$ with the actor deletion of its release ($62.95$ without). Several learned baselines, in contrast, score above their published values (TransFuser $57.59$ against $30.23$, V2X-ViT $53.34$ against $39.35$). The gap of \cod thus reflects the protocol, not the rerun.
Early and late fusion, F-Cooper, V2X-ViT and \cod are the release's configurations of one pipeline: the same planner architecture (MotionNet), each with its own perception and its own planner weights. The no-fusion baseline runs the \cod weights with messages off. Every method receives the route the benchmark gives every agent; \ours reads it as given, most baselines through their own sparse target points. VAD and UniAD run the agents CoLMDriver released for them, with their released weights. The only glue hands their planner the route of the one controlled vehicle, since the agents expect one route per vehicle. The base configurations that their configurations include come from the same release of their code. CoLMDriver and its rule variant run CoLMDriver's released evaluator, agent and configuration on the public V2Xverse runs, with only the server addresses changed. Their language models are served with vLLM 0.8.5 (the release pins no version). The memory budgets fit two servers on one GPU, and one server pair serves three to four simulations. With one controlled vehicle, the negotiation that sets the two apart never runs (no negotiation in any of the $170$ logs). The two are thus one policy, and the paired difference of their driving scores, $+0.47$ ($95\,\%$ interval $[-4.65, +5.55]$), measures the sampling of the vision-language model. CoLMIN runs its own release in the same setup. Its configuration names an action module and a planner configuration that the release does not contain, so it uses the released action module, which holds CoLMIN's method, and the planner configuration of the same checkpoint. Its reflection runs on Qwen2.5-7B-Instruct, since the Qwen2.5-7B weights that its paper names are not released.
On MDrive, which ships no agents for them, WOR, LAV, TransFuser and InterFuser run V2Xverse's released adapters with one instance per vehicle, as on InterDrive. They run through a per-vehicle server that counts a vehicle whose sensors MDrive has already removed as finished and holds a reading missing for one step. On MDrive, TCP runs MDrive's own planner with its released checkpoint (all $286$ tensors load), in MDrive's environment specification with the PyTorch 1.10.1 (CUDA 11.3) wheels instead of conda packages. It averages $77.01$ against the published $76.79$. LMDrive, UniAD and VAD run MDrive's own agents with their released weights (VAD's agent calls MDrive's Bench2Drive controller, which our setup links in). In VAD's Interaction scenario Pedestrian\_Crosswalk/8, all four vehicles collide within $7.6$\,s, and the evaluator hung in its teardown in both attempts without writing a result; a third attempt with the release's RPC timeout completes ($10.51$), so VAD's averages are over all $225$ scenarios.

\noindent\textbf{MDrive harness.}
We run MDrive's released harness with three changes that apply to every method alike and act only where it would leave a scenario without a result (a fourth, for the CoLMDriver family only, follows below).
First, a vehicle that cannot be placed at its start is moved back along its heading by up to $2$\,m, so that the four InterDrive-bucket scenarios above run.
Second, CARLA 0.9.12 sometimes never finishes a frame, typically while a vehicle keeps pushing against a static prop, and the released harness then waits until a watchdog kills the run. A frame that has not finished within $300$\,s now ends the scenario at that frame, scored with what was recorded until then.
Such a run is run once more from the start, as Bench2Drive reruns crashed routes, and is scored at the stall only if it stalls again. Of the $3{,}150$ runs ($14$ method repetitions of $225$ scenarios), $102$ stalled (Defer to Plan $37$, late fusion $36$, \ours $15$, early fusion $9$, F-Cooper $2$, no fusion, V2X-ViT and \cod one each). Of these, $11$ stalled again (late fusion $4$, \ours $3$, Defer to Plan $2$, early fusion and F-Cooper one each). Of the $1{,}350$ runs of TCP, the four camera baselines and the expert, $24$ stalled and $2$ stalled again (TransFuser and LAV, one each).
Of the stalled attempts on the cluster, $62$ sat in MDrive's teardown of a finished vehicle's sensors and $37$ in the world tick itself. The teardown runs a world tick that can hang until the client's RPC timeout. The release sets that timeout to $10$\,s and goes on, while our stall rule had raised it to $300$\,s. The $12$ runs still scored at such a stall (\ours $3$, late fusion $4$, Defer to Plan, early fusion, F-Cooper, LAV and TransFuser one each) were rerun once with the release's timeout and the same seeds. The reruns replace them, and $10$ of the reruns completed. The other $2$ (late fusion and F-Cooper) stalled in the world tick itself, as does one run of Defer to Plan, and stay scored at that stall.
Third, the launcher stops a frozen evaluator after $30$\,min instead of $24$\,h; a run that ends without a result is run once more (further attempts: below).

\noindent\textbf{CoLMDriver, its rule variant, CoLMIN and TCP.}
On InterDrive, CoLMDriver, its rule variant, TCP, VAD and UniAD run CoLMDriver's released evaluator, agents and configurations on all $46$ tests per setting, with only the server addresses changed and, for VAD and UniAD, the interface fixes below. CoLMIN, whose InterDrive rows are published, runs only the two tests shown in \cref{sec:supp_qual}. With the $4{,}096$-token context of CoLMIN's README, its language-model request for the eight vehicles of r44 ($4{,}155$ tokens with the reply) is refused, and its agent retries it without end from $1.75$\,s; for that figure the model is served with $8{,}192$ tokens, and nothing else changes. VAD and UniAD need three interface fixes. VAD's agent imports Bench2DriveZoo's single-route planner and fails with several vehicles; it uses the multi-vehicle route planner of the release's UniAD agent instead. CoLMDriver's sensor interface counted late frames of a finished vehicle's removed sensors toward the frame it waits for, so a running vehicle's camera went missing and most tests of these six-camera agents crashed; it now drops such frames, which changes nothing else. UniAD's runs write every vehicle's record but not the results table that CoLMDriver's runner looks for, so the records decide. Every test then has a result from its first attempt that counts.
The release pins no vLLM version, and its clients set no sampling parameter, so the sampling of the language models depends on the server. From version 0.8 on, vLLM applies the checkpoints' settings (temperature $0.3$ for the vision-language model and $0.7$ for the language model, each with top-$k$ $20$); earlier versions sample with temperature $1$ and no truncation. CoLMDriver's critic asks the language model for a summary of the negotiation and asks again, without limit, until the summary names every vehicle. With the checkpoints' settings, two tests with seven and eight vehicles stayed in this loop for hours without a new frame. Since sampling at temperature $1$ varies the retries more, we serve the models with these earlier defaults on every InterDrive test and ran the tests finished before the change again.
A test is stopped as a livelock when its simulation has finished no frame for $30$\,min while the critic repeats one answer, or when CoLMDriver's own watchdog stops it inside that loop. It is scored at the state reached, from the leaderboard's records or, where the stop left none, with route completion and collisions rebuilt from the recording, and it is not run again. This happened in $8$ of CoLMDriver's $92$ tests, five of them with seven or eight vehicles, and in none of the rule variant's; without them, CoLMDriver scores $77.27$ and $74.85$ (\cref{tab:supp_colm}). In one test without traffic, TCP's agent crashes on the same sensor race in three attempts and completes the fourth.
On MDrive, CoLMDriver and its rule variant run MDrive's own copy of CoLMDriver's agent, which bounds the critic to three retries, with the checkpoints' sampling settings (the default of every vLLM version since March 2025) and MDrive's harness as described above. The agent removes and respawns the roadside sensors of every controlled vehicle every five frames. MDrive's teardown of these sensors runs an extra world tick, which in CARLA 0.9.12 keeps Python's global interpreter lock while it waits for the frame, and with three or more vehicles the runs often stalled in this teardown. For these methods, the teardown sends the same batch and runs the same single tick through the world's own tick call, which releases the lock. Stalls remain, now mostly while the new sensors are spawned: $15$, $47$, $63$ and $62\,\%$ of CoLMDriver's runs with $1$--$2$, $3$--$4$, $5$--$6$ and $7$--$8$ vehicles stalled, and $13$, $46$, $69$ and $80\,\%$ of the rule variant's. Several scenarios stall at the same frame in both runs and on different machines, so a stall follows the scenario. As for every method, a frame unfinished after $300$\,s ends the run, which is run once more and scored at its second stall ($45$ and $56$ of the $225$ scenarios); a run that fails at start-up, before its sensors are warmed up, is run once more and does not count. In two scenarios of the V2X-PnP bucket, CoLMDriver's agent fails at the start in all three attempts (MDrive's launcher reports no route execution after the agent's error in processing the roadside data), while every other method but our CoLMIN adaptation, which uses the same agent, completes them; its averages are over the $223$ scenarios with a result ($68.37$ with these at $0$). MDrive ships no CoLMIN. For \cref{tab:main}, CoLMIN's changes to CoLMDriver (its reflection and its scored candidate critic; the agent file itself is identical) are merged into MDrive's copy of the CoLMDriver agent, served as on InterDrive with its reflection model ($8{,}192$ tokens in the two scenarios whose request exceeds $4{,}096$), and run under the same rules: $14$, $35$, $70$ and $75\,\%$ of its runs stalled, $49$ scenarios were scored at a second stall, and $2$ have no result (the two above). This adaptation scores $74.09$, or $79.04$ without the stall-scored scenarios, $13.9$ and $9.0$ points below \ours on the same scenarios. \Cref{tab:supp_colm} reports CoLMDriver and its rule variant with the stall-scored scenarios and without them; \ours leads in both.
A scenario or test still without a result after its counted attempts (two on MDrive, three on InterDrive) was run again until it had one, at most four times in all (three where every run ended in the same error of the agent); this added $2$ attempts for CoLMDriver, $1$ for its rule variant, $5$ for CoLMIN, $1$ for VAD and $1$ for TCP, and only baselines needed it. These reruns took about $550$ GPU-hours ($94$ on InterDrive, $455$ on MDrive), including the runs that stalled or livelocked.

\begin{table*}[t]
  \centering
  \footnotesize
  \setlength{\tabcolsep}{4pt}
  \begin{tabularx}{\textwidth}{@{}l *{7}{>{\centering\arraybackslash}X}@{}}
    \toprule
    & \multicolumn{2}{c}{InterDrive, no traffic} & \multicolumn{2}{c}{InterDrive, traffic} & \multicolumn{2}{c}{MDrive} & \\
    \cmidrule(lr){2-3}\cmidrule(lr){4-5}\cmidrule(lr){6-7}
    Method & DS$\uparrow$ & SR$\uparrow$ & DS$\uparrow$ & SR$\uparrow$ & DS$\uparrow$ & No coll.$\uparrow$ & Kept \\
    \midrule
    \grouprow{8}{Every test and scenario (as in \cref{tab:main})}
    \quad CoLMDriver & 72.12 & 0.63 & 69.18 & 0.54 & 69.11 & 0.74 & $46$ / $46$ / $223$ \\
    \quad Rule-based$^{\ast\ast}$ & 76.55 & 0.70 & 70.78 & 0.57 & 73.23 & 0.81 & $46$ / $46$ / $225$ \\
    \quad \ours & \sota{97.79} & \sota{0.89} & \sota{90.70} & \sota{0.75} & \sota{88.16} & \sota{0.95} & $46$ / $46$ / $225$ \\
    \grouprow{8}{Without livelocked tests and scenarios scored at a second stall}
    \quad CoLMDriver & 77.27 & 0.69 & 74.85 & 0.60 & 72.07 & 0.71 & $42$ / $42$ / $178$ \\
    \quad \ours, same tests & \sota{97.91} & \sota{0.90} & \sota{90.22} & \sota{0.75} & \sota{87.04} & \sota{0.95} & $42$ / $42$ / $178$ \\
    \quad Rule-based$^{\ast\ast}$ & 76.55 & 0.70 & 70.78 & 0.57 & 79.36 & 0.80 & $46$ / $46$ / $169$ \\
    \quad \ours, same tests & \sota{97.79} & \sota{0.89} & \sota{90.70} & \sota{0.75} & \sota{88.49} & \sota{0.96} & $46$ / $46$ / $169$ \\
    \bottomrule
  \end{tabularx}
  \caption{\textbf{CoLMDriver and its rule variant, rerun on InterDrive and MDrive.} Our reruns of the releases (\cref{sec:supp_setup}): InterDrive with the current result script and the language models sampling with vLLM's earlier defaults; MDrive with MDrive's copy of the agent. A livelocked test (CoLMDriver's critic repeating one answer; the rule variant has none) is scored at the state reached, and a scenario that stalls twice at its second stall. The lower block leaves these out and compares \ours on the same tests and scenarios. Kept: InterDrive tests without and with traffic, MDrive scenarios. $^{\ast\ast}$CoLMDriver negotiating by rules. Paired over the kept MDrive scenarios, \ours leads CoLMDriver by $18.9$ ($[+15.1, +22.9]$) and $15.0$ ($[+10.8, +19.4]$) DS and the rule variant by $14.9$ and $9.1$ ($[+4.9, +13.7]$).}
  \label{tab:supp_colm}
\end{table*}

\subsection{Implementation}\label{sec:implementation}

\noindent\textbf{Training.}
The reported model ($18.0$\,M parameters with the $0.27$\,M of the order gate's precedence network, LiDAR only) is the last stage of the chain of \cref{sec:training}.
This stage trains for $2$ epochs at batch size $4$ on the released V2Xverse recordings: Town01, 02, 03, 04 and 06 for training ($31{,}755$ samples from $68$ routes), Town07 and Town10 for validation.
Town05, the town of all V2Xverse runs, is in neither set; InterDrive and MDrive also test in training towns (Town06; MDrive also Town01 to 03).
We use AdamW~\cite{loshchilov2019adamw} (learning rate $10^{-4}$ for all parameters, weight decay $0.01$, $500$ warm-up iterations, cosine decay, gradient clipping at norm $10$, bfloat16 autocast). Training takes about three hours on one NVIDIA RTX PRO 6000 GPU.
The earlier stages use the same optimizer and schedule, with learning rate $2\cdot10^{-4}$ for the per-agent network and the model without the map and $10^{-4}$ for the route stage. In the two stages after the first, the perception weights loaded from it take a quarter of that rate. The last stage also keeps the first stage's loss on each vehicle's own outputs (weight $1$).
The conflict map is rasterised without gradients, but planning losses still reach the perception through the planner's attention to the fused instances, only not through the map. The order gate's precedence network is trained separately on rule-labelled pairs and merged into the checkpoint. Training seed $0$ (the default) was the reported model as soon as the recipe was fixed; seeds $1$ and $2$ were trained afterwards (\cref{tab:supp_seeds}), and the ablations retrain with the default seed. Besides the right-of-way rule of \cref{sec:supervision}, the virtual-partner labels apply a lane-change rule. An ego that changes lanes yields to a vehicle in the target lane that is not behind it (data/virtual\_partner.py in the release).
A sample stays clean with probability $0.3$. Otherwise, the ego LiDAR is corrupted with probability $0.4$ (sector occlusion, point dropout, beam loss, range limit, noise or fog; five levels). In a sample that is not clean, each partner independently draws a LiDAR corruption (probability $0.4$), a delay of up to $0.6$\,s ($0.5$) and a message loss ($0.3$). It also draws pose noise of standard deviation up to $1.0$\,m and $1.0^\circ$ ($0.5$).

\noindent\textbf{Control at test time.}
The plan's waypoints go to the benchmarks' PID controller~\cite{liu2025codriving}. Its speed target is the plan's mean speed between $0.2$ and $0.8$\,s times the speed factor of the agent configuration ($1.0$ or $1.6$). It steers by the bearing of the mean of the waypoints at $0.8$ and $1.0$\,s instead of a point on the global route.
The controller's forced push after standing is switched off (see below).
Every reported run records its switches (\texttt{jd\_run.json}), and we checked all of them. The V2Xverse runs of the final seeds switch the push off and steer by the waypoints, with the benchmark's partner selection and an ideal channel. The InterDrive runs of the one network also fuse up to three sources, their own roadside unit first and then the nearest partners, and read up to three plan-only partners heard within the $70$\,m range beyond them. They use the order gate with decoding, and their routes continue straight beyond their end (at test time only). The MDrive runs do the same. No reported run reads a stop line. The controller's speed limit is that of each benchmark's agent configuration: on V2Xverse the configuration of each part ($5$ or $8$\,m/s), on InterDrive $8$\,m/s as in CoLMDriver's configuration~\cite{liu2025colmdriver}, on MDrive $5$\,m/s. Baseline reruns keep their own controller and configuration.
All our runs use the simulator without a renderer, which leaves the LiDAR unchanged, and the network in single precision on the CPU. On V2Xverse, with one controlled vehicle, the order gate has nothing to decide, and those runs load the network without its $0.27$\,M gate parameters; the other $17.76$\,M are bit-identical to the one network's.

\noindent\textbf{Training seeds.}
\begin{table}[t]
  \centering
  \footnotesize
  \setlength{\tabcolsep}{3pt}
  \begin{tabularx}{\linewidth}{@{}>{\raggedright\arraybackslash}X cccc@{}}
    \toprule
    & InterDrive & \multicolumn{3}{c}{V2Xverse, repetition 3} \\
    \cmidrule(lr){2-2}\cmidrule(l){3-5}
    Training seed & DS & DS & IS & blocked \\
    \midrule
    \rowcolor{oursbg}
    0 (reported) & 97.79 & 72.24 & 0.787 & 15 \\
    1 & 95.78 & 65.41 & 0.802 & 38 \\
    2 & 94.89 & 71.72 & 0.859 & 35 \\
    \midrule
    Mean & 96.15 & 69.79 & 0.816 & 29 \\
    \bottomrule
  \end{tabularx}
  \caption{\textbf{Training seeds.} The final recipe trained with three seeds (the planner; the order gate of seed $0$ is shared). InterDrive without traffic ($46$ tests), V2Xverse repetition $3$ ($85$ runs). Blocked: runs per $100$ ended by the blocked test.}
  \label{tab:supp_seeds}
\end{table}

Two further seeds of the final recipe (\cref{tab:supp_seeds}) score $95.78$ and $94.89$ on InterDrive without traffic, with paired differences to the reported model of $-2.0$ ($[-4.4, 0.0]$) and $-2.9$ ($[-5.4, -0.9]$). On V2Xverse repetition $3$, they score $65.41$ and $71.72$, with paired differences of $-6.8$ ($[-12.9, -0.8]$) and $-0.5$ ($[-7.1, +5.9]$).
They lose route completion, not safety. They wait longer, at junctions and near pedestrians, and are blocked in $38$ and $35$ of $100$ V2Xverse runs against $15$ for the reported model, with the same or a higher infraction score.
Every seed stays above every baseline of \cref{tab:main} on both benchmarks.

\noindent\textbf{The forced push.}
\begin{table}[t]
  \centering
  \footnotesize
  \setlength{\tabcolsep}{2.5pt}
  \begin{tabularx}{\linewidth}{@{}>{\raggedright\arraybackslash}X cccc@{}}
    \toprule
    & runs & \makecell{with a\\push} & \makecell{vehicle collisions\\${\le}5$\,s after a push} & \makecell{pushed runs\\completed} \\
    \midrule
    \grouprow{5}{V2Xverse, baselines as released (the runs of \cref{tab:main})}
    No fusion$^{\ast}$          & 255 & 14\,\% & 2 of 226  & 6\,\% \\
    Early fusion                & 85  & 16\,\% & 2 of 80   & 14\,\% \\
    Late fusion                 & 85  & 19\,\% & 6 of 55   & 31\,\% \\
    F-Cooper                    & 85  & 14\,\% & 0 of 71   & 0\,\% \\
    V2X-ViT                     & 85  & 16\,\% & 0 of 61   & 0\,\% \\
    \cod                        & 255 & 13\,\% & 2 of 240  & 9\,\% \\
    Defer to Plan$^{\ddagger}$  & 255 & 31\,\% & 64 of 240 & 68\,\% \\
    \grouprow{5}{\ours with the push switched on}
    V2Xverse                    & 85  & 32\,\% & 21 of 26  & 48\,\% \\
    InterDrive, traffic         & 46  & 20\,\% & 13 of 17  & 16\,\% \\
    InterDrive, no traffic      & 46  & 7\,\%  & 0 of 0    & 80\,\% \\
    \bottomrule
  \end{tabularx}
  \caption{\textbf{The benchmark's forced push in the logs.} Runs (InterDrive: tests) with at least one push. Vehicle collisions within $5$\,s after a push of the pushed vehicle, of all vehicle collisions. Share of pushed runs (InterDrive: pushed vehicles) that complete the route. \ours with the push on: repetition $3$, all runs and tests.}
  \label{tab:supp_push}
\end{table}

The benchmark controller, taken from InterFuser~\cite{shao2022interfuser}, counts control steps below $0.2$\,m/s. After $100$ of them ($20$\,s; the agents act at $5$\,Hz), it drives with throttle $0.8$ for $12$ steps ($2.4$\,s), whatever the plan says. InterFuser's own threshold is $1{,}200$ steps.
It rescues policies that stall, and it decides when a waiting vehicle goes, the decision \ours learns. Every \ours run of the paper therefore switches it off, and every baseline keeps it, as released.
A push shows in the logs as throttle $0.8$, above the controller's own limit of $0.75$ (\cref{tab:supp_push}).
On V2Xverse it fires in $13$ to $31\,\%$ of the baselines' runs. It completes some of their routes, and $27\,\%$ of Defer to Plan's vehicle collisions follow a push within $5$\,s.
On InterDrive the baselines stand that long less often (a push in $2$ to $22\,\%$ of the tests), since they collide earlier.
Switched on for \ours, the push fires in $3$ of the $46$ tests without traffic, where four of the five vehicles it pushes then arrive ($+0.8$, $[-0.2, +2.0]$). With traffic, the push changes the driving score by $-3.7$ ($[-8.4, +1.2]$), and on V2Xverse by $-4.5$ ($[-9.9, +0.8]$), where the infraction score changes by $-0.08$ ($[-0.15, -0.02]$). In both settings, it fires when a vehicle yields and precedes $34$ of its $43$ vehicle collisions. On MDrive it changes the driving score by $+1.0$ ($[-0.2, +2.4]$) and precedes $68$ of its $89$ vehicle collisions. The push thus trades blocked runs for collisions.
In InterDrive r23 with traffic, for example, a yielding vehicle is pushed into the crossing traffic and collides four times (DS $7.0$ against $49.3$ without the push). In r8 the push frees a vehicle that waited for good ($100$ against $44.0$).

\noindent\textbf{Conventions.}
Where the controller's speed limit is $8$ instead of $5$\,m/s (V2Xverse parts 1s and 4s, InterDrive), it drives at $1.6$ times the planned speed. The network receives its own speed and history divided by this factor, so its plans are in these units, while its detections and forecasts are metric. The vehicle then reaches the position of row $k$ of a candidate at about $t_k/1.6$, while the forecasts and partner plans in the map stand at $t_k$. We have not tested a map whose time axis is rescaled accordingly. In closed loop, a route that ends within reach is continued straight along its last direction (training repeats its last point). Otherwise every object near the goal would be projected onto the route, and the planner would stop short of it. The waypoints go to the controller in metres in the vehicle frame, without the lateral rescaling of \cod's waypoint space. The network has no input for traffic lights, and stops for a red light alone are relabelled as distance keeping (\cref{sec:supervision}). V2Xverse holds every light at green during evaluation, and across all reported runs the only red-light infractions are on MDrive, in two of its V2X-PnP log replays (once per repetition).

\noindent\textbf{Notes on \cref{tab:main}.}
On InterDrive and MDrive, the reruns of no fusion, early and late fusion, F-Cooper, V2X-ViT and Defer to Plan use repetition $0$, the benchmarks' default seed, as does the rerun of TCP on MDrive. On MDrive, the two strongest in driving score, late fusion and Defer to Plan, also run repetitions $1$ and $2$. On V2Xverse, every rerun runs in its release configuration (rendered, with the release's roadside units) on repetition $0$, and on repetitions $0$ to $2$ for no fusion, \cod and Defer to Plan (\cref{tab:supp_v2xverse}).
The no-fusion row runs the release's no-fusion configuration (\cod weights, messages off); its reported value comes from a separately trained model.
The expert row runs each benchmark's own privileged rule-based expert (V2Xverse's, and the AutoPilots of CoLMDriver on InterDrive and of MDrive) through one wrapper. The wrapper keeps the expert's control law, speeds and decisions and drops its sensors and data collection. The expert does not delete the actors in its way (MDrive's own planner for its AutoPilot stops at its first saved frame on CARLA 0.9.12). It reaches $62.95$ on V2Xverse, $83.89$ and $78.42$ on InterDrive without and with traffic, and $67.52$ on MDrive, below \ours on each. With the actor deletion of V2Xverse's release, V2Xverse's expert reaches $74.24$ on repetition $0$, above \ours (\cref{tab:supp_v2xverse}); on the seeds of \ours it scores $53.93$, below it (\cref{sec:supp_v2xverse}). On MDrive's V2X-PnP bucket it scores $29.57$, against $90.54$ on the InterDrive and $82.45$ on the Interaction bucket. The logged vehicles there are replayed by setting their positions, so the simulator reports almost no velocity for them, and the expert, which predicts conflicts from velocities, does not see them coming.
Four scenarios of MDrive's InterDrive bucket (r25, r43, r44, r45) abort in the released harness for every method. The requested Lincoln is missing in CARLA 0.9.12, and the Tesla that replaces it overlaps a neighbouring vehicle at the start. With that vehicle moved back by at most $2$\,m (\cref{sec:protocol}), the four run for every method; without them, \ours averages $94.68$ on the InterDrive bucket instead of $92.18$.
Our InterDrive scores use the benchmark's current result script. CoLMDriver's tables used an earlier version (see also~\cite{colmdriver_issue6}), while CoLMIN's own rows use the current one. Under the earlier version, our rerun of the released \cod reproduces its published value ($74.46$ against $74.13$, \cref{sec:supp_v2xverse}). Under it, our rerun of LMDrive scores $57.85$ and $54.56$ without and with traffic, against the published $48.83$ and $49.95$. Its route completion is $12$ and $6$ points higher than published, and its infraction score and success rate are within $0.03$. Under the same version, \ours scores $97.79$ and $90.85$, and the reruns score $0.6$ to $12.6$ points more than in \cref{tab:main} (Defer to Plan $76.76$ and $74.73$; VAD and UniAD, which collide most, $10.8$ to $12.6$). The comparison does not depend on the script. CoLMIN's published values by scenario type (\cref{tab:categories}) average to $78.66$ with traffic, while \cref{tab:main} lists its published total ($78.93$).
Our reimplementation of Defer to Plan, trained on the released data, lowers the open-loop error of \cod on the official Town05 samples by $13\,\%$ (ADE $0.543$ against $0.625$\,m). This is more than the $3.4\,\%$ its paper reports, so its closed-loop scores are not those of a weak reimplementation.
The open-loop error of the released \cod matches the published one ($0.625$ against $0.619$\,m): its cooperative path runs as released.

\section{V2Xverse: every metric of every rerun}\label{sec:supp_v2xverse}
\begin{table*}[t]
  \centering
  \footnotesize
  \setlength{\tabcolsep}{2pt}
  \begin{tabularx}{\textwidth}{@{}l *{11}{>{\centering\arraybackslash}X}@{}}
    \toprule
    & & & & & & \multicolumn{4}{c}{runs per 100 with} & \\
    \cmidrule(lr){7-10}
    Method & DS$\uparrow$ & \pub{rep.\,DS} & SR$\uparrow$ & IS$\uparrow$ & RC$\uparrow$ & \makecell{vehicle\\collision$\downarrow$} & \makecell{pedestrian\\collision$\downarrow$} & \makecell{layout\\collision$\downarrow$} & blocked$\downarrow$ & runs \\
    \midrule
    WOR~\cite{chen2021wor} & 55.85 & \pub{12.05} & 0.27 & 0.710 & 80.4 & 31 & 27 & \best{0} & 21 & 85 \\
    LAV~\cite{chen2022lav} & 46.55 & \pub{38.27} & 0.18 & 0.739 & 68.0 & 24 & 26 & 6 & 13 & 85 \\
    TransFuser~\cite{chitta2023transfuser} & 57.59 & \pub{30.23} & 0.28 & 0.706 & 84.9 & 24 & 32 & 4 & 29 & 85 \\
    InterFuser~\cite{shao2022interfuser} & 45.28 & \pub{40.31} & 0.13 & 0.487 & 91.9 & 64 & 28 & 12 & 15 & 85 \\
    TCP~\cite{wu2022tcp} & 53.53 & \pub{47.48} & 0.18 & 0.805 & 68.4 & 16 & 19 & 2 & 65 & 85 \\
    VAD~\cite{jiang2023vad} & 11.92 & -- & 0.00 & 0.632 & 32.6 & 35 & \second{8} & 29 & 81 & 85 \\
    UniAD~\cite{hu2023uniad} & 16.81 & -- & 0.02 & 0.417 & 40.2 & 48 & 13 & 74 & 80 & 85 \\
    LMDrive~\cite{shao2024lmdrive} & 29.73 & -- & 0.05 & \best{0.829} & 36.1 & \second{9} & \best{6} & 19 & 40 & 85 \\
    \midrule
    No fusion$^{\ast}$~\cite{liu2025codriving} & 56.37 & \pub{33.81} & 0.29 & 0.588 & 93.7 & 42 & 31 & 6 & 14 & 255 \\
    Early fusion~\cite{liu2025codriving} & 57.20 & \pub{59.12} & \second{0.32} & 0.592 & 94.7 & 48 & 28 & 8 & 14 & 85 \\
    Late fusion~\cite{liu2025codriving} & \second{59.09} & \pub{52.40} & 0.31 & 0.609 & \best{96.0} & 41 & 33 & 5 & \second{12} & 85 \\
    F-Cooper~\cite{chen2019fcooper} & 51.45 & \pub{44.00} & 0.27 & 0.557 & 91.1 & 46 & 40 & 7 & 18 & 85 \\
    V2X-ViT~\cite{xu2022v2xvit} & 53.34 & \pub{39.35} & 0.19 & 0.602 & 84.6 & 40 & 34 & 5 & 15 & 85 \\
    \cod~\cite{liu2025codriving} & 56.59 & \pub{77.15} & 0.30 & 0.592 & 94.6 & 43 & 31 & 6 & 13 & 255 \\
    Defer to Plan$^{\ddagger}$~\cite{li2026defer} & 56.50 & \pub{79.72} & 0.31 & 0.584 & \second{95.6} & 38 & 38 & 7 & \best{10} & 255 \\
    Rule-based$^{\ast\ast}$~\cite{liu2025colmdriver} & 40.32 & -- & 0.19 & 0.434 & 93.2 & 66 & 41 & 12 & 14 & 85 \\
    CoLMDriver~\cite{liu2025colmdriver} & 40.79 & -- & 0.18 & 0.432 & 93.2 & 69 & 42 & 14 & 15 & 85 \\
    CoLMIN~\cite{colmin2026} & 40.15 & -- & 0.18 & 0.423 & 94.3 & 65 & 45 & 18 & \second{12} & 85 \\
    \midrule
    \gr{Expert, rules (privileged)} & \gr{62.95} & \gr{--} & \gr{0.39} & \gr{0.788} & \gr{79.4} & \gr{9} & \gr{32} & \gr{2} & \gr{42} & \gr{85} \\
    \gr{Expert, with actor deletion (privileged)} & \gr{74.24} & \gr{82.58} & \gr{0.53} & \gr{0.747} & \gr{99.6} & \gr{11} & \gr{38} & \gr{2} & \gr{0} & \gr{85} \\
    \midrule
    \rowcolor{oursbg}
    \textbf{\ours (ours)} & \sota{70.00} & -- & \sota{0.42} & \second{0.825} & 85.2 & \sota{6} & 23 & \second{1} & 29 & 510 \\
    \bottomrule
  \end{tabularx}
  \caption{\textbf{V2Xverse public protocol, all components of the driving score.} Means over runs (repetitions of a route averaged first), as in \cref{tab:main}. Runs per 100 with at least one collision of the kind, or ending by the blocked test. $^{\ast}$, $^{\ast\ast}$, $^{\ddagger}$: as in \cref{tab:main}. With one controlled vehicle, the negotiation of CoLMDriver and CoLMIN never runs, so these three rows are one policy with different samples of its vision-language model. The privileged experts drive on ground truth; the second removes actors that block it for long (the benchmark's own data-collection expert). \ours: repetitions 3 to 8 of the final seeds. Every rerun has all $85$ runs per repetition; the five slowest runs of V2X-ViT needed a longer wall-clock limit of our harness (8 instead of 2.5\,h). \textbf{Bold}: best, \underline{underline}: second, among the values not in gray.}
  \label{tab:supp_v2xverse}
\end{table*}

\begin{figure*}[t]
  \centering
  \includegraphics[width=\textwidth]{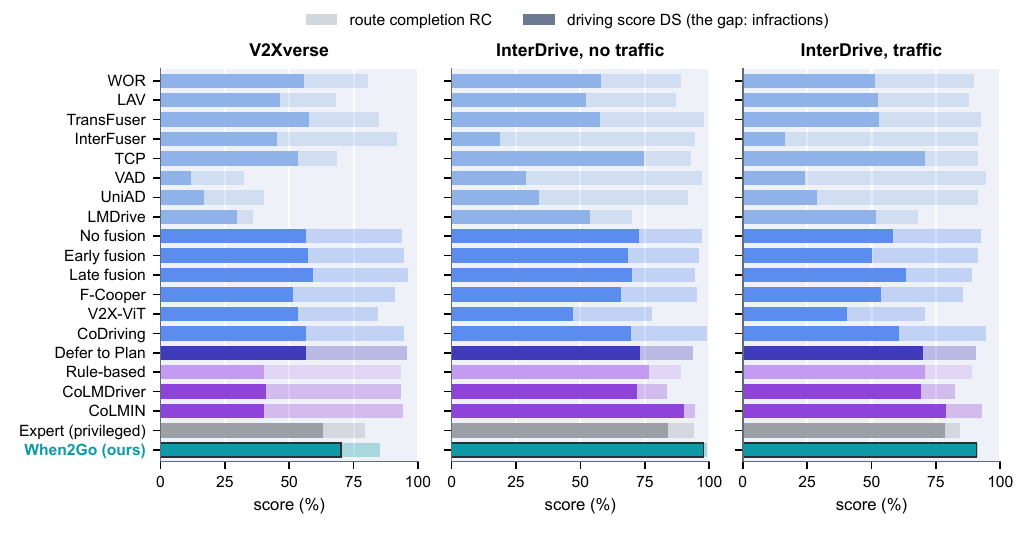}
  \caption{\textbf{Route completion and driving score.} Light bars: route completion (RC, \%); dark bars: driving score of \cref{tab:supp_v2xverse,tab:supp_interdrive} (\ours: the repetitions of \cref{tab:main}; CoLMIN on InterDrive: published, same result script). The gap between the two is the cost of infractions, almost all of them collisions. On InterDrive, \ours loses the least to infractions ($1.2$ and $0.8$ points).}
  \label{fig:supp_rc_is}
\end{figure*}

\Cref{tab:supp_v2xverse} lists the components of the driving score for all methods of the V2Xverse block of \cref{tab:main}.
The driving score of a run is its route completion times its infraction score, so the two trade against each other. A policy that drives on through conflicts completes more of its route and collides more often (\cref{fig:supp_rc_is}: the gap between route completion and driving score).

The rerun fusion baselines complete $85$ to $96\,\%$ of their routes and collide with a vehicle in $38$ to $48$ of $100$ runs. \ours collides with a vehicle in $6$ of $100$ runs and waits more often, which costs route completion when a wait outlasts the benchmark's blocked test ($30$\,s below $0.5$\,m/s).
Among the baselines, LMDrive and TCP have the highest infraction scores ($0.829$ and $0.805$) because they drive little: LMDrive completes $36\,\%$ of its routes, and TCP stands still in $65$ of $100$ runs.

\noindent\textbf{The reruns are faithful.}
Every baseline runs with its released checkpoint (Defer to Plan: our reimplementation) and the benchmark's own agent code, controller and configuration (CoLMDriver, its rule variant, LMDrive, VAD and UniAD: CoLMDriver's released agents), except where \cref{sec:supp_setup} states otherwise. Only the process layout differs (the CARLA client in Python~3.7, the networks in a separate server).
Run on the same route and seed (part 1s, route 13), the unmodified upstream agent and our client--server pair drive the same. Their speeds agree within $0.01$\,m/s over the first $200$ control steps, including the same collision with a cyclist. After that, a waypoint difference of millimetres during a hard brake separates the runs, as it separates two runs of the same agent.
In open loop, the released \cod checkpoint reaches $0.625$ / $1.426$\,m ADE / FDE on the official Town05 samples, against $0.619$ / $1.413$ published.
The gap to the published closed-loop scores comes from the protocol. The benchmark's privileged rule-based expert, which has no network, drops from $82.58$ on the unreleased routes to $74.24$ on the public ones, with the actor deletion of its release in both. Third parties report $55$ to $57$ for the released \cod on the public routes~\cite{v2xverse_issue44,v2xverse_issue48}, as we measure ($56.59$). Leaving routes out does not explain the published $77.15$ on $67$ routes: the best $67$ of our $85$ route means of \cod average $66.3$ (of the best repetition alone, $73.0$).
All methods are therefore compared on the same public routes.
On InterDrive, the released \cod run through the same harness on the reported seeds scores $74.46$ DS without traffic with the benchmark's earlier result script (commit e603368), against $74.13$ reported~\cite{liu2025colmdriver}. The current script, which scores all our runs, gives $69.59$. With background traffic, which is not deterministic, the earlier and the current script give $67.97$ and $60.59$, against $64.50$ reported.
Rerun with that earlier script, CoLMDriver's rule variant scores $78.38$ without traffic, as reported, and $76.40$ with it ($69.71$ reported). CoLMDriver itself scores $77.94$ and $76.32$ on the $42$ tests per setting it finishes, against $88.53$ and $77.09$ reported. The two share all code except the negotiation by the language model, whose sampling the release does not fix (no server version, no sampling parameter, \cref{sec:supp_setup}), and in $4$ tests per setting CoLMDriver's critic retries without end; \cref{tab:main} scores these at the state reached ($72.12$ and $69.18$ under the current script).
\Cref{tab:main} lists the reported values.

\noindent\textbf{What the public protocol measures.}
Each run controls one vehicle among background traffic, so V2Xverse measures what cooperative perception does for a single planner; agreement between controlled vehicles is tested on InterDrive and MDrive.
A run ends as blocked after $30$\,s below $0.5$\,m/s, which also ends runs in which a vehicle waits for a reason. The benchmark's privileged rule-based expert is blocked in $42$ of $100$ runs (\cref{tab:supp_v2xverse}).
Pedestrians are released when the vehicle reaches a trigger point, and not every such encounter can be avoided. The same expert hits a pedestrian in $32$ of $100$ runs, and the expert that deletes blocking actors hits one in $38$. \ours hits one in $23$, fewer than every rerun fusion baseline ($28$ to $40$).
With one repetition resolving about $7$ points, we compare methods paired over the same routes and seeds (\cref{sec:supp_setup}).

\noindent\textbf{The same seeds.}
On the seeds of the \ours runs (repetitions $3$ to $8$), the strongest baselines score lower than in \cref{tab:main}: TransFuser, rendered with its cameras, scores $57.25$. In the setup of \ours (no rendering, LiDAR-only roadside units), late fusion scores $49.16$, Defer to Plan $48.30$, \cod $47.22$ and early fusion $45.76$. On repetitions $3$ to $5$, V2X-ViT scores $53.99$, WOR $49.95$, F-Cooper $49.83$, no fusion $48.07$, TCP $45.51$, LAV $44.77$ and InterFuser $43.92$ (the camera agents rendered). \ours leads each of these twelve by $12.8$ to $24.4$ points, paired over routes and repetitions, with every interval above $+8$. It also leads the privileged rule-based expert on the same seeds, without its actor deletion ($49.87$, $+18.4$, $[+13.3, +23.4]$) and with it ($53.93$, $+14.4$, $[+8.6, +20.2]$; repetitions $3$ to $5$). With the deletion, the expert completes $97.8\,\%$ of its routes but hits a pedestrian in $56$ of $100$ runs, $443$ times in all; $30\,\%$ of these hits repeat a walker that it has already hit and pushes along the road.
Part of the difference is the seeds: in one setup, \cod scores $52.1$ on repetitions $0$ and $2$ and $45.5$ on repetitions $3$ to $5$. Part may be the setup: late fusion's repetition $0$, rerun in it, scores $52.06$ against $59.09$ (paired $-7.0$, $[-14.4, +0.2]$).
\Cref{tab:main} reports every baseline in its release configuration, where the baselines score higher, so its margins are conservative.

\section{InterDrive and MDrive: every metric}\label{sec:supp_id_md}
\begin{table*}[t]
  \centering
  \footnotesize
  \setlength{\tabcolsep}{3pt}
  \begin{tabularx}{\textwidth}{@{}l *{8}{>{\centering\arraybackslash}X}@{}}
    \toprule
    & \multicolumn{4}{c}{InterDrive, no traffic} & \multicolumn{4}{c}{InterDrive, with traffic} \\
    \cmidrule(lr){2-5}\cmidrule(l){6-9}
    Method & DS$\uparrow$ & SR$\uparrow$ & IS$\uparrow$ & RC$\uparrow$ & DS$\uparrow$ & SR$\uparrow$ & IS$\uparrow$ & RC$\uparrow$ \\
    \midrule
    \grouprow{9}{Published}
    TCP~\cite{wu2022tcp} & \pub{73.68} & \pub{0.50} & \pub{0.82} & \pub{90.54} & \pub{74.18} & \pub{0.48} & \pub{0.82} & \pub{91.21} \\
    VAD~\cite{jiang2023vad} & \pub{25.37} & \pub{0.02} & \pub{0.33} & \pub{75.00} & \pub{25.18} & \pub{0.02} & \pub{0.31} & \pub{75.66} \\
    UniAD~\cite{hu2023uniad} & \pub{35.17} & \pub{0.11} & \pub{0.38} & \pub{88.30} & \pub{37.13} & \pub{0.11} & \pub{0.41} & \pub{88.71} \\
    LMDrive~\cite{shao2024lmdrive} & \pub{48.83} & \pub{0.20} & \pub{0.85} & \pub{58.02} & \pub{49.95} & \pub{0.13} & \pub{0.84} & \pub{61.61} \\
    \cod~\cite{liu2025codriving} & \pub{74.13} & \pub{0.57} & \pub{0.76} & \pub{96.31} & \pub{64.50} & \pub{0.47} & \pub{0.67} & \pub{93.58} \\
    Rule-based$^{\ast\ast}$~\cite{liu2025colmdriver} & \pub{78.38} & \pub{0.72} & \pub{0.80} & \pub{91.85} & \pub{69.71} & \pub{0.57} & \pub{0.75} & \pub{87.35} \\
    CoLMDriver~\cite{liu2025colmdriver} & \pub{88.53} & \pub{0.80} & \pub{0.90} & \pub{94.05} & \pub{77.09} & \pub{0.63} & \pub{0.80} & \pub{92.02} \\
    CoLMIN~\cite{colmin2026} & \second{90.12} & \second{0.81} & \second{0.92} & 94.36 & \second{78.93} & \second{0.67} & \second{0.80} & 92.79 \\
    \midrule
    \grouprow{9}{Our reruns}
    WOR & 58.16 & 0.28 & 0.65 & 88.88 & 51.26 & 0.24 & 0.56 & 89.83 \\
    LAV & 52.07 & 0.30 & 0.57 & 87.03 & 52.61 & 0.30 & 0.58 & 87.98 \\
    TransFuser & 57.76 & 0.50 & 0.58 & 98.12 & 53.05 & 0.37 & 0.56 & 92.56 \\
    InterFuser & 18.91 & 0.09 & 0.20 & 94.47 & 16.58 & 0.07 & 0.18 & 91.20 \\
    LMDrive & 53.89 & 0.20 & 0.79 & 69.98 & 51.58 & 0.15 & 0.78 & 68.01 \\
    TCP & 74.58 & 0.59 & 0.80 & 93.11 & 70.91 & 0.46 & 0.78 & 91.22 \\
    VAD & 28.78 & 0.11 & 0.29 & 97.40 & 24.28 & 0.09 & 0.25 & \best{94.59} \\
    UniAD & 33.95 & 0.11 & 0.35 & 91.69 & 29.02 & 0.11 & 0.30 & 91.45 \\
    No fusion$^{\ast}$ & 72.77 & 0.65 & 0.73 & 97.32 & 58.23 & 0.48 & 0.60 & 92.65 \\
    Early fusion & 68.32 & 0.59 & 0.70 & 96.08 & 50.33 & 0.37 & 0.53 & 91.15 \\
    Late fusion & 70.13 & 0.54 & 0.72 & 94.44 & 63.50 & 0.46 & 0.67 & 89.02 \\
    F-Cooper & 65.59 & 0.52 & 0.67 & 95.16 & 53.72 & 0.28 & 0.59 & 85.47 \\
    V2X-ViT & 47.19 & 0.02 & 0.57 & 77.78 & 40.40 & 0.00 & 0.53 & 70.82 \\
    \cod & 69.59 & 0.59 & 0.70 & \best{99.09} & 60.59 & 0.48 & 0.62 & \second{94.30} \\
    Defer to Plan$^{\ddagger}$ & 73.10 & 0.59 & 0.76 & 93.81 & 69.87 & 0.48 & 0.75 & 90.57 \\
    Rule-based$^{\ast\ast}$ & 76.55 & 0.70 & 0.80 & 89.10 & 70.78 & 0.57 & 0.75 & 88.84 \\
    CoLMDriver & 72.12 & 0.63 & 0.81 & 83.56 & 69.18 & 0.54 & 0.75 & 82.29 \\
    \gr{Expert, rules (privileged)} & \gr{83.89} & \gr{0.67} & \gr{0.89} & \gr{94.09} & \gr{78.42} & \gr{0.48} & \gr{0.94} & \gr{84.29} \\
    \midrule
    \rowcolor{oursbg}
    \textbf{\ours (ours)} & \sota{97.79} & \sota{0.89} & \sota{0.99} & \second{98.99} & \sota{90.70} & \sota{0.75} & \sota{0.99} & 91.47 \\
    \bottomrule
  \end{tabularx}
  \caption{\textbf{InterDrive: all components of the driving score.} Route completion (RC, \%), infraction score (IS) and success rate (SR) behind \cref{tab:main} and \cref{fig:teaser}. Published values as reported~\cite{liu2025colmdriver,colmin2026}. \pub{gray}: CoLMDriver's tables, scored with the benchmark's earlier result script, which leaves the first vehicle's collisions out. CoLMIN, our reruns (released checkpoints, repetition $0$) and \ours are scored with the current one. WOR, LAV, TransFuser and InterFuser run V2Xverse's released adapters, one instance per vehicle; as in its V2Xverse row, InterFuser's adapter drives the predicted waypoints with V2Xverse's controller instead of InterFuser's safety-aware controller. TCP, VAD, UniAD, CoLMDriver and its rule variant run CoLMDriver's released agents (VAD and UniAD with interface fixes); CoLMDriver's livelocked tests (four per setting) are scored at the state reached (\cref{sec:supp_setup}). \gr{Expert}: CoLMDriver's privileged rule-based AutoPilot, run as on V2Xverse (\cref{sec:supp_setup}). $^{\ast}$, $^{\ddagger}$: as in \cref{tab:main}. \textbf{Bold}: best, \underline{underline}: second, among the values not in gray.}
  \label{tab:supp_interdrive}
\end{table*}

\begin{table*}[t]
  \centering
  \scriptsize
  \setlength{\tabcolsep}{2.2pt}
  \begin{tabularx}{\textwidth}{@{}l *{12}{>{\centering\arraybackslash}X}@{}}
    \toprule
    & \multicolumn{3}{c}{InterDrive (ID)} & \multicolumn{3}{c}{Interaction (Int.)} & \multicolumn{3}{c}{V2X-PnP (PnP)} & \multicolumn{3}{c}{Avg} \\
    \cmidrule(lr){2-4}\cmidrule(lr){5-7}\cmidrule(lr){8-10}\cmidrule(l){11-13}
    Method & No coll.$\uparrow$ & DS$\uparrow$ & SR$\uparrow$ & No coll.$\uparrow$ & DS$\uparrow$ & SR$\uparrow$ & No coll.$\uparrow$ & DS$\uparrow$ & SR$\uparrow$ & No coll.$\uparrow$ & DS$\uparrow$ & SR$\uparrow$ \\
    \midrule
    TCP~\cite{wu2022tcp} & -- & \pub{83.55} & \pub{67.80} & -- & \pub{75.80} & \pub{55.25} & -- & \pub{71.02} & \pub{58.70} & -- & \pub{76.79} & \pub{60.58} \\
    UniAD~\cite{hu2023uniad} & -- & \pub{57.48} & \pub{26.40} & -- & \pub{53.08} & \pub{31.10} & -- & \pub{72.55} & \pub{68.48} & -- & \pub{61.04} & \pub{41.99} \\
    VAD~\cite{jiang2023vad} & -- & \pub{57.49} & \pub{23.30} & -- & \pub{48.83} & \pub{27.14} & -- & \pub{67.20} & \pub{66.30} & -- & \pub{57.84} & \pub{38.91} \\
    LMDrive~\cite{shao2024lmdrive} & -- & \pub{56.74} & \pub{27.10} & -- & \pub{57.45} & \pub{29.08} & -- & \pub{71.56} & \pub{69.57} & -- & \pub{61.92} & \pub{41.92} \\
    \cod~\cite{liu2025codriving} & -- & \pub{83.99} & \pub{66.10} & -- & \pub{79.06} & \pub{70.53} & -- & \pub{84.97} & \pub{82.84} & -- & \pub{82.67} & \pub{73.16} \\
    Rule-based$^{\ast\ast}$~\cite{mdrive2026} & -- & \pub{83.77} & \pub{75.10} & -- & \pub{66.70} & \pub{53.65} & -- & \pub{77.89} & \pub{73.13} & -- & \pub{76.12} & \pub{67.29} \\
    CoLMDriver~\cite{liu2025colmdriver} & -- & \pub{85.29} & \pub{76.85} & -- & \pub{67.91} & \pub{51.93} & -- & \pub{78.27} & \pub{75.37} & -- & \pub{77.16} & \pub{68.05} \\
    \midrule
    WOR, our rerun~\cite{chen2021wor} & 70.4 & 66.59 & 43.09 & 73.3 & 60.05 & 33.38 & 88.8 & 84.65 & 73.88 & 77.5 & 70.43 & 50.12 \\
    LAV, our rerun~\cite{chen2022lav} & 57.1 & 63.36 & 47.94 & 58.8 & 68.62 & 55.42 & 81.3 & 83.80 & 70.15 & 65.8 & 71.93 & 57.84 \\
    TransFuser, our rerun~\cite{chitta2023transfuser} & 60.4 & 64.84 & 53.11 & 63.4 & 62.86 & 48.18 & 88.8 & 71.23 & 61.19 & 70.9 & 66.31 & 54.16 \\
    InterFuser, our rerun~\cite{shao2022interfuser} & 9.8 & 32.15 & 8.83 & 23.1 & 42.78 & 23.08 & 85.1 & 87.67 & 77.61 & 39.3 & 54.20 & 36.50 \\
    UniAD, our rerun~\cite{hu2023uniad} & 32.9 & 50.04 & 24.31 & 35.6 & 54.01 & 28.91 & 74.6 & 79.74 & 68.66 & 47.7 & 61.27 & 40.63 \\
    VAD, our rerun~\cite{jiang2023vad} & 29.0 & 30.30 & 8.46 & 39.6 & 48.10 & 24.33 & 79.9 & 67.31 & 50.75 & 49.5 & 48.57 & 27.85 \\
    LMDrive, our rerun~\cite{shao2024lmdrive} & 72.2 & 53.68 & 30.37 & 79.3 & 58.85 & 32.89 & 84.3 & 76.49 & 58.96 & 78.6 & 63.01 & 40.74 \\
    TCP, our rerun~\cite{wu2022tcp} & 82.2 & 77.34 & 60.51 & \second{88.3} & 73.86 & 52.63 & \best{92.5} & 79.83 & 55.22 & \second{87.7} & 77.01 & 56.12 \\
    No fusion$^{\ast}$~\cite{liu2025codriving} & 72.6 & 78.92 & 72.01 & 71.9 & 79.94 & \second{71.01} & 88.8 & \second{91.19} & \second{84.33} & 77.8 & 83.35 & 75.78 \\
    Early fusion~\cite{liu2025codriving} & 67.9 & 75.47 & 66.94 & 73.4 & 79.47 & 68.03 & 89.6 & \best{92.41} & \best{86.57} & 77.0 & 82.45 & 73.84 \\
    Late fusion~\cite{liu2025codriving} & 78.2 & 82.21 & 75.85 & 77.5 & 81.01 & \best{72.08} & 89.1 & 91.15 & \second{84.33} & 81.6 & 84.79 & \best{77.42} \\
    F-Cooper~\cite{chen2019fcooper} & 71.6 & 75.53 & 64.31 & 76.2 & \second{81.47} & 70.82 & 89.6 & 90.09 & 82.84 & 79.1 & 82.36 & 72.66 \\
    V2X-ViT~\cite{xu2022v2xvit} & 63.3 & 59.10 & 27.34 & 64.5 & 67.51 & 30.79 & 82.1 & 79.82 & 53.73 & 70.0 & 68.81 & 37.29 \\
    \cod, our rerun~\cite{liu2025codriving} & 71.7 & 78.19 & 71.74 & 71.2 & 79.20 & 69.14 & 88.8 & 91.00 & \second{84.33} & 77.3 & 82.80 & 75.07 \\
    Defer to Plan$^{\ddagger}$~\cite{li2026defer} & 85.1 & \second{86.82} & \second{78.14} & 78.7 & 79.59 & 70.84 & 87.1 & 88.99 & 82.59 & 83.6 & \second{85.13} & 77.19 \\
    Rule-based$^{\ast\ast}$, our rerun~\cite{liu2025colmdriver} & 82.6 & 75.05 & 66.67 & 73.7 & 64.79 & 49.45 & 88.1 & 79.85 & 67.91 & 81.4 & 73.23 & 61.34 \\
    CoLMDriver, our rerun~\cite{liu2025colmdriver} & 78.7 & 74.85 & 62.54 & 59.3 & 58.81 & 39.48 & 84.6 & 73.66 & 59.23 & 74.2 & 69.11 & 53.75 \\
    CoLMIN, our adaptation~\cite{colmin2026} & \second{88.1} & 78.09 & 70.35 & 74.9 & 69.06 & 51.44 & 86.9 & 75.13 & 62.31 & 83.3 & 74.09 & 61.37 \\
    \gr{Expert, rules (privileged)} & \gr{95.8} & \gr{90.54} & \gr{85.71} & \gr{93.5} & \gr{82.45} & \gr{73.94} & \gr{50.7} & \gr{29.57} & \gr{22.39} & \gr{80.0} & \gr{67.52} & \gr{60.68} \\
    \midrule
    \rowcolor{oursbg}
    \textbf{\ours (ours)} & \sota{97.4} & \sota{92.18} & \sota{81.82} & \sota{95.8} & \sota{84.82} & 69.83 & \second{91.8} & 87.48 & 80.35 & \sota{95.0} & \sota{88.16} & \second{77.33} \\
    \bottomrule
  \end{tabularx}
  \caption{\textbf{MDrive per bucket.} Share of controlled vehicles without a collision (No coll., \%), driving score and success rate (\%). Top: published values from~\cite{mdrive2026}, which does not report collisions. Middle: our reruns, repetition $0$ (late fusion and Defer to Plan: $0$ to $2$). \ours: repetitions $3$ to $5$. Every rerun and \ours cover all $225$ scenarios (CoLMDriver: $223$, two fail at start in its own agent; CoLMIN, our adaptation of its release to MDrive: $223$), and the repetitions of a scenario are averaged first. Scenarios that stall twice are scored at the second stall ($56$ of the rule variant's, $45$ of CoLMDriver's and $49$ of CoLMIN's, \cref{sec:supp_setup}). Avg: mean of the bucket values. \gr{Expert}: MDrive's privileged AutoPilot, as in \cref{tab:main} (\cref{sec:supp_setup}). WOR, LAV, TransFuser and InterFuser: V2Xverse's released adapters, one instance per vehicle, as on InterDrive (InterFuser without its safety-aware controller). $^{\ast}$, $^{\ddagger}$, $^{\ast\ast}$: as in \cref{tab:main}. \textbf{Bold}: best, \underline{underline}: second, among the values not in gray.}
  \label{tab:supp_mdrive}
\end{table*}

\noindent\textbf{MDrive's collision rule.}
MDrive's release ends a vehicle's route at its first collision (the vehicle stays in the scene), although the collision criterion of each route is built not to end it. The check came with a release of April 2026, and the published InterDrive-bucket values likely predate it. We keep the release as it is for every method. The rule costs most where vehicles collide most. If every vehicle whose route ended at a collision had finished the rest of it without a further infraction, \ours would gain at most $1.7$ points and the reruns $3.3$ to $13.2$. With half of the remaining route finished instead, our gain would be $0.9$, against $1.6$ to $6.6$ for the reruns. The driving-score margins over the strongest reruns, Defer to Plan ($3.0$) and late fusion ($3.4$), would then shrink to about $1.5$ with the half estimate and close with the upper bound. The share of collision-free vehicles does not depend on the rule.

\noindent\textbf{Matched information.}
On InterDrive and MDrive, the released cooperative baselines fuse each vehicle's LiDAR with that of its own roadside unit only; CoLMDriver and CoLMIN also exchange intents. We restrict \ours to the same inputs, its own roadside unit and no plan or intended trajectory of any other vehicle, so that the order gate never acts ($0$ of $30{,}443$ planning steps). It then scores $94.14$ without and $85.70$ with traffic (the repetitions of \cref{tab:main}), against $97.79$ and $90.70$ with its full inputs and $90.12$ and $78.93$ for CoLMIN. Paired with the full inputs, the differences are $-3.7$ ($95\,\%$ interval $[-8.4, +0.1]$) and $-5.0$ ($[-8.9, -1.7]$). The loss lies mostly in lane changes, which score $92.00$ and $80.29$ against $98.25$ and $96.84$. The restricted network has $9$ and $18$ vehicle collisions per $100$ tests, against $0$ and $1$ with the full inputs. With the same perception plus the broadcast plans and intended trajectories of up to three other vehicles (as CoLMDriver exchanges intents), it scores $96.06$ and $90.59$. Paired with the full inputs, the differences are $-1.7$ ($[-4.1, -0.1]$) and $-0.1$ ($[-1.8, +1.6]$). This variant has $0$ and $2$ vehicle collisions per $100$ tests, and lane changes recover to $94.97$ and $96.13$. The plans and the order on them thus carry the margin, and fusing the other vehicles' LiDAR adds little.

\Cref{tab:supp_interdrive,tab:supp_mdrive} list the components of the InterDrive and MDrive scores of \cref{tab:main} and the values of \cref{fig:teaser}.
On MDrive, \ours keeps the most vehicles free of collisions and has the highest driving score in the InterDrive and Interaction buckets and on average. Its driving score is $88.16$, $88.14$ and $88.17$ in repetitions $3$ to $5$, so repetition $3$ alone, used in \cref{tab:control} and \cref{sec:supp_gate,sec:supp_stock}, equals the mean; its share of collision-free vehicles there is $94.9\,\%$, against $95.0\,\%$ over the three. In the InterDrive and Interaction buckets and on average, its shares of collision-free vehicles are $97.4$, $95.8$ and $95.0\,\%$, against at most $88.1$, $88.3$ and $87.7\,\%$ for the baselines. On the V2X-PnP bucket, TCP keeps more vehicles free of collisions ($92.5$ against $91.8\,\%$). In driving score, InterFuser, Defer to Plan, \cod and its fusion variants but V2X-ViT score higher on V2X-PnP ($87.67$ to $92.41$ against $87.48$). In success rate, \ours is ahead in the InterDrive bucket. On average it is second, $0.1$ behind late fusion ($77.42$ against $77.33$; paired $-0.1$, $95\,\%$ interval $[-4.8, +4.7]$) and ahead of Defer to Plan ($77.19$).

\noindent\textbf{Learning against the rule-based expert.}
Each benchmark's privileged rule-based expert sees the ground truth. Yet, without deleting the actors in its way, it scores below \ours on all three: $83.89$ and $78.42$ on InterDrive without and with traffic, $62.95$ on V2Xverse and $67.52$ on MDrive. V2Xverse's expert with its deletion scores $74.24$ (repetition $0$; $53.93$ on the seeds of \ours).
\Cref{tab:supp_expert} splits InterDrive by category.
On lane merges both reach $100$ without traffic. The gap lies at intersections ($73.43$ against $95.12$) and lane changes ($67.15$ against $98.25$), where two controlled vehicles must settle who goes first.
There each expert vehicle decides on its own, from straight-line predictions of the others and without a shared order; the order gate alone adds $4.1$ points to \ours (\cref{tab:order}).
Two further gaps come from the inputs and the decision rule.
The expert predicts conflicts from the velocities the simulator reports, which the replayed traffic of MDrive's V2X-PnP bucket lacks ($29.57$ against $87.48$), while \ours estimates motion from its own sensor history.
Its stop rule is binary: the expert ends $42$ of $100$ V2Xverse runs blocked and still hits a pedestrian in $32$, against $29$ and $23$ for \ours.
\ours instead chooses among speed profiles with a learned collision risk, and its supervision drops the expert's stops that its own horizon cannot explain (\cref{sec:supervision}).

\begin{table}[t]
  \centering
  \footnotesize
  \setlength{\tabcolsep}{2.6pt}
  \begin{tabularx}{\linewidth}{@{}>{\raggedright\arraybackslash}X cccccccc@{}}
    \toprule
    & \multicolumn{4}{c}{no traffic} & \multicolumn{4}{c}{traffic} \\
    \cmidrule(lr){2-5}\cmidrule(l){6-9}
    Method & IC & LM & LC & all & IC & LM & LC & all \\
    \midrule
    Rule-based expert & 73.43 & \best{100.0} & 67.15 & 83.89 & \second{69.70} & 89.62 & \second{70.04} & 78.42 \\
    CoLMIN~\cite{colmin2026} & \second{90.71} & \second{98.75} & \second{67.20} & \second{90.12} & 66.47 & \second{94.11} & 67.45 & \second{78.93} \\
    \rowcolor{oursbg}
    \textbf{\ours} & \best{95.12} & \best{100.0} & \best{98.25} & \best{97.79} & \best{80.30} & \best{97.59} & \best{96.84} & \best{90.70} \\
    \bottomrule
  \end{tabularx}
  \caption{\textbf{InterDrive by category against the rule-based expert.} Intersections (IC, $18$ tests), lane merges (LM, $20$) and lane changes (LC, $8$), current result script. The expert sees ground truth; CoLMIN: published values. \textbf{Bold}: best, \underline{underline}: second.}
  \label{tab:supp_expert}
\end{table}

\Cref{tab:supp_scale} splits the InterDrive scores by the number of controlled vehicles (\cref{fig:scale}) and by town. With two vehicles, negotiation works: CoLMDriver and its rule variant score $97.1$ and $97.5$ without traffic and $95.3$ and $92.9$ with it, against $100.0$ and $95.2$ for \ours. With more vehicles it does not: with three or four they score $66.0$ and $64.0$, and with six to eight $15.8$ and $44.1$ without traffic, against $100.0$ and $87.3$; seven or eight vehicles also cause most of CoLMDriver's livelocks. The margin of \ours over the strongest cooperative rerun is thus $2.5$, $30.4$ and $32.3$ points with $2$, $3$--$4$ and $6$--$8$ vehicles without traffic, and $-0.1$, $29.8$ and $14.8$ with it. By town, it is largest in Town05, which training never saw ($25.0$ and $24.0$ points), and $9.7$ and $5.4$ in Town06. Town07, the validation town, holds only two-vehicle tests, where CoLMDriver matches \ours without traffic and leads with it ($93.6$ against $87.7$).

\begin{table*}[t]
  \centering
  \footnotesize
  \setlength{\tabcolsep}{3pt}
  \begin{tabularx}{\textwidth}{@{}l *{12}{>{\centering\arraybackslash}X}@{}}
    \toprule
    & \multicolumn{6}{c}{InterDrive, no traffic} & \multicolumn{6}{c}{InterDrive, with traffic} \\
    \cmidrule(lr){2-7}\cmidrule(l){8-13}
    & \multicolumn{3}{c}{controlled vehicles} & \multicolumn{3}{c}{town} & \multicolumn{3}{c}{controlled vehicles} & \multicolumn{3}{c}{town} \\
    \cmidrule(lr){2-4}\cmidrule(lr){5-7}\cmidrule(lr){8-10}\cmidrule(l){11-13}
    Method & 2 & 3--4 & 6--8 & 05$^{\circ}$ & 06 & 07$^{\circ}$ & 2 & 3--4 & 6--8 & 05$^{\circ}$ & 06 & 07$^{\circ}$ \\
    \midrule
    No fusion$^{\ast}$ & 91.3 & 63.6 & 40.3 & 67.8 & 85.3 & 68.0 & 75.1 & 45.8 & 36.8 & 44.2 & 85.3 & 62.8 \\
    Early fusion & 84.0 & \second{69.6} & 22.8 & 65.6 & 75.5 & 64.9 & 71.0 & 39.2 & 15.6 & 37.3 & 66.1 & 74.9 \\
    Late fusion & 84.5 & 61.2 & 48.4 & 60.2 & 88.4 & 75.3 & 73.2 & 55.2 & 53.4 & 51.6 & 88.4 & 63.3 \\
    F-Cooper & 79.9 & 60.3 & 36.6 & 57.3 & 73.6 & 85.5 & 64.1 & 46.2 & 40.1 & 36.7 & 83.1 & 66.4 \\
    V2X-ViT & 65.2 & 43.7 & 4.8 & 46.8 & 47.1 & 49.2 & 57.3 & 35.4 & 3.9 & 41.8 & 42.1 & 30.6 \\
    \cod & 85.5 & 61.6 & 42.0 & 62.1 & 85.8 & 68.0 & 77.5 & 46.1 & 43.0 & 47.7 & 85.8 & 63.8 \\
    Defer to Plan$^{\ddagger}$ & 87.0 & 63.1 & \second{55.0} & 64.8 & \second{89.2} & 75.7 & 79.4 & 60.9 & \second{61.6} & 58.2 & \second{93.3} & 71.5 \\
    Rule-based$^{\ast\ast}$ & \second{97.5} & 64.0 & 44.1 & \second{71.7} & 80.1 & \second{90.8} & 92.9 & 49.2 & 53.1 & \second{63.5} & 79.9 & 83.6 \\
    CoLMDriver~\cite{liu2025colmdriver} & 97.1 & 66.0 & 15.8 & 71.5 & 60.5 & \best{100.0} & \best{95.3} & \second{61.8} & 12.1 & 63.0 & 70.8 & \best{93.6} \\
    \midrule
    \rowcolor{oursbg}
    \textbf{\ours (ours)} & \sota{100.0} & \sota{100.0} & \sota{87.3} & \sota{96.7} & \sota{98.9} & \sota{100.0} & \second{95.2} & \sota{91.6} & \sota{76.4} & \sota{87.5} & \sota{98.7} & \second{87.7} \\
    \bottomrule
  \end{tabularx}
  \caption{\textbf{InterDrive by number of controlled vehicles and by town.} Driving score per group of tests ($22$, $16$ and $8$ tests with $2$, $3$--$4$ and $6$--$8$ controlled vehicles; $27$, $13$ and $6$ tests in Town05, 06 and 07). $^{\circ}$: town not in the training data (Town07: validation). Our reruns: repetition $0$ (CoLMDriver: its livelocked tests at the state reached). \ours: repetition $3$ (with traffic $3$ to $5$). Published methods report no per-test scores. $^{\ast}$, $^{\ddagger}$: as in \cref{tab:main}. \textbf{Bold}: best, \underline{underline}: second, among the values not in gray.}
  \label{tab:supp_scale}
\end{table*}

\begin{table*}[t]
  \centering
  \footnotesize
  \setlength{\tabcolsep}{4pt}
  \begin{tabularx}{\textwidth}{@{}>{\raggedright\arraybackslash}X cc cc cc@{}}
    \toprule
    & \multicolumn{2}{c}{V2Xverse} & \multicolumn{2}{c}{InterDrive, no traffic} & \multicolumn{2}{c}{InterDrive, traffic} \\
    \cmidrule(lr){2-3}\cmidrule(lr){4-5}\cmidrule(l){6-7}
    Variant & $\Delta$DS & $95\,\%$ interval & $\Delta$DS & $95\,\%$ interval & $\Delta$DS & $95\,\%$ interval \\
    \midrule
    \grouprow{7}{Route-time conflict map}
    \quad w/o map (tokens only) & {\boldmath$-6.4$} & {\boldmath$[-12.9, -0.1]$} & $-4.6$ & $[-10.6, +0.3]$ & $-0.5$ & $[-4.4, +3.6]$ \\
    \quad w/o partner plans in the map & $-2.0$ & $[-9.1, +4.9]$ & {\boldmath$-6.0$} & {\boldmath$[-11.8, -1.5]$} & $-1.6$ & $[-5.7, +2.4]$ \\
    \quad w/o forecasts in the map & $-1.3$ & $[-8.0, +5.4]$ & {\boldmath$-3.3$} & {\boldmath$[-6.2, -0.8]$} & $-2.0$ & $[-6.0, +1.4]$ \\
    \quad one class channel & $+3.0$ & $[-4.0, +9.8]$ & $-0.5$ & $[-2.2, +1.1]$ & $-1.0$ & $[-5.1, +3.3]$ \\
    \quad BEV occupancy instead of the map & $-5.9$ & $[-13.3, +1.3]$ & $-1.8$ & $[-5.3, +0.6]$ & $-1.3$ & $[-5.3, +2.5]$ \\
    \grouprow{7}{Cooperative fusion and planner}
    \quad w/o reliability estimate & $-6.4$ & $[-14.1, +1.1]$ & {\boldmath$-2.2$} & {\boldmath$[-3.9, -0.7]$} & $+1.2$ & $[-1.9, +4.9]$ \\
    \quad w/o risk envelope (regressed profile) & $-1.0$ & $[-7.6, +5.6]$ & {\boldmath$-4.2$} & {\boldmath$[-9.0, -0.3]$} & $-1.4$ & $[-4.8, +1.8]$ \\
    \quad w/o candidate imitation loss & $-1.9$ & $[-8.6, +4.8]$ & $+0.4$ & $[-0.5, +1.5]$ & $+0.3$ & $[-3.2, +3.7]$ \\
    \grouprow{7}{Supervision}
    \quad expert plans as recorded & $-6.4$ & $[-14.6, +1.8]$ & {\boldmath$-4.5$} & {\boldmath$[-8.4, -1.2]$} & $-2.5$ & $[-8.4, +3.5]$ \\
    \quad unexplained stops released only & $+0.5$ & $[-6.7, +7.4]$ & {\boldmath$-2.2$} & {\boldmath$[-4.7, -0.4]$} & $-3.0$ & $[-8.4, +2.6]$ \\
    \quad distance keeping only & $-3.9$ & $[-11.4, +3.7]$ & $-2.3$ & $[-6.4, +0.6]$ & $+0.3$ & $[-4.1, +4.5]$ \\
    \quad w/o commit or yield early & $-4.2$ & $[-11.3, +3.0]$ & $-3.0$ & $[-7.2, +0.1]$ & $-4.5$ & $[-9.3, +0.4]$ \\
    \quad w/o box test in the risk labels & $-3.6$ & $[-11.1, +3.8]$ & {\boldmath$-10.9$} & {\boldmath$[-17.6, -5.0]$} & $-4.8$ & $[-10.2, +0.6]$ \\
    \quad w/o virtual partners & $-0.7$ & $[-7.4, +6.2]$ & {\boldmath$-6.1$} & {\boldmath$[-11.2, -1.9]$} & $-1.4$ & $[-5.4, +3.0]$ \\
    \bottomrule
  \end{tabularx}
  \caption{\textbf{Ablations with paired $95\,\%$ intervals.} The rows of \cref{tab:ablation}, repetition $3$; differences to the first line of \cref{tab:ablation}, paired over the same routes ($85$ V2Xverse runs) and tests ($46$ per InterDrive setting), intervals by bootstrap. Bold: the interval excludes zero. InterDrive without traffic is deterministic, so its intervals reflect the spread over tests only. With traffic, the first line is the run the ablations were compared with ($90.24$); the reported network scores $91.11$ on the same seeds.}
  \label{tab:supp_ablation_ci}
\end{table*}

\section{The benchmarks' default controller}\label{sec:supp_stock}
Our runs change two settings of the benchmarks' PID controller, on our side only: it steers by the network's waypoints instead of the route point, and its forced push is off (\cref{sec:training}). With both default settings, the same weights score as follows (repetition $3$; paired over routes or scenarios against the reported repetition $3$, which with traffic scores $91.11$; the reference run of \cref{tab:ablation} is another run, $90.24$).
\emph{V2Xverse} (repetitions $3$ to $5$): $59.77$ against $68.31$ (paired $-8.5$, $95\,\%$ interval $[-12.7, -4.3]$), route completion $91.5$ against $85.7\,\%$, infraction score $0.636$ against $0.797$. Vehicle collisions rise from $16$ to $128$ in the $255$ runs, and $100$ of the $128$ come within $5$\,s after a push, which fires in $97$ of the runs. On the same seeds the default-controlled network still scores above every baseline (\cref{sec:supp_v2xverse}), resolved for all but TransFuser ($+3.3$, $[-1.9, +8.5]$). The network leads late fusion by $+7.9$ ($[+2.7, +12.8]$) and V2X-ViT by $+5.8$ ($[+0.5, +10.9]$). With every interval above $+4$, it leads WOR by $+9.8$, F-Cooper by $+9.9$, no fusion by $+11.7$, Defer to Plan by $+12.5$, early fusion by $+13.0$, TCP by $+14.3$, \cod by $+14.6$, LAV by $+14.9$ and InterFuser by $+15.9$. It also leads the privileged rule-based expert without its actor deletion ($+9.9$, $[+4.1, +15.5]$) and, within noise, the expert with it ($+5.9$, $[-0.2, +12.0]$).
\emph{MDrive}: $89.49$ against $88.16$ (paired $+1.3$, $[-0.1, +2.9]$; all $225$ scenarios). In repetition $3$, the share of collision-free vehicles falls from $94.9$ to $89.6\,\%$, still above the safest baseline ($87.7\,\%$), and $59$ of the $80$ vehicle collisions follow a push within $5$\,s.
\emph{InterDrive with traffic}: $90.21$ against $91.11$ (paired $-0.9$, $[-5.0, +3.5]$), with $7$ vehicle collisions in $4$ of $46$ tests against none, all within $5$\,s after a push.
The forced push thus decides most of the extra collisions; it is the one rule of the controller that overrides the plan.

\section{Ablations with confidence intervals}\label{sec:supp_ablation_ci}
\begin{figure*}[t]
  \centering
  \includegraphics[width=0.9\textwidth]{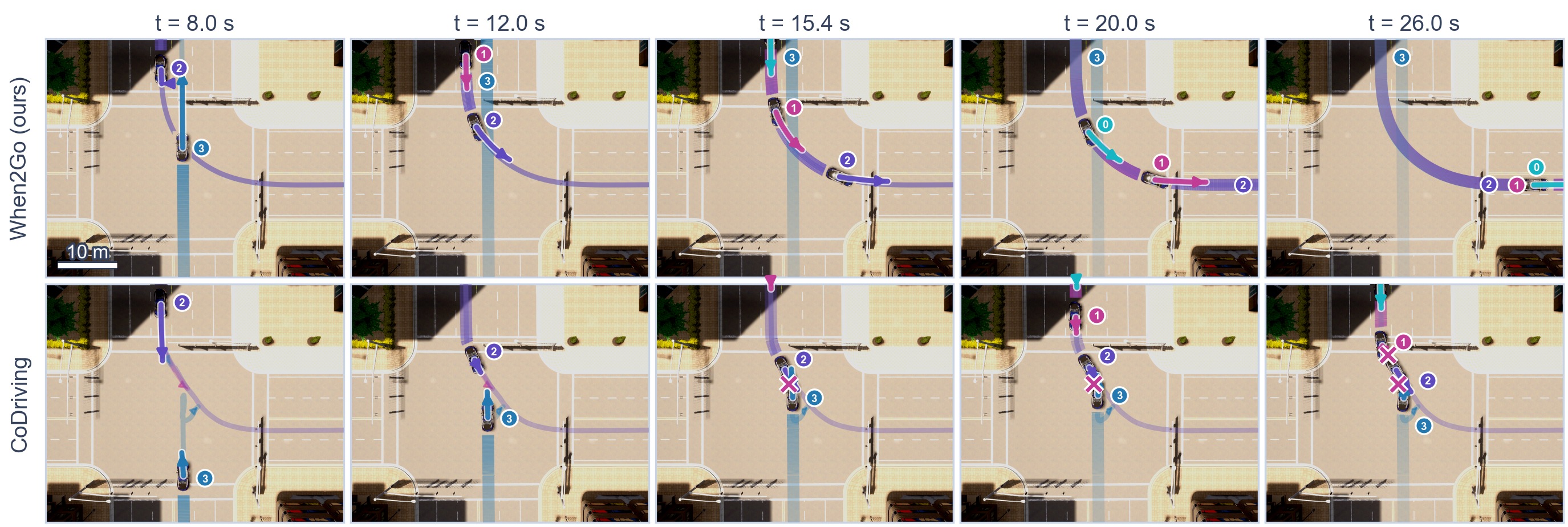}
  \caption{\textbf{Four vehicles at a left turn: the MDrive row of \cref{fig:qual} over time.} MDrive Unprotected Left Turn (Town05), repetition 3, one fixed camera. Top: \ours lets vehicle 3 cross first while 2 waits at the entry; 2, 1 and 0 then turn one after another, and all four arrive without a collision (DS $100$). Bottom: \cod~\cite{liu2025codriving} as released turns vehicle 2 into the path of 3, and they collide at $15.3$\,s (\textcolor[HTML]{C13C95}{$\times$}); vehicle 1 runs into 2 at $25.3$\,s, and none of the four arrives (DS $16.6$).}
  \label{fig:qual_mdrive_seq}
\end{figure*}
\begin{figure*}[t]
  \centering
  \includegraphics[width=0.9\textwidth]{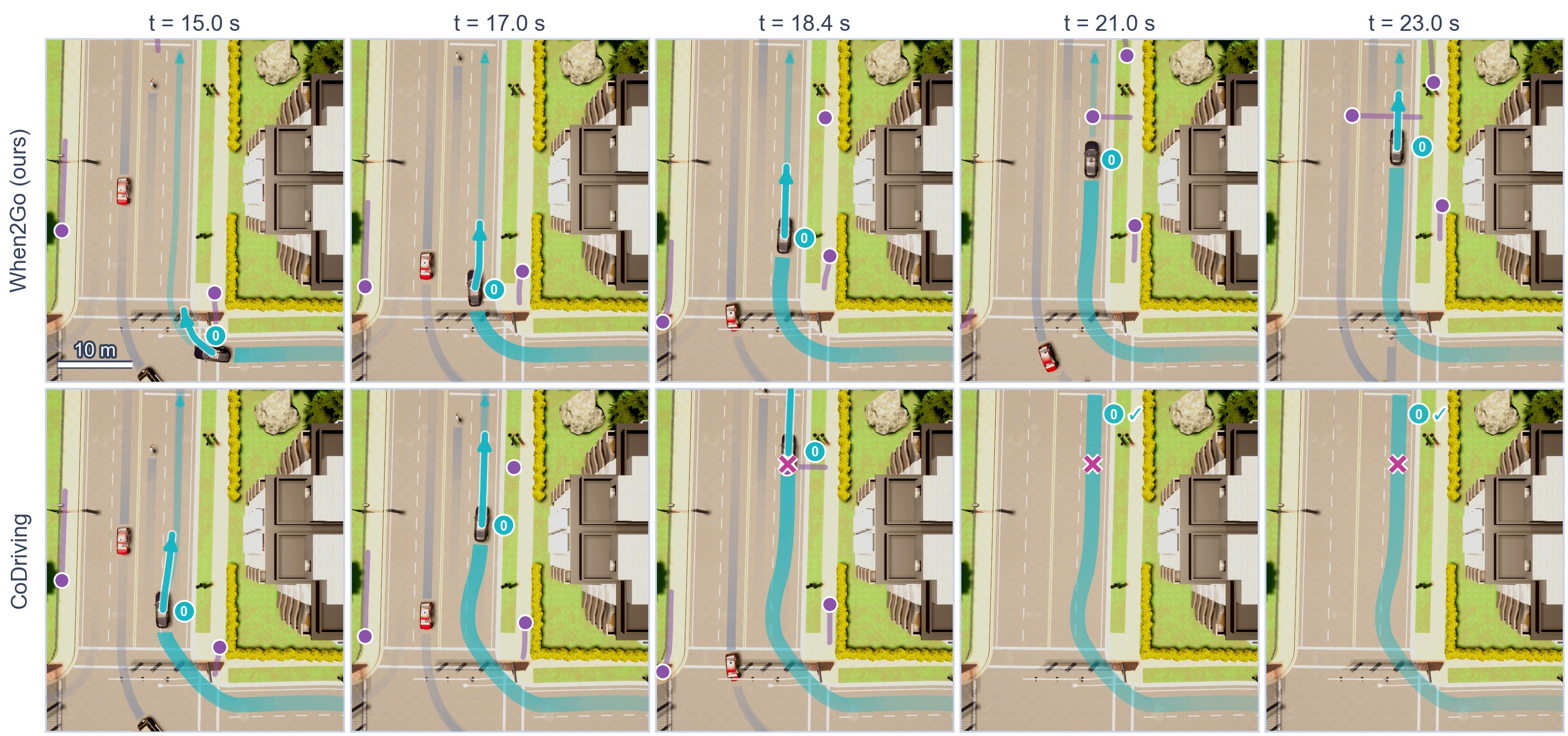}
  \caption{\textbf{A crossing pedestrian: the full sequence of the V2Xverse row of \cref{fig:qual}.} V2Xverse public route r3: one controlled vehicle among background vehicles (gray paths) and pedestrians (purple dots with their last $3$\,s); same camera and instants in both rows. The benchmark releases the pedestrian when the vehicle reaches a trigger point. \cod hits the pedestrian at $18.3$\,s (\textcolor[HTML]{C13C95}{$\times$}) and completes the route at $19.5$\,s (DS $50$); \ours stops in front of the pedestrian, drives on once it has crossed and completes the route at $25.8$\,s without an infraction (DS $100$).}
  \label{fig:qual_v2xverse}
\end{figure*}

\begin{figure*}[t]
  \centering
  \includegraphics[width=\textwidth]{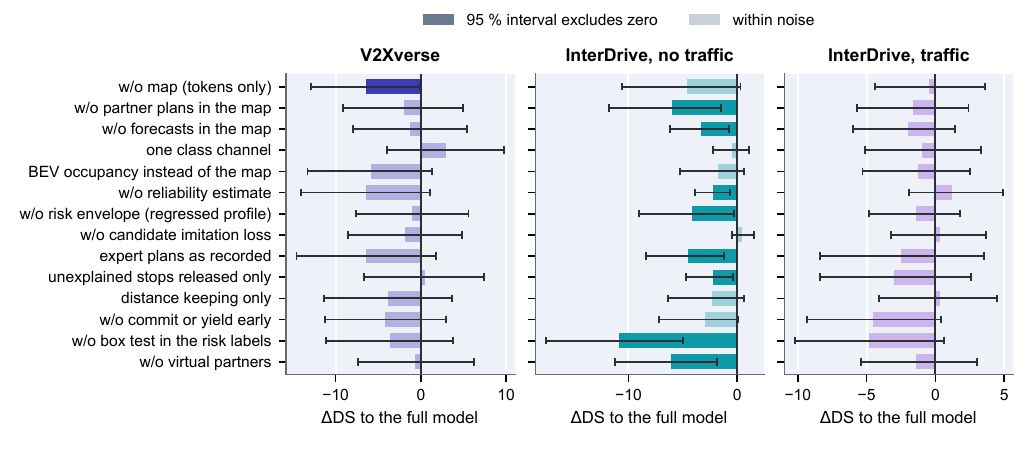}
  \caption{\textbf{Ablations with paired $95\,\%$ intervals.} The values of \cref{tab:supp_ablation_ci}; solid bars: the interval excludes zero.}
  \label{fig:supp_ablation}
\end{figure*}

\Cref{tab:supp_ablation_ci,fig:supp_ablation} give the paired $95\,\%$ intervals of every ablation of \cref{tab:ablation}. Without traffic, where InterDrive is deterministic, eight of the fourteen changes cost significantly. On V2Xverse only the map's effect is resolved ($-6.4$, $[-12.9, -0.1]$), and with traffic none is.

\noindent\textbf{Three training seeds.}
The rows of \cref{tab:ablation} are single retrainings, and another training seed of the full recipe alone moves the score by up to $2.9$ points without traffic (\cref{tab:supp_seeds}). We therefore also trained the two variants that replace the map, tokens only and BEV occupancy, with seeds $1$ and $2$ and ran them on InterDrive without traffic (repetition $3$, $46$ tests). The full recipe scores $97.79$, $95.78$ and $94.89$ with seeds $0$ to $2$, the variant without the map $93.22$, $95.30$ and $96.10$, and the BEV variant $96.00$, $95.77$ and $97.67$; every variant stays above the best published method ($90.12$). Averaged over the three seeds, the full recipe leads the variant without the map by $1.3$ ($[-1.2, +4.4]$, paired over tests; $[-2.0, +5.5]$ when the seeds are resampled as well), and its difference to the BEV variant is $-0.3$ ($[-2.1, +1.4]$). Without traffic, the three ways to draw the inputs are thus within the seed spread, and the single-seed difference of \cref{tab:ablation} ($-4.6$) lies at its end.

\section{Qualitative results}\label{sec:supp_qual}
\begin{figure*}[t]
  \centering
  \includegraphics[width=\textwidth]{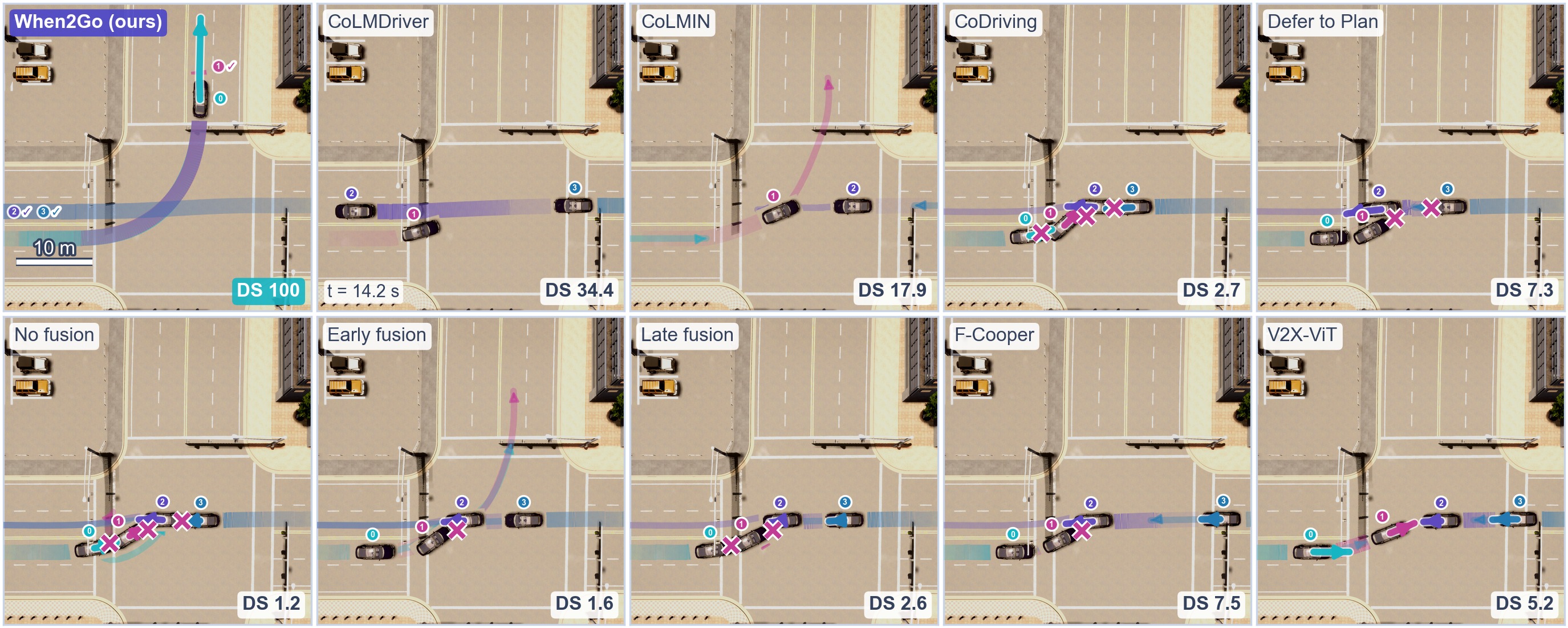}
  \caption{\textbf{One junction, ten methods, the same instant.} InterDrive r32 at $t{=}18$\,s: vehicles 0 and 1 turn left across the oncoming 2 and 3. \ours lets 2 and 3 pass first, and all four arrive without a collision (the last, 0, at $18.4$\,s); the other methods collide with the oncoming vehicles (\textcolor[HTML]{C13C95}{$\times$}) or block the junction (CoLMIN's 1 and 2 collide at $28.2$\,s). CoLMDriver's test was stopped at $14.4$\,s, after its critic had repeated one answer for $30$\,min without a new frame, so its panel shows its last recorded instant ($14.2$\,s) and its DS is that of the state reached, as in the tables. Corner: DS of this test; \checkmark: arrived. The scene of \cref{fig:overview}; its full sequence for \ours and \cod is \cref{fig:qual_interdrive_full}.}
  \label{fig:qual_r32_methods}
\end{figure*}

\begin{figure*}[t]
  \centering
  \includegraphics[width=\textwidth]{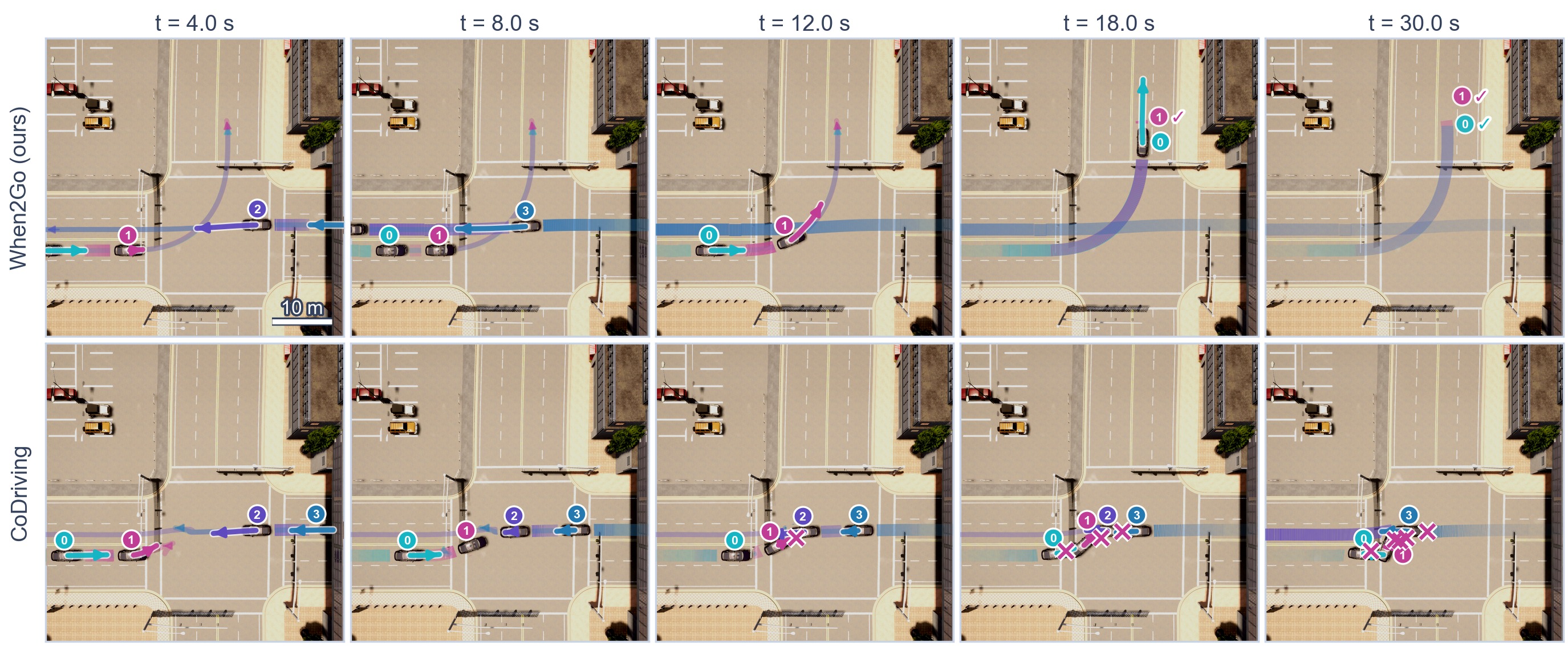}
  \caption{\textbf{Same junction, same instants: the full sequence of \cref{fig:qual_r32_methods}.} InterDrive r32 (four vehicles, Town05), one fixed camera; vehicles 0 and 1 turn left across the path of the oncoming vehicles 2 and 3. Top: \ours lets 2 and 3 pass, then turns; all four routes are completed without a collision (DS $100$). Bottom: \cod~\cite{liu2025codriving} as released turns vehicle 1 into the path of vehicle 2 ($9.9$\,s); the vehicles then block each other in the junction, collide four more times (\textcolor[HTML]{C13C95}{$\times$}) and three of them time out (DS $2.71$). A check mark marks a vehicle that has arrived, which the benchmark removes from the scene.}
  \label{fig:qual_interdrive_full}
\end{figure*}

\begin{figure*}[t]
  \centering
  \includegraphics[width=0.93\textwidth]{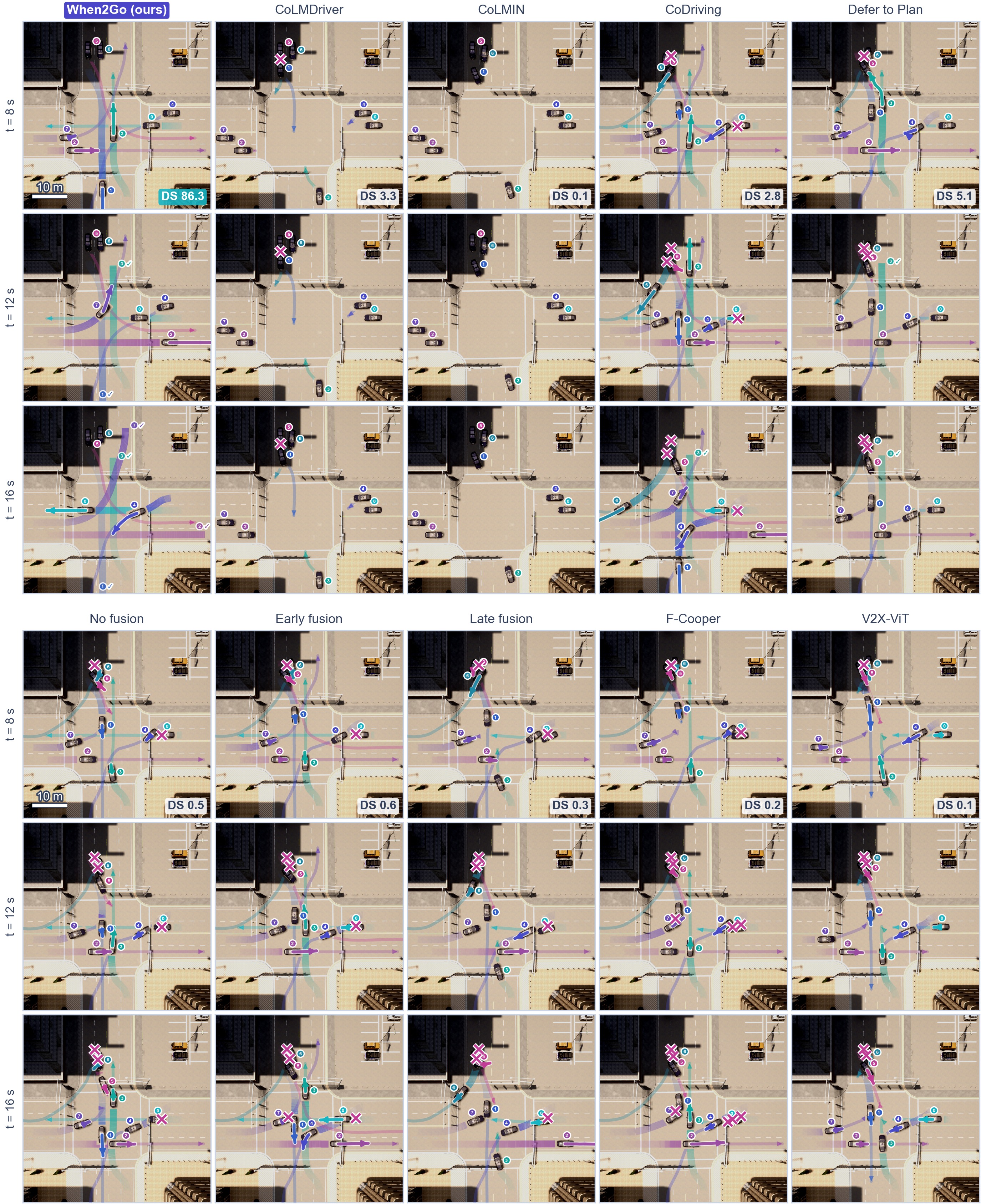}
  \caption{\textbf{Eight vehicles, ten methods, three instants: the InterDrive row of \cref{fig:qual} over time, with CoLMIN.} InterDrive r44 (Town05, no background traffic), one fixed camera. Every rerun baseline but CoLMIN has collided by $8$\,s (\textcolor[HTML]{C13C95}{$\times$}) and keeps colliding. CoLMIN's vehicles hardly move (by $20.9$\,s seven have completed none of their route, the eighth $3\,\%$), two of them collide at $20.8$\,s, and its agent then stops with an error in its negotiation code; with the $4{,}096$-token context of its README its request for eight vehicles does not fit, so this run serves its language model with $8{,}192$ tokens. \ours has no collision in the whole test, and two of its vehicles do not finish (one is blocked, one times out; DS $86.3$). Corner: DS of the test (CoLMIN: of the state in which its agent stopped); \checkmark: arrived. CoLMIN's InterDrive panels here and in \cref{fig:qual,fig:qual_r32_methods} come from figure runs in the setup of CoLMDriver's reruns (its InterDrive rows are published), its V2Xverse panel from its rerun of \cref{tab:main} and its MDrive panel from our adaptation (\cref{sec:supp_setup}).}
  \label{fig:qual_r44_six}
\end{figure*}

The figures replay runs made for them with the reported weights (\ours) and the reruns' configurations; a corner value is the driving score of the run shown, and no figure run enters a table. The scenes (busy junctions with turns, where the methods differ most) illustrate behaviour and are not a sample. Per vehicle, the figures show the driven path (fading with age), the path still ahead (light), the broadcast plan (bold arrow) and its ID; a check mark marks an arrived vehicle, a cross a collision.
\Cref{fig:qual_mdrive_seq,fig:qual_v2xverse,fig:qual_r44_six} follow the rows of \cref{fig:qual} over time, and \cref{fig:qual_r32_methods,fig:qual_interdrive_full} show the scene of \cref{fig:overview} for all ten methods and over time.

\end{document}